\documentclass{article}

\usepackage{microtype}
\usepackage{graphicx}
\usepackage{subfigure}
\usepackage{multirow}
\usepackage{array}
\usepackage{booktabs} % for professional tables
\usepackage{amsmath,amsfonts,bm}
\usepackage{amssymb}
\usepackage{dblfloatfix}

\usepackage{hyperref}

\usepackage[accepted]{icml2020}

\icmltitlerunning{Stabilizing DARTS with Amended Gradient Estimation on Architectural Parameters}

\begin{document}

\twocolumn[
\icmltitle{Stabilizing DARTS\\with Amended Gradient Estimation on Architectural Parameters}



\icmlsetsymbol{equal}{*}

\begin{icmlauthorlist}
\icmlauthor{Kaifeng Bi}{tsinghua,huawei}
\icmlauthor{Changping Hu}{tsinghua,huawei}
\icmlauthor{Lingxi Xie}{huawei}
\icmlauthor{Xin Chen}{tongji,huawei}
\icmlauthor{Longhui Wei}{huawei}
\icmlauthor{Qi Tian}{huawei}
\end{icmlauthorlist}

\icmlaffiliation{tsinghua}{Tsinghua University}
\icmlaffiliation{huawei}{Huawei Inc.}
\icmlaffiliation{tongji}{Tongji University}

\icmlcorrespondingauthor{Lingxi Xie}{198808xc@gmail.com}
\icmlcorrespondingauthor{Qi Tian}{tian.qi1@huawei.com}

\icmlkeywords{Machine Learning, ICML}

\vskip 0.3in
]



\printAffiliationsAndNotice{\icmlEqualContribution} % otherwise use the standard text.

\begin{abstract}
DARTS is a popular algorithm for neural architecture search (NAS). Despite its great advantage in search efficiency, DARTS often suffers \textbf{weak stability}, which reflects in the large variation among individual trials as well as the sensitivity to the hyper-parameters of the search process. This paper owes such instability to an \textbf{optimization gap} between the super-network and its sub-networks, namely, improving the validation accuracy of the super-network does not necessarily lead to a higher expectation on the performance of the sampled sub-networks. Then, we point out that the gap is due to the inaccurate estimation of the architectural gradients, based on which we propose an amended estimation method. Mathematically, our method guarantees \textbf{a bounded error} from the true gradients while the original estimation does not. Our approach bridges the gap from two aspects, namely, amending the estimation on the architectural gradients, and unifying the hyper-parameter settings in the search and re-training stages. Experiments on CIFAR10 and ImageNet demonstrate that our approach largely improves search stability and, more importantly, enables DARTS-based approaches to explore much larger search spaces that have not been investigated before.
\end{abstract}

\section{Introduction}
\label{introduction}

Neural architecture search (NAS) has been an important topic in the research area of automated machine learning (AutoML). The idea is to replace the manual design of neural network architectures with an automatic algorithm, by which deep learning methods become more flexible in fitting complex data distributions, \textit{e.g.}, large-scale image datasets. Early efforts of NAS adopted heuristic search methods such as reinforcement learning~\cite{zoph2017neural,zoph2018learning} and evolutionary algorithms~\cite{real2017large,xie2017genetic} to sample networks from a large search space, and optimizing each sampled network individually to evaluate its quality. Despite notable success by this methodology, it often requires a vast amount of computation, which obstacles its applications in the scenarios of limited resources. Inspired by reusing and sharing parameters among trained networks, DARTS~\cite{liu2019darts} was designed as a `one-shot' solution of NAS. The idea is to incorporate all possibilities into a \textit{super-network} and then adopt a differentiable mechanism to optimize model weights (such as convolution) and architectural weights simultaneously, following which the best \textit{sub-network} is sampled to be the final architecture.

Although DARTS reduced the search cost by $1$--$2$ orders of magnitudes (to a few hours on a single GPU), it suffers a critical weakness known as \textbf{instability}. Researchers reported that DARTS-based algorithms can sometimes generate weird architectures that produce considerably worse accuracy than those generated in other individual runs~\cite{zela2020understanding}. Although some practical methods~\cite{chen2019progressive,nayman2019xnas} have been developed to reduce search variance, the following property of DARTS persists and has not been studied thoroughly: when DARTS gets trained for sufficiently long, \textit{e.g.}, extending the default number of $50$ epochs to $200$ epochs, almost \textbf{all} DARTS-based approaches converge to a dummy architecture in which all edges are occupied by \textit{skip-connect}. These architectures, with few trainable parameters, are often far from producing high accuracy, in particular, on large datasets like ImageNet, although the validation accuracy in the search stage continues growing with more epochs.

In essence, this observation refers to that an improved validation accuracy of the super-network does not necessarily lead to high-quality sub-networks to be sampled. We name it the \textbf{optimization gap} between search and re-training, and investigate it from a perspective which is less studied before. We reveal that DARTS has been using an inaccurate approximation to calculate the gradients with respect to the architectural parameters ($\boldsymbol{\alpha}$ as in the literature), and we amend the error by slightly modifying the second-order term in gradient computation. Mathematically, we prove that the amended term has a bounded error, \textit{i.e.}, the angle between the true and estimated gradients is smaller than $90^\circ$, while the original DARTS did not guarantee so. With this modification, the search performance is stabilized and the dummy all-\textit{skip-connect} architecture does not appear even after a very long search process. Consequently, one can freely allow the search process to arrive at convergence, and hence the sensitivity to hyper-parameters is alleviated.

Practically, our approach involves using an amended second-order gradient, so that the computational overhead is comparable to the \textit{second-order} version of DARTS. Experiments are performed on two popular image classification datasets, namely, CIFAR10 and ImageNet. In all experiments, the search process, after arriving at convergence, produces competitive architectures and classification accuracy comparable to the state-of-the-arts.

%The remainder of this paper is organized as follows. We briefly review related work in Section~\ref{related_work}, and illustrate our approach of amending architectural gradients in Section~\ref{approach}. After experiments are shown in Section~\ref{experiments}, we conclude this work in Section~\ref{conclusions}.

\section{Related Work}
\label{related_work}

With the era of big data and powerful computational resources, deep learning~\cite{lecun2015deep}, in particular, deep neural networks~\cite{krizhevsky2012imagenet}, have rapidly grown up to be the standard tool for learning representations in a complicated feature space. Recent years have witnessed the trend of using deeper~\cite{he2016deep} and denser~\cite{huang2017densely} networks to boost recognition performance, while there is no justification that whether these manually designed architectures are best for each specific task, \textit{e.g.}, image classification. Recently, researchers started considering the possibility of learning network architectures automatically from data, which led to the appearance of neural architecture search (NAS)~\cite{zoph2017neural}, which is now popular and known as a sub research field in automated machine learning (AutoML).

The common pipeline of NAS starts with a pre-defined space of network operators. Since the search space is often large (\textit{e.g.}, containing $10^{10}$ or even more possible architectures), it is unlikely that exhaustive search is tractable, and thus heuristic search methods are widely applied for speedup. Typical examples include reinforcement learning~\cite{zoph2017neural,zoph2018learning,liu2018progressive} and evolutionary algorithms~\cite{real2017large,xie2017genetic,real2019regularized}. These approaches followed a general pipeline that samples a set of architectures from a learnable distribution, evaluates them and learns from rewards by updating the distribution. In an early age, each sampled architecture undergoes an individual training process from scratch and thus the overall computational overhead is large, \textit{e.g.}, hundreds of even thousands of GPU-days. To alleviate the burden, researchers started to share computation among the sampled architectures, with the key lying in reusing network weights trained previously~\cite{cai2018efficient} or starting from a well-trained super-network~\cite{pham2018efficient}. These efforts shed light on the `one-shot' architecture search methods, which require training the super-network only once and thus run more efficiently, \textit{e.g.}, $2$--$3$ orders of magnitude faster than conventional approaches.

Within the scope of one-shot architecture search, an elegant solution lies in jointly formulating architecture search and approximation, so that it is possible to apply end-to-end optimization for training network and architectural parameters simultaneously. This methodology is known today as differentiable NAS, and a typical example is DARTS~\cite{liu2019darts}, which constructed a super-network with all possible operators contained and decoupled, and the goal is to determine the weights of these architectural parameters, followed by pruning and re-training stages. This kind of approach allowed more flexible search space to be constructed, unlike conventional approaches with either reinforcement or evolutionary learning, which suffer from the computational burden and thus must constrain search within a relatively small search space~\cite{tan2019efficientnet}.

Despite the inspirations by differentiable NAS, these approaches still suffer a few critical issues that narrow down their applications in practice. One significant drawback lies in the lack of stability~\cite{li2019random,sciuto2019evaluating}, which reflects in the way that results of differentiable search can be impacted by very small perturbations, \textit{e.g.}, initialization of architectural weights, training hyper-parameters, and even randomness in the training process. Existing solutions include running search for several individual times and choosing the best one in validation~\cite{liu2019darts}, or using other kinds of techniques such as decoupling modules~\cite{cai2019proxylessnas,guo2019single}, adjusting search space during optimization~\cite{noy2019asap,chen2019progressive,nayman2019xnas}, regularization~\cite{xu2020pc}, early termination~\cite{liang2019darts+}, \textit{etc}., however, these approaches seem to develop heuristic remedies rather than analyze it from the essence, \textit{e.g.}, how instability happens in mathematics, which this paper delves deep into this problem and presents a preliminary solution.

%In this paper, we investigate the stability issue in mathematics and show that the results produced by the current approaches are much less reliable than people used to think. Then, we fix this issue by amending optimization of the architectural parameters, so that each step of the update gets closer to the correct direction. We show great improvement on stability in a fundamental task, image classification, while we believe our approach can be applied to a wide range of tasks including object detection~\cite{ghiasi2019fpn}, semantic segmentation~\cite{liu2019auto}, hyper-parameter learning~\cite{cubuk2019autoaugment}, \textit{etc}.

\section{Stabilizing DARTS with Amended Architectural Gradient Estimation}
\label{approach}

\subsection{Preliminaries: DARTS}
\label{approach:preliminaries}

Differentiable NAS approaches start with defining a \textbf{super-network}, which is constrained in a search space with a pre-defined number of layers and a limited set of neural operators. The core idea is to introduce a `soft' way operator selection (\textit{i.e.}, using a weighted sum over the outputs of a few operators instead of taking the output of only one), so that optimization can be done in an end-to-end manner. Mathematically, the super-network is a function $\mathbf{f}\!\left(\mathbf{x};\boldsymbol{\omega},\boldsymbol{\alpha}\right)$, with $\mathbf{x}$ being input, and parameterized by network parameters $\boldsymbol{\omega}$ (\textit{e.g.}, convolutional kernels) and architectural parameters $\boldsymbol{\alpha}$ (\textit{e.g.}, indicating the importance of each operator between each pair of layers). $\mathbf{f}\!\left(\mathbf{x};\boldsymbol{\omega},\boldsymbol{\alpha}\right)$ is differentiable to both $\boldsymbol{\omega}$ and $\boldsymbol{\alpha}$, so that gradient-based approaches (\textit{e.g.}, SGD) can be applied for optimization.

In the example of DARTS, $\mathbf{f}\!\left(\mathbf{x};\boldsymbol{\omega},\boldsymbol{\alpha}\right)$ is composed of a few cells, each of which contains $N$ nodes, and there is a pre-defined set, $\mathcal{E}$, denoting which pairs of nodes are connected. For each connected node pair $\left(i,j\right)$, $i<j$, node $j$ takes $\mathbf{x}_i$ as input and propagates it through a pre-defined operator set, $\mathcal{O}$, and sums up all outputs:
\begin{equation}
\label{eqn:node_output}
{\mathbf{y}^{\left(i,j\right)}\left(\mathbf{x}_i\right)}={\sum_{o\in\mathcal{O}}\frac{\exp(\alpha_o^{\left(i,j\right)})}{\sum_{o'\in\mathcal{O}}\exp(\alpha_o^{\left(i,j\right)})}\cdot o\!\left(\mathbf{x}_i\right)}.
\end{equation}
Here, normalization is performed by computing softmax on the architectural weights. Within each unit of the search process, $\boldsymbol{\omega}$ and $\boldsymbol{\alpha}$ get optimized alternately. After that, the operator $o$ with the maximal value of $\alpha_o^{\left(i,j\right)}$ is preserved for each edge $\left(i,j\right)$. All network parameters $\boldsymbol{\omega}$ are discarded and the obtained architecture is re-trained from scratch.

\subsection{The Optimization Gap of DARTS}
\label{approach:gap}

\begin{figure}[!t]
\centering
\includegraphics[width=0.40\textwidth]{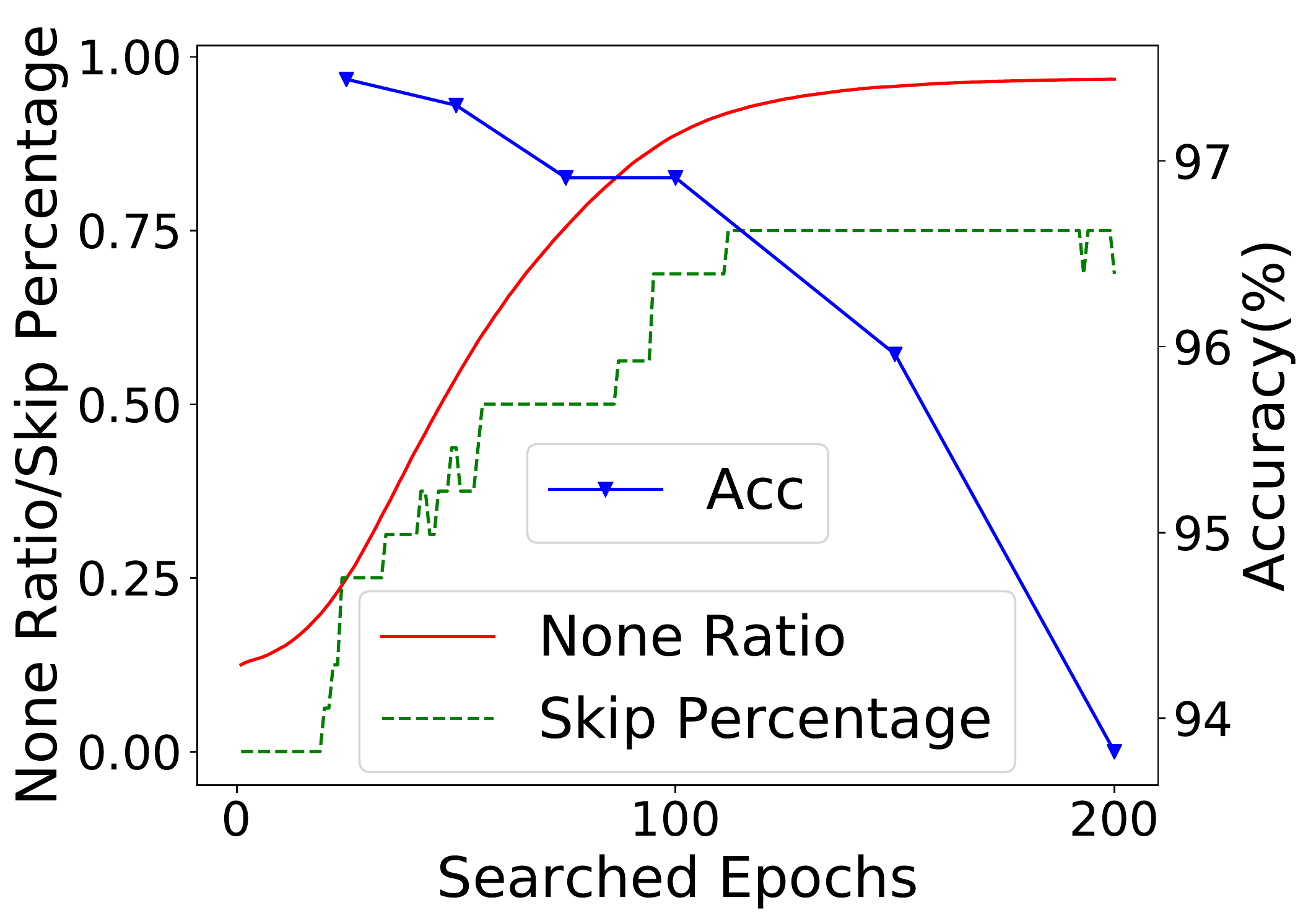}
\vspace{-0.2cm}
\includegraphics[width=0.45\textwidth]{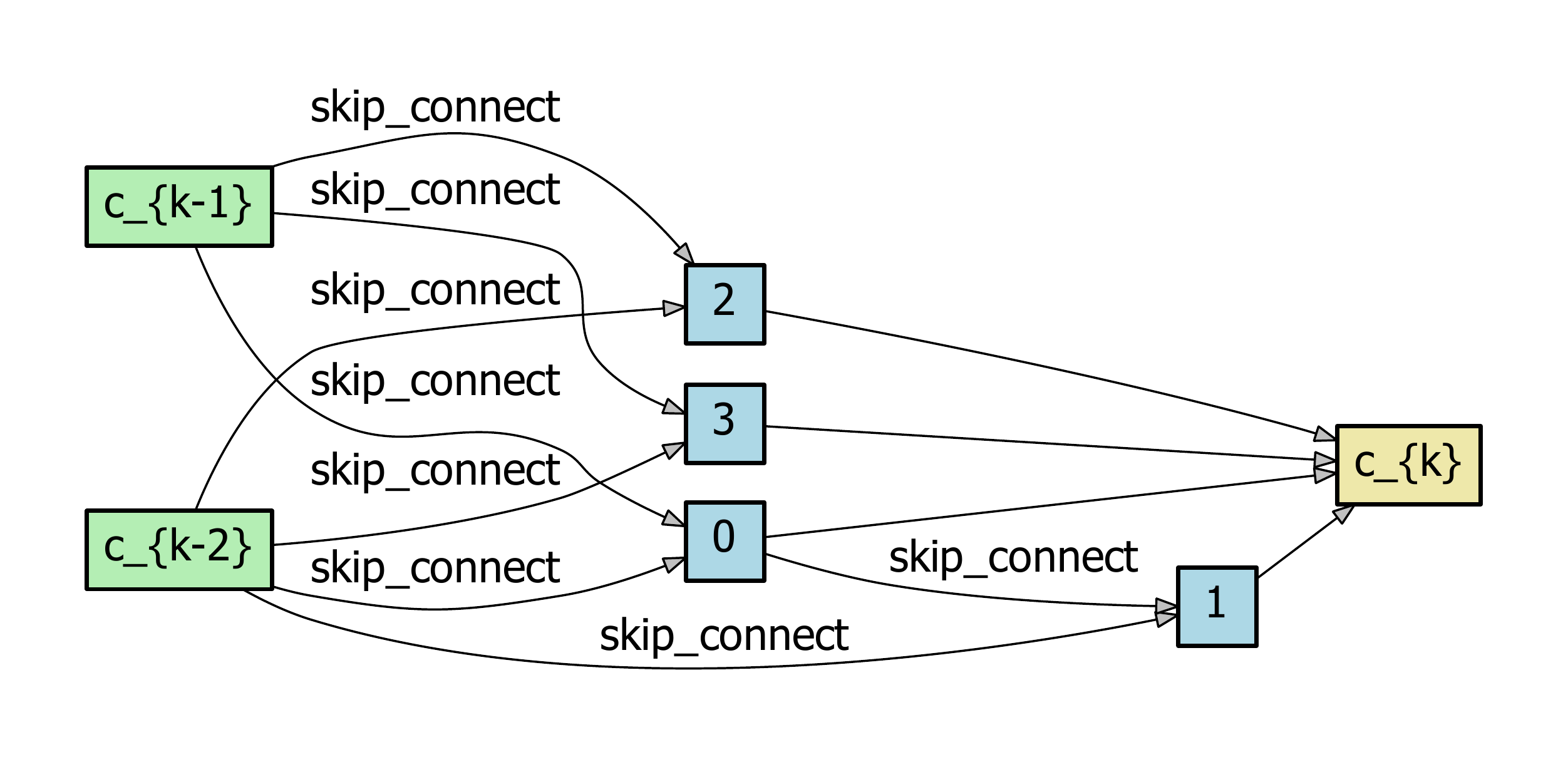}
\vspace{-0.4cm}
\caption{\textbf{Top}: a typical search process of the \textit{first-order} DARTS, in which $200$ search epochs are used. Red, green and blue lines indicate the average weight of \textit{none}, the ratio of preserved \textit{skip-connect} operators, and the re-training accuracy, respectively, with respect to the number of search epochs. \textbf{Bottom}: the normal cell after $200$ search epochs, in which all operators are \textit{skip-connect}. \textbf{Such failure consistently happens in several individual runs of both \textit{first-order} and \textit{second-order} DARTS.}}
\label{fig:convergence_performance}
\end{figure}

Our research is motivated by an observation that DARTS, at the end of a regular training process with, say, $50$ epochs~\cite{liu2019darts}, has not yet arrived at convergence. To verify this, we increase the length of each training stage from $50$ to $200$ epochs, and observe two weird facts shown in Figure~\ref{fig:convergence_performance}. \textbf{First}, the weight of the \textit{none} operator monotonically goes up -- at $200$ epochs, the weight has achieved $0.95$ on most edges of the normal cells, however, the \textit{none} operator is not considered in the final architecture. \textbf{Second}, almost all preserved operators are the \textit{skip-connect} (\textit{a.k.a.}, \textit{identity}) operator, a parameter-free operator that contributes little to feature learning -- and surprisingly, it occupies $30\%$ to $70\%$ of the weight remained by the \textit{none} operator. Such a network has very few trainable parameters, and thus usually reports unsatisfying performance at the re-training stage, in particular, lower than a randomly sampled network (please refer to the experiments in Section~\ref{experiments:CIFAR10:stabilization}).

Despite dramatically bad sub-networks are produced, the validation accuracy of the super-network keeps growing as the search process continues. In a typical run of the \textit{first-order} DARTS on CIFAR10, from $50$ to $200$ search epochs, the validation accuracy of the super-network is boosted from $88.82\%$ to $91.06\%$, while the re-training accuracy of the final architecture reduced from $97.00\%$ to $93.82\%$. This implies an \textbf{optimization gap} between the super-network and its sub-networks. Specifically, the search process aims to improve the validation accuracy of the super-network, but this does not necessarily result in high accuracy of the optimal sub-network determined by the architectural parameters\footnote{This opinion is different from that of~\cite{zela2020understanding}, which believed a super-network with higher validation accuracy must be better, and owed unsatisfying sub-network performance to the final discretization step of DARTS. Please refer to Appendix~\ref{maths:gap} for the detailed elaboration.}, $\boldsymbol{\alpha}$. A practical solution is to terminate the search process early~\cite{liang2019darts+}, however, despite its effectiveness, early termination makes the search result sensitive to the initialization ($\boldsymbol{\alpha}$ and $\boldsymbol{\omega}$), the hyper-parameters of search (\textit{e.g.}, learning rate), and the time of termination. Consequently, the stability of DARTS is inevitably weakened.

\subsection{Amending the Architectural Gradients}
\label{approach:mathematics}

We point out that the failure owes to inaccurate estimation of $\left.\nabla_{\boldsymbol{\alpha}}\mathcal{L}_\mathrm{val}\!\left(\boldsymbol{\omega}^\star\!\left(\boldsymbol{\alpha}\right),\boldsymbol{\alpha}\right)\right|_{\boldsymbol{\alpha}=\boldsymbol{\alpha}_t}$, the gradient with respect to $\boldsymbol{\alpha}$. Following the chain rule of gradients, this quantity equals to
\begin{eqnarray}
\label{eqn:chain_rule}
\nonumber
\left.\nabla_{\boldsymbol{\alpha}}\mathcal{L}_\mathrm{val}\!\left(\boldsymbol{\omega},\boldsymbol{\alpha}\right)\right|_{\boldsymbol{\omega}=\boldsymbol{\omega}^\star\!\left(\boldsymbol{\alpha}_t\right),\boldsymbol{\alpha}=\boldsymbol{\alpha}_t}+\left.\nabla_{\boldsymbol{\alpha}}\boldsymbol{\omega}^\star\!\left(\boldsymbol{\alpha}\right)\right|_{\boldsymbol{\alpha}=\boldsymbol{\alpha_t}}\cdot\\
\left.\left.\nabla_{\boldsymbol{\omega}}\mathcal{L}_\mathrm{val}\!\left(\boldsymbol{\omega},\boldsymbol{\alpha}\right)\right|_{\boldsymbol{\omega}=\boldsymbol{\omega}^\star\!\left(\boldsymbol{\alpha}_t\right),\boldsymbol{\alpha}=\boldsymbol{\alpha}_t}\right..
\end{eqnarray}
We denote ${\mathbf{g}_1}={\left.\nabla_{\boldsymbol{\alpha}}\mathcal{L}_\mathrm{val}\!\left(\boldsymbol{\omega},\boldsymbol{\alpha}\right)\right|_{\boldsymbol{\omega}=\boldsymbol{\omega}^\star\!\left(\boldsymbol{\alpha}_t\right),\boldsymbol{\alpha}=\boldsymbol{\alpha}_t}}$ and ${\mathbf{g}_2}={\left.\nabla_{\boldsymbol{\alpha}}\boldsymbol{\omega}^\star\!\left(\boldsymbol{\alpha}\right)\right|_{\boldsymbol{\alpha}=\boldsymbol{\alpha_t}}\cdot\left.\left.\nabla_{\boldsymbol{\omega}}\mathcal{L}_\mathrm{val}\!\left(\boldsymbol{\omega},\boldsymbol{\alpha}\right)\right|_{\boldsymbol{\omega}=\boldsymbol{\omega}^\star\!\left(\boldsymbol{\alpha}_t\right),\boldsymbol{\alpha}=\boldsymbol{\alpha}_t}\right.}$. $\mathbf{g}_1$ is easy to compute as is done in the \textit{first-order} DARTS, while $\mathbf{g}_2$, in particular $\left.\nabla_{\boldsymbol{\alpha}}\boldsymbol{\omega}^\star\!\left(\boldsymbol{\alpha}\right)\right|_{\boldsymbol{\alpha}=\boldsymbol{\alpha_t}}$, is not (see Appendix~\ref{maths:toy_example}). To estimate $\left.\nabla_{\boldsymbol{\alpha}}\boldsymbol{\omega}^\star\!\left(\boldsymbol{\alpha}\right)\right|_{\boldsymbol{\alpha}=\boldsymbol{\alpha_t}}$, we note that $\boldsymbol{\omega}^\star\!\left(\boldsymbol{\alpha}\right)$ has arrived at the optimality on the training set, hence ${\left.\nabla_{\boldsymbol{\omega}}\mathcal{L}_\mathrm{train}\!\left(\boldsymbol{\omega},\boldsymbol{\alpha}\right)\right|_{\boldsymbol{\omega}=\boldsymbol{\omega}^\star\!\left(\boldsymbol{\alpha}^\dagger\right),\boldsymbol{\alpha}=\boldsymbol{\alpha}^\dagger}}\equiv{\mathbf{0}}$ holds for any $\boldsymbol{\alpha}^\dagger$. Differentiating with respect to any $\boldsymbol{\alpha}^\dagger$ on both sides, we have ${\nabla_{\boldsymbol{\alpha}^\dagger}\left(\left.\nabla_{\boldsymbol{\omega}}\mathcal{L}_\mathrm{train}\!\left(\boldsymbol{\omega},\boldsymbol{\alpha}\right)\right|_{\boldsymbol{\omega}=\boldsymbol{\omega}^\star\!\left(\boldsymbol{\alpha}^\dagger\right),\boldsymbol{\alpha}=\boldsymbol{\alpha}^\dagger}\right)}\equiv{\mathbf{0}}$.
When $\boldsymbol{\alpha}^\dagger=\boldsymbol{\alpha_t}$, it becomes:
\begin{equation}
{\left.\nabla_{\boldsymbol{\alpha^\dagger}}\left(\left.\nabla_{\boldsymbol{\omega}}\mathcal{L}_\mathrm{train}\!\left(\boldsymbol{\omega},\boldsymbol{\alpha}\right)\right|_{\boldsymbol{\omega}=\boldsymbol{\omega}^\star\!\left(\boldsymbol{\alpha}^\dagger\right),\boldsymbol{\alpha}=\boldsymbol{\alpha}^\dagger}\right)\right|_{\boldsymbol{\alpha^\dagger}=\boldsymbol{\alpha}_t}}=\mathbf{0}.
\end{equation}
Again, applying the chain rule to the left-hand side gives:
\begin{eqnarray}
\nonumber
{\left.\nabla_{\boldsymbol{\alpha},\boldsymbol{\omega}}^2\mathcal{L}_\mathrm{train}\!\left(\boldsymbol{\omega},\boldsymbol{\alpha}\right)\right|_{\boldsymbol{\omega}=\boldsymbol{\omega}^\star\!\left(\boldsymbol{\alpha}_t\right),\boldsymbol{\alpha}=\boldsymbol{\alpha}_t}+\left.\nabla_{\boldsymbol{\alpha}}\boldsymbol{\omega}^\star\!\left(\boldsymbol{\alpha}\right)\right|_{\boldsymbol{\alpha}=\boldsymbol{\alpha}_t}}\cdot\\
{\left.\nabla_{\boldsymbol{\omega}}^2\mathcal{L}_\mathrm{train}\!\left(\boldsymbol{\omega},\boldsymbol{\alpha}\right)\right|_{\boldsymbol{\omega}=\boldsymbol{\omega}^\star\!\left(\boldsymbol{\alpha}_t\right),\boldsymbol{\alpha}=\boldsymbol{\alpha}_t}}={\mathbf{0}},
\end{eqnarray}
where we use the notation of ${\nabla_{\boldsymbol{\alpha},\boldsymbol{\omega}}^2\!\left(\cdot\right)}\doteq{\nabla_{\boldsymbol{\alpha}}\!\left(\nabla_{\boldsymbol{\omega}}\!\left(\cdot\right)\right)}$ throughout the remaining part of this paper. We denote ${\mathbf{H}}\doteq{\left.\nabla_{\boldsymbol{\omega}}^2\mathcal{L}_\mathrm{train}\!\left(\boldsymbol{\omega},\boldsymbol{\alpha}\right)\right|_{\boldsymbol{\omega}=\boldsymbol{\omega}^\star\!\left(\boldsymbol{\alpha}_t\right),\boldsymbol{\alpha}=\boldsymbol{\alpha}_t}}$, the Hesse matrix corresponding to the optimum, $\boldsymbol{\omega}^\star\!\left(\boldsymbol{\alpha}_t\right)$. $\mathbf{H}$ is symmetric and positive-definite, and thus invertible, which gives that ${\left.\nabla_{\boldsymbol{\alpha}}\boldsymbol{\omega}^\star\!\left(\boldsymbol{\alpha}\right)\right|_{\boldsymbol{\alpha}=\boldsymbol{\alpha}_t}}=-\left.\nabla_{\boldsymbol{\alpha},\boldsymbol{\omega}}^2\mathcal{L}_\mathrm{train}\!\left(\boldsymbol{\omega},\boldsymbol{\alpha}\right)\right|_{\boldsymbol{\omega}=\boldsymbol{\omega}^\star\!\left(\boldsymbol{\alpha_t}\right),\boldsymbol{\alpha}=\boldsymbol{\alpha}_t}\cdot{\mathbf{H}^{-1}}$. Substituting it into $\mathbf{g}_2$ gives
\begin{eqnarray}
\nonumber
{\mathbf{g}_2}={-\left.\left.\nabla_{\boldsymbol{\alpha},\boldsymbol{\omega}}^2\mathcal{L}_\mathrm{train}\!\left(\boldsymbol{\omega},\boldsymbol{\alpha}\right)\right|_{\boldsymbol{\omega}=\boldsymbol{\omega}^\star\!\left(\boldsymbol{\alpha}_t\right),\boldsymbol{\alpha}=\boldsymbol{\alpha}_t}\right.\cdot\mathbf{H}^{-1}\cdot}\\
\label{eqn:gradient_computation}
{\left.\nabla_{\boldsymbol{\omega}}\mathcal{L}_\mathrm{val}\!\left(\boldsymbol{\omega},\boldsymbol{\alpha}\right)\right|_{\boldsymbol{\omega}=\boldsymbol{\omega}^\star\!\left(\boldsymbol{\alpha}_t\right),\boldsymbol{\alpha}=\boldsymbol{\alpha}_t}}.
\end{eqnarray}
Note that no approximation has been made till now.

To compute $\mathbf{g}_2$, the main difficulty lies in $\mathbf{H}^{-1}$ which, due to the high dimensionality of $\mathbf{H}$ (over one million in DARTS), is computationally intractable. We directly replace $\mathbf{H}^{-1}$ with $\mathbf{H}$, which leads to an approximated term:
\begin{eqnarray}
\nonumber
{\mathbf{g}_2'}={-\eta\cdot\left.\left.\nabla_{\boldsymbol{\alpha},\boldsymbol{\omega}}^2\mathcal{L}_\mathrm{train}\!\left(\boldsymbol{\omega},\boldsymbol{\alpha}\right)\right|_{\boldsymbol{\omega}=\boldsymbol{\omega}^\star\!\left(\boldsymbol{\alpha}_t\right),\boldsymbol{\alpha}=\boldsymbol{\alpha}_t}\right.\cdot\mathbf{H}\cdot}\\
\label{eqn:gradient_approximation}
{\left.\nabla_{\boldsymbol{\omega}}\mathcal{L}_\mathrm{val}\!\left(\boldsymbol{\omega},\boldsymbol{\alpha}\right)\right|_{\boldsymbol{\omega}=\boldsymbol{\omega}^\star\!\left(\boldsymbol{\alpha}_t\right),\boldsymbol{\alpha}=\boldsymbol{\alpha}_t}},
\end{eqnarray}
where ${\eta}>{0}$ is named the \textbf{amending coefficient}, the only hyper-parameter of our approach, and its effect will be discussed in the experimental section. A nice property of $\mathbf{g}_2'$ is that the angle between $\mathbf{g}_2$ and $\mathbf{g}_2'$ does not exceed $90^\circ$(\textit{i.e.}, ${\left\langle\mathbf{g}_2',\mathbf{g}_2\right\rangle}\geqslant{0}$, as proved in Appendix~\ref{maths:inner_product}).

As the final step, we compute $\mathbf{g}_1$ and $\mathbf{g}_2'$ following the \textit{second-order} DARTS (see Appendix~\ref{maths:computation}). Overall, the computation of Eqn~\eqref{eqn:gradient_approximation} requires similar computational costs of the \textit{second-order} DARTS. On an NVIDIA Tesla-V100 GPU, each search epoch requires around $0.02$ GPU-days on the standard $8$-cell search space on CIFAR10.

\subsection{Why Is Our Approach Better Than DARTS?}
\label{approach:comparison}

DARTS fails mostly due to inaccurate estimation of $\mathbf{g}_2$. The \textit{first-order} DARTS directly discarded this term, \textit{i.e.}, setting ${\mathbf{g}_2^\mathrm{1st}}={\mathbf{0}}$, and the \textit{second-order} DARTS used $\mathbf{I}$ (the identity matrix) to replace $\mathbf{H}^{-1}$ in Eqn~\eqref{eqn:gradient_computation}. Let us denote the approximated term as $\mathbf{g}_2^\mathrm{2nd}$, then the property that ${\left\langle\mathbf{g}_2^\mathrm{2nd},\mathbf{g}_2\right\rangle}\geqslant{0}$ does not hold. Consequently, there can be a significant gap between the true and estimated values of $\left.\nabla_{\boldsymbol{\alpha}}\mathcal{L}_\mathrm{val}\!\left(\boldsymbol{\omega}^\star\!\left(\boldsymbol{\alpha}\right),\boldsymbol{\alpha}\right)\right|_{\boldsymbol{\alpha}=\boldsymbol{\alpha}_t}$. Such inaccuracy accumulates with every update on $\boldsymbol{\alpha}$, and eventually causes $\boldsymbol{\alpha}$ to converge to the degenerated solutions. Although all (up to $100$) our trials of the original DARTS search converge to architectures with all \textit{skip-connect} operators, we do not find a theoretical explanation why this is the only ending point, which is left as an open problem for future research.

On the contrary, with an amended approximation, $\mathbf{g}_2'$, our approach can survive after a sufficiently long search process. The longest search process in our experiments has $500$ epochs, after which the architecture remains mostly the same as that after $100$ epochs. In addition, the final architecture seems converged, \textit{i.e.}, will not change even with more search epochs, as (i) the \textit{none} operator does not dominate any edge; and (ii) the weight of the dominating operator in each edge is still gradually increasing.

The rationality of our approach is also verified by the validation accuracy of the search stage. In a typical search process on CIFAR10, the \textit{first-order} DARTS, by directly discarding $\mathbf{g}_2$, reports an average validation accuracy of $90.5\%$. The \textit{second-order} DARTS adds $\mathbf{g}_2^\mathrm{2nd}$ but reports a reduced validation accuracy. Our approach, by adding $\mathbf{g}_2'$, achieves an improved validation accuracy of $91.5\%$, higher than both versions of DARTS. This indicates that our approach indeed provides more accurate approximation so that the super-network is better optimized.

\subsection{Hyper-Parameter Consistency}
\label{approach:consistency}

Our goal is to bridge the optimization gap between the search and re-training phases. Besides amending the architectural gradients to avoid `over-fitting' the super-network, another important factor is to make the hyper-parameters used in search and re-training consistent. In the contexts of DARTS, examples include using different depths (\textit{e.g.}, DARTS used $8$ cells in search and $20$ cells in re-training) and widths (\textit{e.g.}, DARTS used a basic channel number of $16$ in search and $36$ in re-training), as well as using different training strategies (\textit{e.g.}, during re-training, a few regularization techniques including Cutout~\cite{devries2017improved}, Dropout~\cite{srivastava2014dropout} and auxiliary loss~\cite{szegedy2015going} were used, but none of them appeared in search). More importantly, the final step of search (removing $6$ out of $14$ edges from the structure) can cause another significant gap. In Section~\ref{experiments:CIFAR10:stabilization}, we will discuss some practical ways to bridge these gaps towards higher stability. For more analysis on this point, please refer to Appendix~\ref{details:consistency}.

\subsection{Discussions and Relationship to Prior Work}
\label{approach:discussions}

A few prior differentiable search approaches noticed the issue of instability, but they chose to solve it in different manners. For example, P-DARTS~\cite{chen2019progressive} fixed the number of preserved \textit{skip-connect} operators, PC-DARTS~\cite{xu2020pc} used edge normalization to eliminate the \textit{none} operator, while XNAS~\cite{nayman2019xnas} and DARTS+~\cite{liang2019darts+} introduced a few human expertise to stabilize search. However, we point out that (i) either P-DARTS or PC-DARTS, with carefully designed methods or tricks, can also fail in a long enough search process (more than $200$ epochs); and that (ii) XNAS and DARTS+, by adding human expertise, somewhat violated the design principle of AutoML, in which one is expected to avoid introducing too many hand-designed rules.

A recent work~\cite{zela2020understanding} also tried to robustify DARTS in a simplified search space, and using various optimization strategies, including $\ell_2$-regularization, adding Dropout and using early termination. We point out that these methods are still sensitive to initialization and hyper-parameters, while our approach provides a mathematical explanation and enjoys a better convergence property (see Section~\ref{experiments:CIFAR10:stabilization} and Appendix~\ref{extra:understanding}).

Another line of NAS, besides differentiable methods, is to use either reinforcement learning or an evolutionary algorithm as a controller of heuristic search and train each sampled network to get some kind of rewards, \textit{e.g.}, validation accuracy. In the viewpoint of optimization, this pipeline mainly differs from the differentiable one in that optimizing $\boldsymbol{\alpha}$ is decoupled from optimizing $\boldsymbol{\omega}$, so that it does not require $\boldsymbol{\omega}$ to arrive at $\boldsymbol{\omega}^\ast$, but only need a reasonable approximation of $\boldsymbol{\omega}^\ast$ to predict model performance -- this is an important reason that such algorithms often produce stable results. Our approach sheds light on introducing a similar property, \textit{i.e.}, robustness to approximated $\boldsymbol{\omega}^\star$, which helps in stabilizing differentiable search approaches.

\section{Experiments}
\label{experiments}

\subsection{Results on CIFAR10}
\label{experiments:CIFAR10}

The CIFAR10 dataset~\cite{krizhevsky2009learning} has $50\rm{,}000$ training and $10\rm{,}000$ testing images, equally distributed over $10$ classes. We mainly use this dataset to evaluate the stability of our approach, as well as analyze the impacts of different search options and parameters.

We search and re-train similarly as DARTS. During the search, all operators are assigned equal weights on each edge. The batch size is set to be $96$. An Adam optimizer is used to update architectural parameters, with a learning rate of $0.0003$, a weight decay of $0.001$ and a momentum of $\left(0.5,0.999\right)$. The number of epochs is to be discussed later. During re-training, the base channel number is increased to $36$. An SGD optimizer is used with an initial learning rate starting from $0.025$, decaying with cosine annealing, and arriving at $0$ after $600$ epochs. The weight decay is set to be $0.0003$, and the momentum is $0.9$.

\subsubsection{Impact of the Amending Coefficient}
\label{experiments:CIFAR10:coefficient}

\begin{figure}[!t]
\centering
\subfigure[${\eta}\leqslant{0.01}$, CIFAR10 Test Error: 6.18\%]{\includegraphics[height=2.0cm]{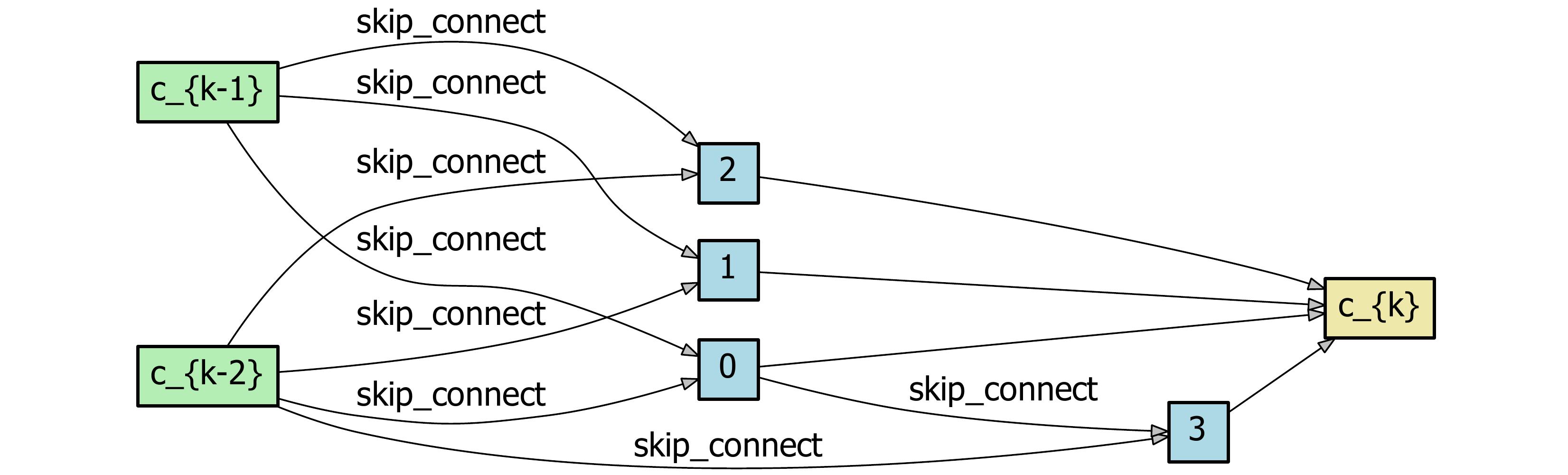}}
\subfigure[${\eta}={0.1}$, CIFAR10 Test Error: 3.08\%]{\includegraphics[height=2.3cm]{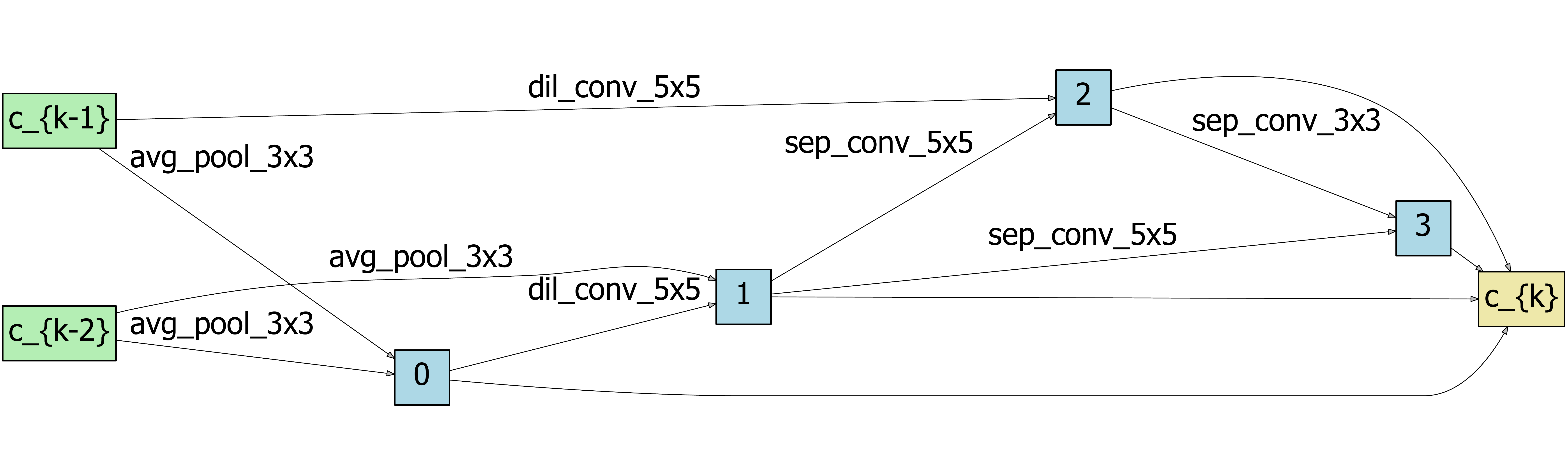}}
\subfigure[${\eta}\geqslant{1}$, CIFAR10 Test Error: 7.16\%]{\includegraphics[height=2.0cm]{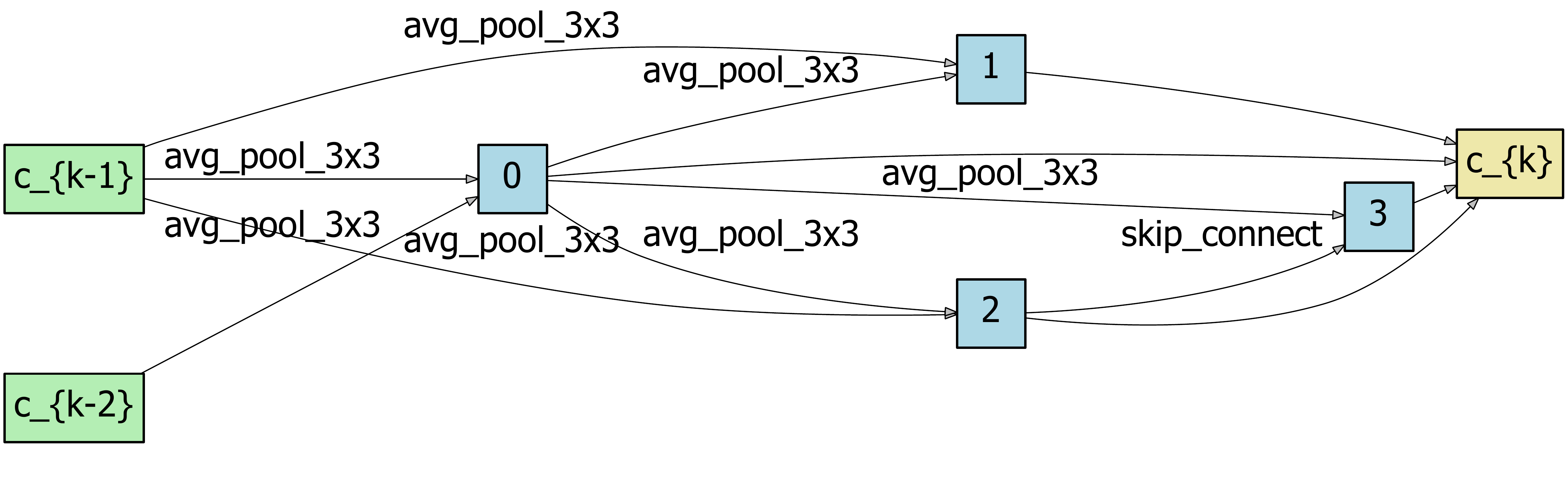}}
\caption{The normal cells (in the standard DARTS space) obtained by different amending coefficients.}
\label{fig:amending_coefficient}
\end{figure}

\begin{figure*}[!b]
\centering
\includegraphics[width=0.55\textwidth]{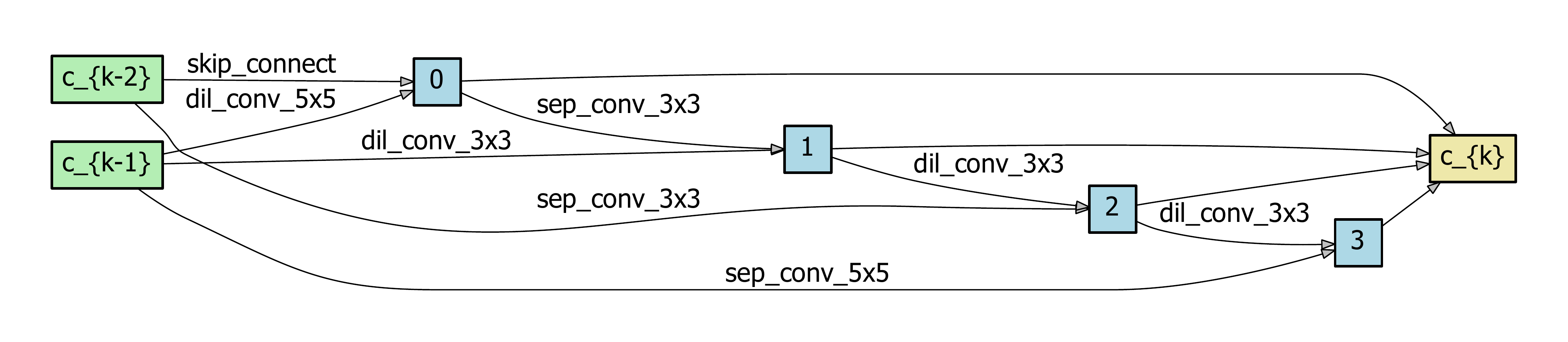}
\includegraphics[width=0.42\textwidth]{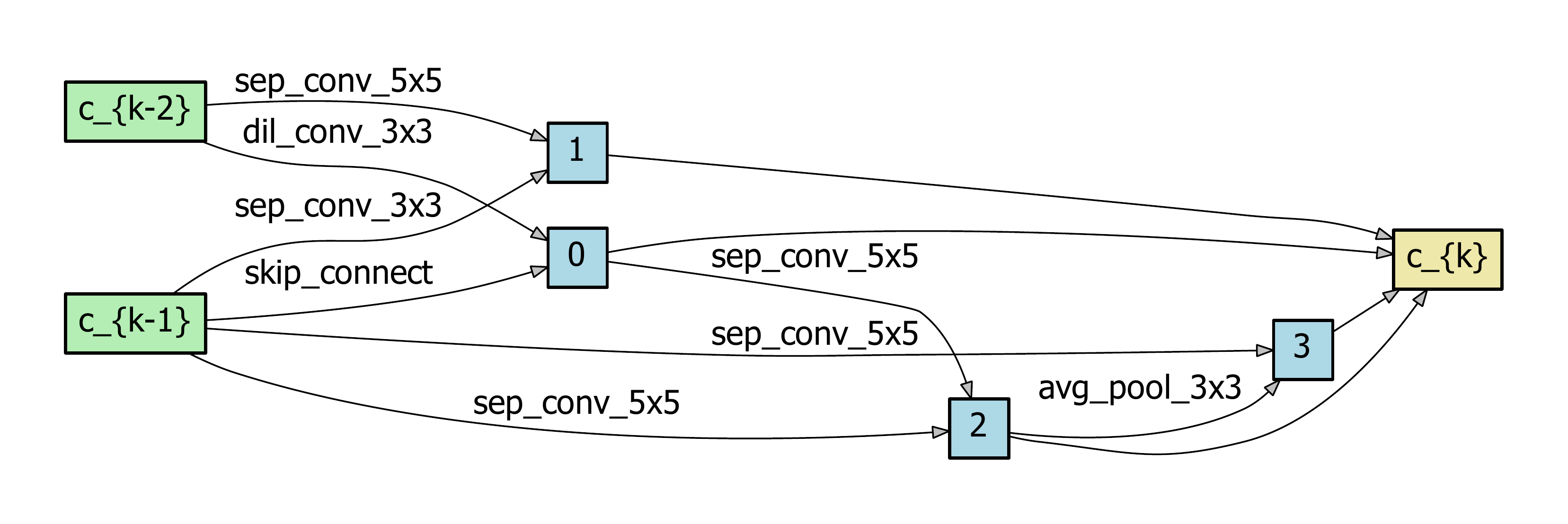}\\
\vspace{0.2cm}
\includegraphics[width=1.00\textwidth]{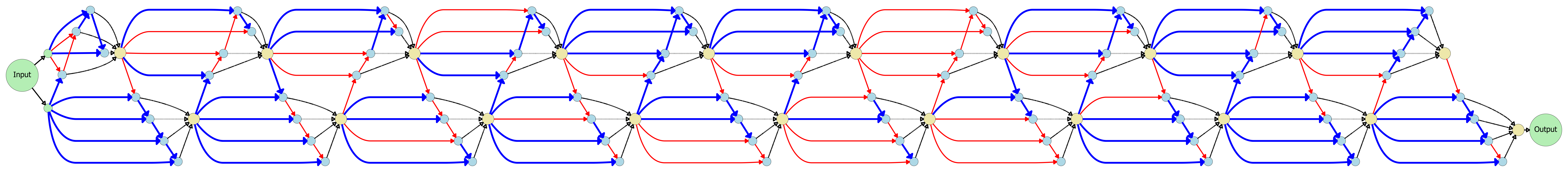}
\vspace{0.2cm}
\includegraphics[width=0.90\textwidth]{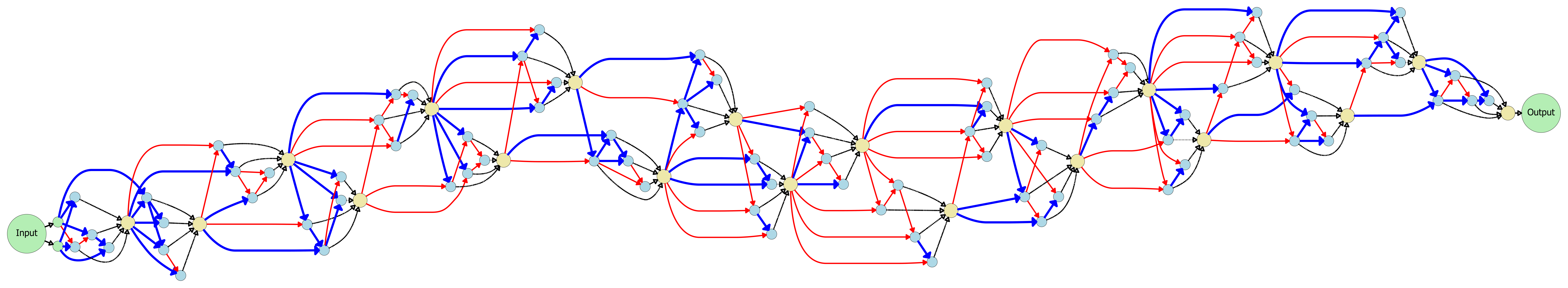}
\vspace{-0.4cm}
\caption{\textbf{Top}: the normal and reduction cells found in $\mathcal{S}_1$ with edge pruning, \textit{i.e.}, following the original DARTS to search on full ($14$) edges. \textbf{Middle} \& \textbf{Bottom}: the overall architecture found in $\mathcal{S}_2$ with fixed and searched edges, in which red thin, blue bold, and black dashed arrows indicate \textit{skip-connect}, \textit{sep-conv-3x3}, and \textit{channel-wise concatenation}, respectively. \textit{This figure is best viewed in color.}}
\label{fig:architectures_CIFAR10}
\end{figure*}

We first investigate how the amending coefficient, $\eta$, defined in Eqn~\eqref{eqn:gradient_approximation}, impacts architecture search. To arrive at convergence, we run the search stage for $500$ epochs. We evaluate different $\eta$ values from $0$ to $1$, and the architectures corresponding to small, medium and large amending coefficients are summarized in Figure~\ref{fig:amending_coefficient}.

We can see that after $500$ epochs, ${\eta}={0.1}$ produces a reasonable architecture that achieves an error rate of $3.08\%$ on CIFAR10. Actually, even with more search epochs, this architecture is not likely to change, as the preserved operator on each edge has a weight not smaller than $0.5$, and most of these weights are still growing gradually.

When $\eta$ is very small, \textit{e.g.}, ${\eta}={0.001}$ or ${\eta}={0.01}$, the change brought by this amending term to architecture search is negligible, and our approach shows almost the same behavior as the \textit{first-order} version of DARTS, \textit{i.e.}, ${\eta}={0}$. In addition, in such scenarios, although the search process eventually runs into an architecture with all \textit{skip-connect} operators, the number of epochs needed increases significantly, which verifies that the amending term indeed pulls architecture search away from degeneration.

On the other hand, if we use a sufficiently large $\eta$ value, \textit{e.g.}, ${\eta}={1}$, the amending term, $\mathbf{g}_2'$, can dominate optimization, so that the first term, \textit{i.e.}, the gradient of architectural parameters, has limited effects in updating $\boldsymbol{\alpha}$. Note that the amending term is closely related to network regularization, therefore, in the scenarios of a large $\eta$, the network significantly prefers \textit{avg-pool-3x3} to other operators, as average pooling can smooth feature maps and avoid over-fitting. However, pooling is also a parameter-free operator, so the performance of such architectures is also below satisfaction.

Following these analyses, we simply use ${\eta}={0.1}$ for all later experiments. We do not tune $\eta$ very carefully, though it is possible to determine $\eta$ automatically using a held-out validation set. Besides, we find that the best architecture barely changes after $100$ search epochs, so we fix the search length to be $100$ epochs to reduce computational costs.

\subsubsection{Towards Stabilized Search Performance}
\label{experiments:CIFAR10:stabilization}

Next, we investigate the stability of our approach by comparing it (with ${\eta}={0.1}$) to DARTS~\cite{liu2019darts}, P-DARTS~\cite{chen2019progressive}, and PC-DARTS~\cite{xu2020pc}. We run all the competitors for $200$ search epochs\footnote{For P-DARTS, we run each of its three search stages for $100$ epochs, and do not use the heuristic rule that preserves exactly two \textit{skip-connect} operators.} to guarantee convergence in the final architecture. Results are summarized in Table~\ref{tab:degeneration_CIFAR10}. One can see that all others, except PC-DARTS, produce lower accuracy than that of random search (some of them, DARTS and P-DARTS, even degenerate to all-\textit{skip-connect} architectures), but our approach survives, indicating the amended approximation effectively boosts search robustness. Moreover, we verify the search stability by claiming a $0.58\%$ advantage over random search.

\begin{table}[!t]
\begin{center}
\small
\begin{tabular}{lccc}
\hline
\textbf{\multirow{2}{*}{Architecture}} & \textbf{Test Err.} & \textbf{Params} & \textbf{\#P} \\
\cmidrule(lr){2-4}
 & \textbf{ (\%)} & \textbf{(M)} & \\
\hline
Random Search$^\dagger$          & 3.29 & 3.2 & - \\
\hline
DARTS (\textit{first-order})     & 6.18 & 1.4 & 0 \\
DARTS (\textit{second-order})    & 5.15 & 1.5 & 0 \\
P-DARTS                          & 5.38 & 1.5 & 0 \\
PC-DARTS                         & 3.15 & 2.4 & 3 \\
\hline
Our Approach                     & 2.71 & 3.3 & 7 \\
\quad \textit{w/o amending term} & 3.15 & 3.9 & 6 \\
\quad \textit{w/o consistency}   & 3.08 & 3.3 & 5 \\
\hline
\end{tabular}
\end{center}
\vspace{-0.4cm}
\caption{The performance of DARTS, P-DARTS, PC-DARTS and our approach after $200$ search epochs on CIFAR10. `\#P' means the number of parametric operators (excluding \textit{skip-connect}, \textit{avg-pool-3x3}, \textit{max-pool-3x3}) in the final normal cell. $^\dagger$: we borrow the results from DARTS~\cite{liu2019darts}.}
\label{tab:degeneration_CIFAR10}
\end{table}

\begin{table*}[!b]
\begin{center}
\begin{tabular}{lccccc}
\hline
\textbf{\multirow{2}{*}{Architecture}} & \textbf{Test Err.} & \textbf{Params} & \textbf{Search Cost} & \textbf{\multirow{2}{*}{Search Method}} \\
\cmidrule(lr){2-4}
&                            \textbf{ (\%)} & \textbf{(M)} & \textbf{(GPU-days)} &\\
\hline
DenseNet-BC~\cite{huang2017densely}                         & 3.46          & 25.6 & -           & manual         \\
\hline
ENAS~\cite{pham2018efficient} w/ Cutout                     & 2.89          & 4.6  & 0.5         & RL             \\
NASNet-A~\cite{zoph2018learning} w/ Cutout                  & 2.65          & 3.3  & 1800        & RL             \\
NAONet-WS~\cite{luo2018neural}                              & 3.53          & 3.1  & 0.4         & NAO            \\
PNAS~\cite{liu2018progressive}                              & 3.41$\pm$0.09 & 3.2  & 225         & SMBO           \\
Hireachical Evolution~\cite{liu2018hierarchical}            & 3.75$\pm$0.12 & 15.7 & 300         & evolution      \\
AmoebaNet-B~\cite{real2019regularized} w/ Cutout            & 2.55$\pm$0.05 & 2.8  & 3150        & evolution      \\
\hline
SNAS (moderate)~\cite{xie2018snas} w/ Cutout                & 2.85$\pm$0.02 & 2.8  & 1.5         & gradient-based \\
ProxylessNAS~\cite{cai2019proxylessnas} w/ Cutout           & 2.08          & -    & 4.0         & gradient-based \\
DARTS (\textit{first-order})~\cite{liu2019darts} w/ Cutout  & 3.00$\pm$0.14 & 3.3  & 0.4         & gradient-based \\
DARTS (\textit{second-order})~\cite{liu2019darts} w/ Cutout & 2.76$\pm$0.09 & 3.3  & 1.0         & gradient-based \\
P-DARTS~\cite{chen2019progressive} w/ Cutout                & 2.50          & 3.4  & 0.3         & gradient-based \\
BayesNAS~\cite{zhou2019bayesnas} w/ Cutout                  & 2.81$\pm$0.04 & 3.4  & 0.2         & gradient-based \\
PC-DARTS~\cite{xu2020pc} w/ Cutout                          & 2.57$\pm$0.07 & 3.6  & 0.1         & gradient-based \\
\hline
Amended-DARTS, $\mathcal{S}_1$, pruned edges, w/ Cutout     & 2.71$\pm$0.09 & 3.3  & 1.7         & gradient-based \\
Amended-DARTS, $\mathcal{S}_2$, fixed edges, w/ Cutout      & 2.60$\pm$0.15 & 3.6  & 1.1         & gradient-based \\
Amended-DARTS, $\mathcal{S}_2$, searched edges, w/ Cutout   & 2.63$\pm$0.09 & 3.0  & 3.1$^\ast$  & gradient-based \\
\hline
\end{tabular}
\end{center}
\vspace{-0.4cm}
\caption{Comparison with state-of-the-art network architectures on CIFAR10. $^\ast$: Edge search, the (optional) first step of our approach, is very costly, \textit{e.g.}, occupying $2.1$ out of $3.1$ GPU-days. For a detailed analysis of search cost, please refer to Appendix~\ref{details:costs}.}
\label{tab:comparison_CIFAR10}
\end{table*}

Table~\ref{tab:degeneration_CIFAR10} also provides an ablation study by switching off the amending term or the consistency of training hyper-parameters (\textit{i.e.}, adding Cutout, Dropout, and the auxiliary loss tower to the search stage with the same parameters, \textit{e.g.}, the Dropout ratio, as they are used in re-training). Without any one of them, the error rate significantly increases to more than $3\%$, still better than random search but the advantage becomes much weaker. These results verify our motivation, \textit{i.e.}, shrinking the optimization gap from any aspects can lead to better search performance. The architectures with and without unified hyper-parameters are shown in Figure~\ref{fig:architectures_CIFAR10} (top) and Figure~\ref{fig:amending_coefficient} (middle), respectively.

We also perform experiments in the simplified search space defined by~\cite{zela2020understanding}. Without bells and whistles, we obtain an error rate of $2.55\%$ which is significantly better than random search ($3.05\%$) reported in the paper. Please refer to Appendix~\ref{extra:understanding} for more results under this setting.

\subsubsection{Exploring More Complex Search Spaces}
\label{experiments:CIFAR10:complexity}

Driven by the benefit of shrinking the optimization gap, we further apply two modifications. \textbf{First}, to avoid edge removal, we partition the search stage into two sub-stages: the former chooses $8$ active edges from the $14$ candidates, and the latter, restarting from scratch, determines the operators on each preserved edge. Technical details are provided in Appendix~\ref{details:edge_selection}. Another option that can save computational costs is to fix the edges in each cell, \textit{e.g.}, each node $i$ is connected to node $i-1$ and the least indexed node (denoted by $c_{k-2}$ in most conventions). Note that our approach also works well with all $14$ edges preserved, but we have used $8$ edges to be computationally fair to DARTS. \textbf{Second}, we use the same width (\textit{i.e.}, the number of basic channels, $36$) and depth (\textit{i.e.}, the number of cells, $20$) in both search and re-training. The modification on depth reminds us of the \textit{depth gap}~\cite{chen2019progressive} between search and re-training (the network has $8$ cells in search, but $20$ cells in re-training). Instead of using a progressive search method, we directly search in an augmented space (see the next paragraph), thanks to the improved stability of our approach.

We denote the original search space used in DARTS as $\mathcal{S}_1$, which has six normal cells and two reduction cells, and all cells of the same type share the architectural parameters. This standard space contains $1.1\times10^{18}$ distinct architectures. We also explore a more complex search space, denoted by $\mathcal{S}_2$, in which we relax the constraint of sharing architectural parameters, meanwhile the number of cells increases from $8$ to $20$, \textit{i.e.}, the same as in the re-training stage. Here, since the GPU memory is limited, we cannot search with all seven operators, so we only choose two, namely \textit{skip-connect} and \textit{sep-conv-3x3}, which have very different properties. This setting allows a total of $1.9\times10^{93}$ architectures to appear which significantly surpasses the capacity of most existing cell-based search spaces.

The searched results in $\mathcal{S}_1$ and $\mathcal{S}_2$ with fixed or searched edges are shown in Figure~\ref{fig:architectures_CIFAR10}, and their performance summarized in Table~\ref{tab:comparison_CIFAR10}. By directly searching in the target space, $\mathcal{S}_2$, the error is further reduced from $2.71\%$ to $2.60\%$ and $2.63\%$ with fixed and searched edges, respectively. The improvement seems small on CIFAR10, but when we transfer these architectures to ImageNet, the corresponding advantages become more significant ($0.4\%$, see Table~\ref{tab:comparison_ILSVRC2012}).

We also execute DARTS (with early termination, otherwise it fails dramatically) and random search on $\mathcal{S}_2$ with the fixed-edge setting, and they report $0.25\%$ and $0.29\%$ deficits compared to our approach (DARTS is slightly better than random search). When we transfer these architectures to ImageNet, the deficits become more significant ($1.7\%$ and $0.8\%$, respectively, and DARTS performs even worse). This provides a side evidence to randomly-wired search~\cite{xie2019exploring}, advocating for the importance of designing stabilized approach on large search spaces.

\begin{table*}[!t]
\begin{center}
\begin{tabular}{lcccccc}
\hline
\textbf{\multirow{2}{*}{Architecture}} & \multicolumn{2}{c}{\textbf{Test Err. (\%)}} & \textbf{Params} & $\times+$ & \textbf{Search Cost} & \textbf{\multirow{2}{*}{Search Method}} \\
\cmidrule(lr){2-3}
&                            \textbf{top-1} & \textbf{top-5} & \textbf{(M)} & \textbf{(M)} & \textbf{(GPU-days)} &\\
\hline
Inception-v1~\cite{szegedy2015going}                     & 30.2 & 10.1 & 6.6     & 1448 & -            & manual         \\
MobileNet~\cite{howard2017mobilenets}                    & 29.4 & 10.5 & 4.2     & 569  & -            & manual         \\
ShuffleNet 2$\times$ (v1)~\cite{zhang2018shufflenet}     & 26.4 & 10.2 & $\sim$5 & 524  & -            & manual         \\
ShuffleNet 2$\times$ (v2)~\cite{ma2018shufflenet}        & 25.1 & -    & $\sim$5 & 591  & -            & manual         \\
\hline
NASNet-A~\cite{zoph2018learning}                         & 26.0 & 8.4  & 5.3     & 564  & 1800         & RL             \\
MnasNet-92~\cite{tan2019mnasnet}                         & 25.2 & 8.0  & 4.4     & 388  & -            & RL             \\
PNAS~\cite{liu2018progressive}                           & 25.8 & 8.1  & 5.1     & 588  & 225          & SMBO           \\
AmoebaNet-C~\cite{real2019regularized}                   & 24.3 & 7.6  & 6.4     & 570  & 3150         & evolution      \\
\hline
SNAS (mild)~\cite{xie2018snas}                           & 27.3 & 9.2  & 4.3     & 522  & 1.5          & gradient-based \\
ProxylessNAS (GPU)$^\ddagger$~\cite{cai2019proxylessnas} & 24.9 & 7.5  & 7.1     & 465  & 8.3          & gradient-based \\
DARTS (\textit{second-order})~\cite{liu2019darts}        & 26.7 & 8.7  & 4.7     & 574  & 4.0          & gradient-based \\
BayesNAS~\cite{zhou2019bayesnas}                         & 26.5 & 8.9  & 3.9     & -    & 0.2          & gradient-based \\
P-DARTS (CIFAR10)~\cite{chen2019progressive}             & 24.4 & 7.4  & 4.9     & 557  & 0.3          & gradient-based \\
PC-DARTS~\cite{xu2020pc}$^\ddagger$                      & 24.2 & 7.3  & 5.3     & 597  & 3.8          & gradient-based \\
\hline
Amended-DARTS, $\mathcal{S}_1$, pruned edges             & 24.7 & 7.6  & 5.2     & 586  & 1.7          & gradient-based \\
Amended-DARTS, $\mathcal{S}_2$, fixed edges              & 24.3 & 7.4  & 5.5     & 590  & 1.1          & gradient-based \\
Amended-DARTS, $\mathcal{S}_2$, searched edges           & 24.3 & 7.3  & 5.2     & 596  & 3.1          & gradient-based \\
\hline
\end{tabular}
\end{center}
\vspace{-0.5cm}
\caption{Comparison with state-of-the-arts on ILSVRC2012, under the \textit{mobile setting}. $^\ddagger$: these architectures are searched on ImageNet.}
\vspace{-0.2cm}
\label{tab:comparison_ILSVRC2012}
\end{table*}

\subsubsection{Comparison to the State-of-the-Arts}
\label{experiments:CIFAR10:comparison}

Finally, we compare our approach with recent approaches, in particular, differentiable ones. Result are shown in Table~\ref{tab:comparison_CIFAR10}. Our approach produces competitive results among state-of-the-arts, although it does not seem to beat others. We note that existing approaches often used additional tricks, \textit{e.g.}, P-DARTS assumed a fixed number of \textit{skip-connect} operators, which shrinks the search space (so as to guarantee stability). More importantly, all these differentiable search approaches must be terminated in an early stage, which makes them less convincing as search has not arrived at convergence. These tricks somewhat violate the ideology of neural architecture search; in comparison, our approach, though not producing the best performance, promises a more theoretically convinced direction.

\subsection{Results on ImageNet}
\label{experiments:ImageNet}

We use ILSVRC2012~\cite{russakovsky2015imagenet}, the most commonly used subset of ImageNet~\cite{deng2009imagenet}, for experiments. It contains $1.3\mathrm{M}$ training images and $50\mathrm{K}$ testing images, which are almost evenly distributed over all $1\rm{,}000$ categories. We directly use the architectures searched on CIFAR10 and compute a proper number of basic channels, so that the FLOPs of each architecture does not exceed $600\mathrm{M}$, \textit{i.e.}, the mobile setting. During re-training, there are a total of $250$ epochs. We use an SGD optimizer with an initial learning rate of $0.5$ (decaying linearly after each epoch), a momentum of $0.9$ and a weight decay of $3\times10^{-5}$. On eight NVIDIA Tesla V100 GPUs, the entire re-training process takes around $3$ days.

The comparison of our approach to existing work is shown in Table~\ref{tab:comparison_ILSVRC2012}. In the augmented search space, $\mathcal{S}_2$, our approach reports a top-$1$ error rate of $24.3\%$ without either AutoAugment~\cite{cubuk2019autoaugment} or Squeeze-and-Excitation~\cite{hu2018squeeze}. This result, obtained after search convergence, is competitive among state-of-the-arts. In comparison, without the amending term, DARTS converges to a weird architecture in which some cells are mostly occupied by \textit{skip-connect} and some others by \textit{sep-conv-3x3}. This architecture reports a top-$1$ error of $26.0\%$, which is even inferior to random search ($25.1\%$). Last but not least, the deficit of $\mathcal{S}_1$, compared to $\mathcal{S}_2$, becomes more significant on ImageNet. This again verifies the usefulness of shrinking the optimization gap, in particular for challenging tasks.

\subsection{Results on Penn Treebank}
\label{experiments:PTB}

\begin{table*}[!t]
\begin{center}
\begin{tabular}{lcccccc}
\hline
\textbf{\multirow{2}{*}{Architecture}} & \multicolumn{2}{c}{\textbf{Perplexity}} & \textbf{Params} & \textbf{Search Cost} & \textbf{\multirow{2}{*}{Search Method}} \\
\cmidrule(lr){2-3}
&                            \textbf{val} & \textbf{test} & \textbf{(M)} & \textbf{(GPU-days)} & \\
\hline
NAS~\cite{zoph2017neural}                         & - & 64.0 & 25     & $\sim$10000         & RL             \\
ENAS~\cite{pham2018efficient}$^\dagger$                         & 60.8 & 58.6 & 24     & 0.5         & RL             \\
\hline
Random search baseline$^\dagger$              & 61.8 & 59.4 & 23     &  2.0        & random \\
DARTS (first order)~\cite{liu2019darts}              & 60.2 & 57.6 & 23     &  0.5        & gradient-based \\
DARTS (second order)~\cite{liu2019darts}              & 58.1 & 55.7 & 23     &  1.0        & gradient-based \\
GDAS~\cite{dong2019searching}              & 59.8 & 57.5 & 23     &  0.4        & gradient-based \\
NASP~\cite{yao2020efficient}              & 59.9 & 57.3 & 23     &  0.1        & gradient-based \\
R-DARTS~\cite{zela2020understanding}              & - & 57.6 & 23     &  -        & gradient-based \\
\hline
Amended-DARTS              & \textbf{57.1} & \textbf{54.8} & 23     & 1.0         & gradient-based \\
\hline
\end{tabular}
\end{center}
\vspace{-0.5cm}
\caption{Comparison with NAS approaches on Penn Treebank. $^\dagger$ We borrow these results from~\cite{liu2019darts}.}
\vspace{-0.2cm}
\label{tab:PTB}
\end{table*}

We also evaluate our approach on the Penn Treebank dataset, a popular language modeling task on which the recurrent cell connecting LSTM units are being searched~\cite{zoph2017neural,pham2018efficient,liu2019darts}. We follow the implementation of DARTS to build the search pipeline and make two modifications, namely, (i) amending the second-order term according to Eqn~\eqref{eqn:gradient_approximation}, (ii) unify the weight decay and variational Dropout ratio between search and evaluation. The search process continues till the architecture does not change for sufficiently long, and the evaluation stage is executed for $8\rm{,}000$ epochs (\textit{i.e.}, until convergence, following the released code of DARTS), on which the best snapshot on the validation set is transferred to the test set.

\begin{figure}[!t]
\centering
\includegraphics[width=0.45\textwidth]{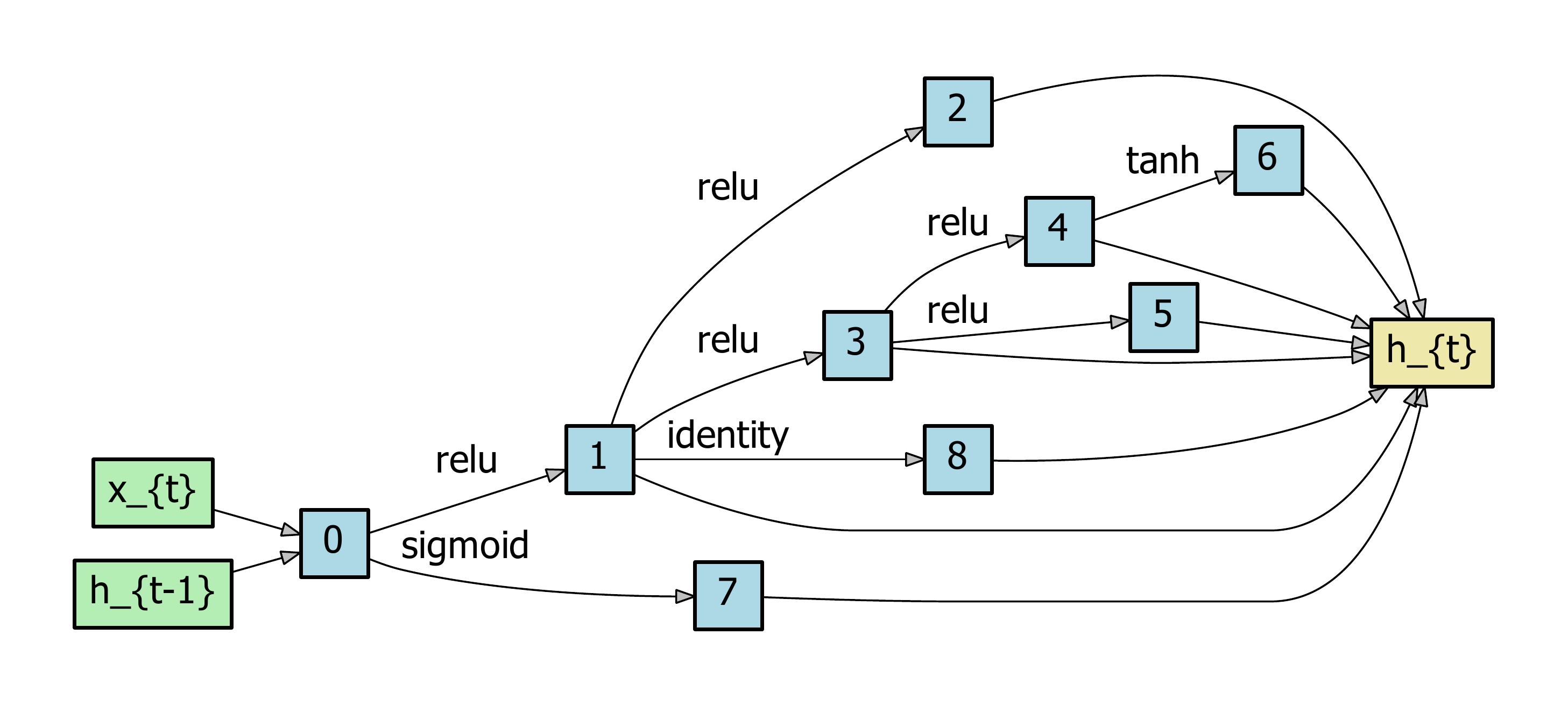}
\vspace{-0.4cm}
\caption{The recurrent cell searched by our approach on the Penn Treebank dataset. The random seed is $3$.}
\label{fig:PTB_recurrent_cell}
\end{figure}

Results are shown in Table~\ref{tab:PTB}, and the searched recurrent cell shown in Figure~\ref{fig:PTB_recurrent_cell}. Note that the DARTS paper reported $58.1$/$55.7$ validation/test perplexity (ppl), but our reproduction obtains $58.5$/$56.3$, slightly lower than the original implementation. Our approach with amended gradients reports $57.1$/$54.8$, showing a significant gain over the baseline. As far as we know, this is the \textbf{best} results ever reported in the DARTS space. Prior DARTS-based approaches mostly reported worse results than the original DARTS (see Table~\ref{tab:PTB}), or restricted to image-level operations (\textit{e.g.}, PC-DARTS~\cite{xu2020pc}), but our approach is a fundamental improvement over DARTS that boosts both computer vision and language modeling tasks. More importantly, we emphasize that after gradient fixation, the search process becomes more robust: we have achieved similar performance, in terms of ppl, using a few different random seeds, while the original DARTS seems quite sensitive to the random seed.

\section{Conclusions}
\label{conclusions}

In this paper, we present an effective approach for stabilizing DARTS, the state-of-the-art differentiable search method. Our motivation comes from that DARTS-based approaches mostly converge to all-\textit{skip-connect} architectures when they are executed for a sufficient number of epochs. We analyze this weird phenomenon mathematically and find the reason to be in the dramatic inaccuracy in gradient computation of the architectural parameters. With an alternative approximation based on the optimality of the network parameters, for the first time we can prove that the error of estimation is bounded, while previous work cannot. In standard image classification tasks, our approach shows improved stability, with which we are able to explore much larger search spaces and obtain better performance.

Our research sheds light on NAS research in several aspects. \textbf{First}, we reveal the importance of proper approximation in differentiable architecture search. \textbf{Second}, by fixing the error, we provide a platform for fairly comparing the ability of NAS itself rather than designing tricks. \textbf{Third}, thanks to improved stability, our algorithm can explore larger search spaces, which we believe is the future trend of NAS.

\bibliography{example_paper}
\bibliographystyle{icml2020}

\clearpage

\appendix

\onecolumn

\icmltitle{\textit{Appendix:} Stabilizing DARTS\\with Amended Gradient Estimation on Architectural Parameters}

\section{Mathematical Proofs and Analyses}
\label{maths}

In this section, we provide some details to complement the theoretical part of the main article.

Recall that the main goal of this paper is to compute $\left.\nabla_{\boldsymbol{\alpha}}\mathcal{L}_\mathrm{val}\!\left(\boldsymbol{\omega}^\star\!\left(\boldsymbol{\alpha}\right),\boldsymbol{\alpha}\right)\right|_{\boldsymbol{\alpha}=\boldsymbol{\alpha}_t}$, the gradient with respect to the architectural parameters, $\boldsymbol{\alpha}$. It is composed into the sum of $\mathbf{g}_1+\mathbf{g}_2$, and we use $\mathbf{g}_2'$ to approximate $\mathbf{g}_2$. The form of $\mathbf{g}_1$, $\mathbf{g}_2$ and $\mathbf{g}_2'$ is:
\begin{eqnarray}
{\mathbf{g}_1}&=&{\left.\nabla_{\boldsymbol{\alpha}}\mathcal{L}_\mathrm{val}\!\left(\boldsymbol{\omega},\boldsymbol{\alpha}\right)\right|_{\boldsymbol{\omega}=\boldsymbol{\omega}^\star\!\left(\boldsymbol{\alpha}_t\right),\boldsymbol{\alpha}=\boldsymbol{\alpha}_t}}.\\
\nonumber
{\mathbf{g}_2}&=&{\left.\nabla_{\boldsymbol{\alpha}}\boldsymbol{\omega}^\star\!\left(\boldsymbol{\alpha}\right)\right|_{\boldsymbol{\alpha}=\boldsymbol{\alpha_t}}\cdot\left.\left.\nabla_{\boldsymbol{\omega}}\mathcal{L}_\mathrm{val}\!\left(\boldsymbol{\omega},\boldsymbol{\alpha}\right)\right|_{\boldsymbol{\omega}=\boldsymbol{\omega}^\star\!\left(\boldsymbol{\alpha}_t\right),\boldsymbol{\alpha}=\boldsymbol{\alpha}_t}\right.}\\
&=&{-\left.\left.\nabla_{\boldsymbol{\alpha},\boldsymbol{\omega}}^2\mathcal{L}_\mathrm{train}\!\left(\boldsymbol{\omega},\boldsymbol{\alpha}\right)\right|_{\boldsymbol{\omega}=\boldsymbol{\omega}^\star\!\left(\boldsymbol{\alpha}_t\right),\boldsymbol{\alpha}=\boldsymbol{\alpha}_t}\right.\cdot\mathbf{H}^{-1}\cdot}{\left.\nabla_{\boldsymbol{\omega}}\mathcal{L}_\mathrm{val}\!\left(\boldsymbol{\omega},\boldsymbol{\alpha}\right)\right|_{\boldsymbol{\omega}=\boldsymbol{\omega}^\star\!\left(\boldsymbol{\alpha}_t\right),\boldsymbol{\alpha}=\boldsymbol{\alpha}_t}}.\\
{\mathbf{g}_2'}&=&{-\eta\cdot\left.\left.\nabla_{\boldsymbol{\alpha},\boldsymbol{\omega}}^2\mathcal{L}_\mathrm{train}\!\left(\boldsymbol{\omega},\boldsymbol{\alpha}\right)\right|_{\boldsymbol{\omega}=\boldsymbol{\omega}^\star\!\left(\boldsymbol{\alpha}_t\right),\boldsymbol{\alpha}=\boldsymbol{\alpha}_t}\right.\cdot\mathbf{H}\cdot\left.\nabla_{\boldsymbol{\omega}}\mathcal{L}_\mathrm{val}\!\left(\boldsymbol{\omega},\boldsymbol{\alpha}\right)\right|_{\boldsymbol{\omega}=\boldsymbol{\omega}^\star\!\left(\boldsymbol{\alpha}_t\right),\boldsymbol{\alpha}=\boldsymbol{\alpha}_t}}.
\end{eqnarray}

\subsection{Our Opinions on the Optimization Gap}
\label{maths:gap}

\textit{This part corresponds to Section~\ref{approach:gap} in the main article.}

\iffalse
\begin{table}[!b]
\begin{center}
\small
\begin{tabular}{lccc}
\hline
\textbf{\multirow{2}{*}{Architecture}} & \multicolumn{2}{c}{\textbf{Validation/Test Err.(\%)}} & \multirow{2}{*}{Discretization}\\
\cmidrule{2-3} & In the search phase & In the re-training phase\\
\hline
Super-network, w/ the amending term  & 10.5 & 5.4 & $\times$     \\
Super-network, w/o the amending term & 12.8 & 7.4 & $\times$     \\
Sub-network, w/ the amending term    & -    & 6.2 & $\checkmark$ \\
Sub-network, w/o the amending term   & -    & 7.4 & $\checkmark$ \\
\hline
\end{tabular}
\end{center}
\vspace{-0.4cm}
\caption{The performance of super-network and sub-network on CIFAR10. We trained on 50,000 training images and test on 10,000 testing images in re-training phase, while hold out half of the training data as the validation data in search phase.}
\label{tab:discretization}
\end{table}
\fi

In~\cite{zela2020understanding}, the authors believed that a super-network with a higher validation accuracy always has a higher probability of generating strong sub-networks, and they owed the weird behavior of DARTS to the discretization step (preserving the operator with the highest weight on each edge and discarding others) after the differentiable search phase. Here, we provide a different opinion, detailed as follows:
\begin{enumerate}
\item \textbf{A high validation accuracy does not necessarily indicate a better super-network.} In training a weight-sharing super-network, \textit{e.g.}, in DARTS, the improvement of validation accuracy can be brought by two factors, namely, the super-network configuration (corresponding to $\boldsymbol{\alpha}$) becomes better or the network weights (corresponding to $\boldsymbol{\omega}$) are better optimized. We perform an intuitive experiment in which we fix the initialized $\boldsymbol{\alpha}$ and only optimize $\boldsymbol{\omega}$. After $50$ epochs, the validation accuracy of the super-network is boosted from $10\%$ (random guess) to $88.28\%$. While the performance seems competitive among a regular training process, the sampled sub-network is totally random.
\item \textbf{The advantage of our approach persists even without discretization-and-pruning.} From another perspective, we try to skip the discretization-and-pruning step at the end of the search stage and re-train the super-network (with $\boldsymbol{\alpha}$ fixed) directly. To reduce computational costs, we use a shallower super-network, and perform both search and re-training for $600$ epochs on CIFAR10 to guarantee convergence. Without the amending term, the validation error in the search stage and the testing error in the re-training stage are $12.8\%$ and $7.4\%$, respectively, and these numbers become $10.5\%$ and $5.4\%$ after the amending term is added. This indicates that amending the gradient estimation indeed improves the super-network, and the advantage persists even when discretization-and-pruning is not used.
\item \textbf{Inaccuracy of gradient estimation also contributes to poor performance.} As detailed in Section~\ref{maths:toy_example}, we construct a toy example that applies bi-level optimization to the loss function of ${\mathcal{L}\left(\boldsymbol{\omega},\boldsymbol{\alpha};\boldsymbol{x}\right)}={\left(\boldsymbol{\omega}\boldsymbol{x}-\boldsymbol{\alpha}\right)^2}$. We show that even when $\boldsymbol{\omega}^\star$ is achieved at every iteration, the gradient of $\boldsymbol{\alpha}$, $\left.\nabla_{\boldsymbol{\alpha}}\mathcal{L}_\mathrm{val}\!\left(\boldsymbol{\omega}^\star\!\left(\boldsymbol{\alpha}\right),\boldsymbol{\alpha}\right)\right|_{\boldsymbol{\alpha}=\boldsymbol{\alpha}_t}$, can be totally incorrect. When this happens in architecture search, the super-network can be pushed towards sub-optimal or even random architectures, while the validation accuracy can continue growing.
\end{enumerate}

Therefore, our opinion is that the `optimization gap' is brought by the inconsistency between search and re-training -- edge pruning after search is one aspect, and the inaccuracy of gradient estimation is another one which seems more important. We alleviate the pruning issue by first selecting (or fixing) a few edges and then determining the operator on each preserved edge. However, such a two-stage search process can be very unstable if the gradient estimation remains inaccurate. In other words, amending errors in gradient computation lays the foundation of hyper-parameter consistency.

% From the experiments above, we point out:
% \begin{enumerate}
% \item The validation accuracy increases mainly because the network parameters are trained as the epochs grow, and will decrease when network parameters are fully trained(see the left part of Figure~\ref{fig:toy example loss} for a simple example).
% \item The interaction between architectural parameters and network parameters makes different super-networks have different final validation accuracy, but the impact is always small compared to the impact of training network parameters(1\% compared to 90\%). So when we compare two super-networks, we must train each super-network from scratch with the same settings to ensure the super-network is trained fully and fairly.
% \item There is still a key underlying problem lead to the weird results of DARTS, the inaccurate estimate of gradient, named `optimization gap'. The gap exists without the discretization stage and could be enlarged after it in some cases.
% \item The amending term will help to find good super-networks and sub-networks with a proper amending coefficient.
% \end{enumerate}

\begin{figure*}[!t]
\centering
\subfigure[A search process without the amending term.]{\includegraphics[width=0.45\textwidth]{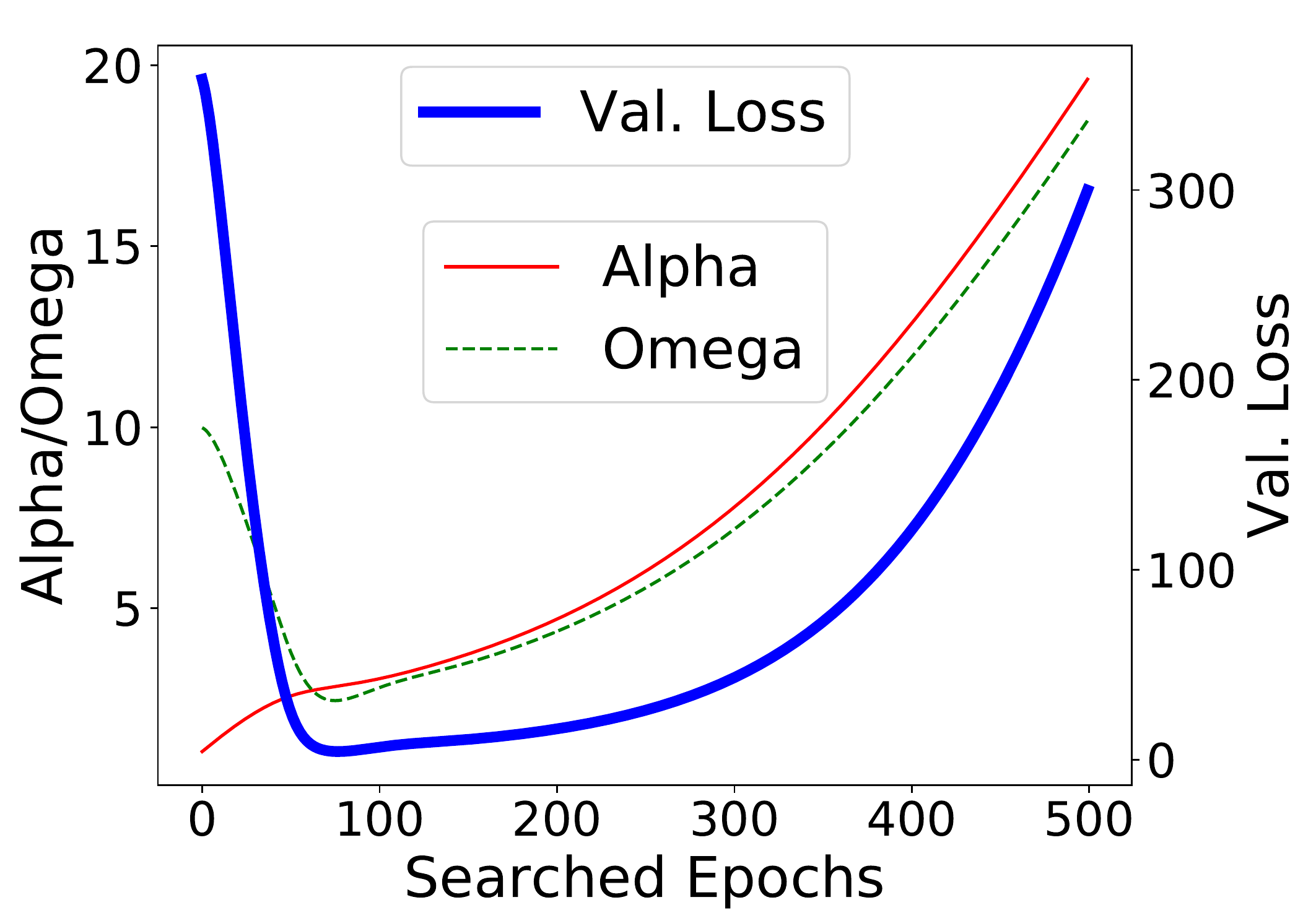}}
\subfigure[A search process with the amending term.]{\includegraphics[width=0.45\textwidth]{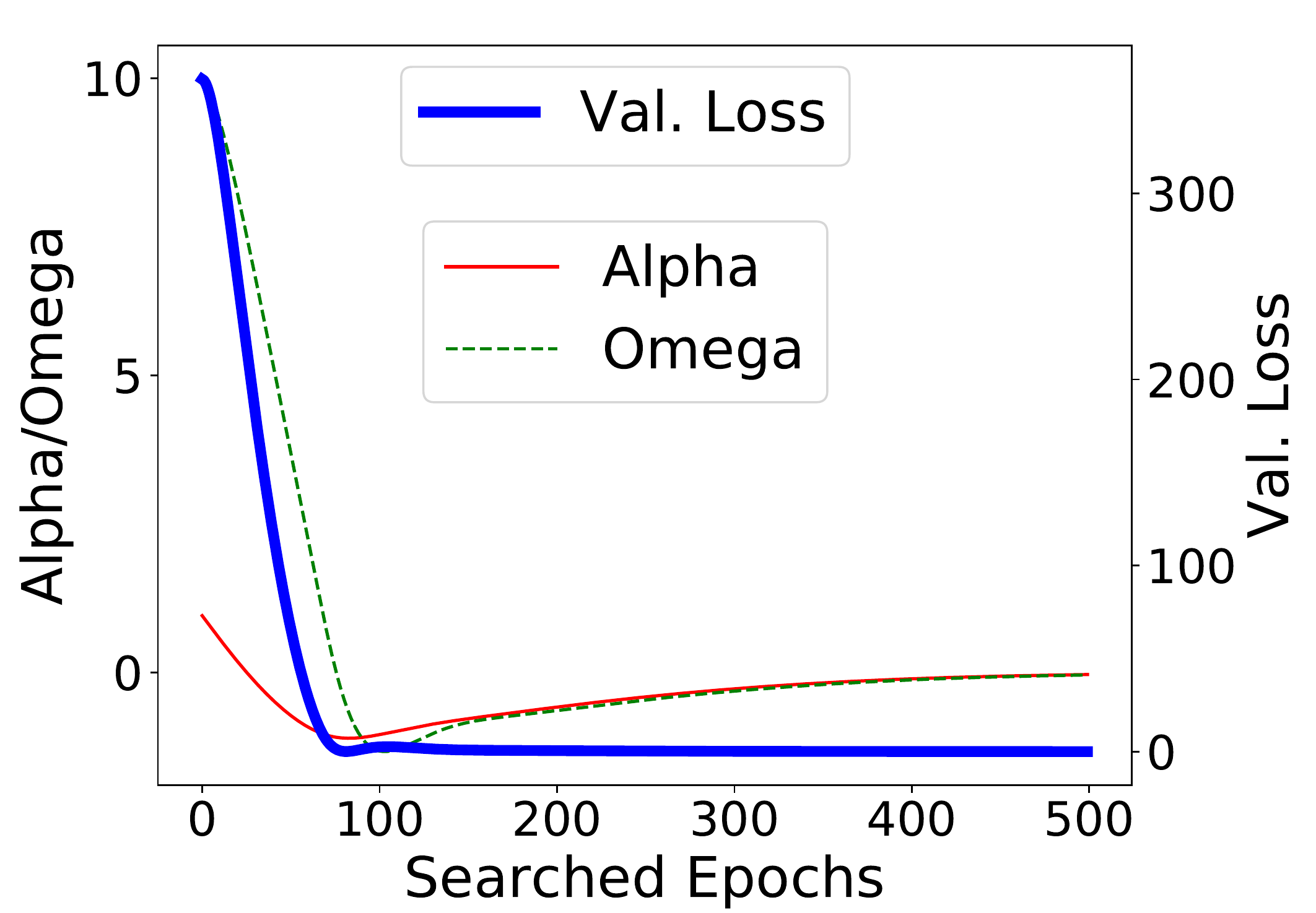}}\\
\vspace{-0.4cm}
\caption{Two search processes with a simple loss function $\mathcal{L}\left(\boldsymbol{\omega},\boldsymbol{\alpha};\mathbf{x}\right)=\left(\boldsymbol{\omega}\cdot\mathbf{x}-\boldsymbol{\alpha}\right)^2$, with all of $\boldsymbol{\omega}$, $\boldsymbol{\alpha}$ and $\mathbf{x}$ being $1$-dimensional scalars. Red, green and blue curves indicate the value of $\boldsymbol{\alpha}$, $\boldsymbol{\omega}$ and the validation loss, respectively.}
\label{fig:toy_example_curves}
\end{figure*}

\subsection{Importance of the Second-Order Gradient Term, $\mathbf{g}_2$}
\label{maths:toy_example}

\textit{This part corresponds to Section~\ref{approach:mathematics} in the main article.}

DARTS~\cite{liu2019darts} owed the inaccuracy in computing $\mathbf{g}_2$ to that $\boldsymbol{\omega}^\star\!\left(\boldsymbol{\alpha}\right)$ is difficult to arrive at (\textit{e.g.}, requiring a lot of computation), and believed that $\mathbf{g}_2$ goes to $\mathbf{0}$ when the optimum is achieved. We point out that this is not correct even in a very simple example of convex optimization, detailed as follows.

Let the loss function be ${\mathcal{L}\left(\boldsymbol{\omega},\boldsymbol{\alpha};\mathbf{x}\right)}={\left(\boldsymbol{\omega}\cdot\mathbf{x}-\boldsymbol{\alpha}\right)^2}$. Then, the only difference between ${\mathcal{L}_\mathrm{train}\left(\boldsymbol{\omega},\boldsymbol{\alpha}\right)}={\mathcal{L}\left(\boldsymbol{\omega},\boldsymbol{\alpha};\mathbf{x}_\mathrm{train}\right)}$ and ${\mathcal{L}_\mathrm{val}\left(\boldsymbol{\omega},\boldsymbol{\alpha}\right)}={\mathcal{L}\left(\boldsymbol{\omega},\boldsymbol{\alpha};\mathbf{x}_\mathrm{val}\right)}$ lies in the input, $\mathbf{x}$. Assume that the training dataset contains a sample, ${\mathbf{x}_\mathrm{train}}={\mathbf{1}}$ and validation dataset contains another sample, ${\mathbf{x}_\mathrm{val}}={\mathbf{2}}$. It is easy to derive that the local optimum of $\mathcal{L}_\mathrm{train}\left(\boldsymbol{\omega},\boldsymbol{\alpha}\right)$ is ${\boldsymbol{\omega}^\star\left(\boldsymbol{\alpha}\right)}={\boldsymbol{\alpha}}$. Substituting ${\mathbf{x}_\mathrm{val}}={\mathbf{2}}$ into the loss function yields ${\mathcal{L}_\mathrm{val}\left(\boldsymbol{\omega},\boldsymbol{\alpha}\right)}={\left(\boldsymbol{\omega}\cdot\mathbf{2}-\boldsymbol{\alpha}\right)^2}$. When ${\boldsymbol{\alpha}}={\boldsymbol{\alpha}_t}$ ($\boldsymbol{\alpha}_t$ denotes the current status of $\boldsymbol{\alpha}$), $\boldsymbol{\omega}$ arrives at $\boldsymbol{\omega}^\star\left(\boldsymbol{\alpha}_t\right)$, ${\mathbf{g}_1}={2\left(\boldsymbol{\alpha}_t-2\boldsymbol{\alpha}_t\right)}={-2\boldsymbol{\alpha}_t}$ and ${\mathbf{g}_2}={4\boldsymbol{\alpha}_t}\neq{\mathbf{0}}$. In other words, when $\boldsymbol{\omega}$ arrives at $\boldsymbol{\omega}^\star\left(\boldsymbol{\alpha}\right)$, ${\mathcal{L}_\mathrm{val}\left(\boldsymbol{\omega}^\star\left(\boldsymbol{\alpha}\right),\boldsymbol{\alpha}\right)}={\boldsymbol{\alpha}^2}$, we have ${\left.\nabla_{\boldsymbol{\alpha}}\mathcal{L}_\mathrm{val}\!\left(\boldsymbol{\omega}^\star\!\left(\boldsymbol{\alpha}\right),\boldsymbol{\alpha}\right)\right|_{\boldsymbol{\alpha}=\boldsymbol{\alpha}_t}}={2\boldsymbol{\alpha}_t}\neq{\mathbf{g}_1}$.

In summary, $\mathbf{g}_2$ will not be zero even when $\boldsymbol{\omega}^\star$ is achieved, which implies that ignoring $\mathbf{g}_2$ may lead to a wrong direction of optimization -- in the toy example above, the loss will increase monotonically towards infinity if $\mathbf{g}_2$ is ignored. We believe that this inaccuracy also contributed to the weird phenomena in real-world data (\textit{i.e.}, when DARTS was run on CIFAR10, the optimization process collapses to a dummy network architecture which produces poor classification results).

To provide better understanding of the search process, we plot the statistics from two search processes, without and with the amending term, in Figure~\ref{fig:toy_example_curves}. Without the amending term, \textit{i.e.}, the computation of $\mathbf{g}_2$ suffers considerable inaccuracy, $\boldsymbol{\alpha}$ (shown as the red curve) is always going towards an incorrect direction. The validation loss is first reduced (because the initialized $\boldsymbol{\omega}$ is poor and thus in the first half of the search process, the accuracy improvement brought by $\boldsymbol{\omega}$ surpasses the accuracy drop brought by $\boldsymbol{\alpha}$), but, as the search process continues, the first-order gradient is reduced and the inaccuracy of $\mathbf{g}_2$ dominates $\left.\nabla_{\boldsymbol{\alpha}}\mathcal{L}_\mathrm{val}\!\left(\boldsymbol{\omega}^\star\!\left(\boldsymbol{\alpha}\right),\boldsymbol{\alpha}\right)\right|_{\boldsymbol{\alpha}=\boldsymbol{\alpha}_t}$. Consequently, both parameters are quickly pushed away from the optimum and the search process `goes wild' (\textit{i.e.}, the loss `converges' to infinity). This problem is perfectly solved after the amending term is applied.

\subsection{Proof of ${\left\langle\mathbf{g}_2',\mathbf{g}_2\right\rangle}\geqslant{0}$}
\label{maths:inner_product}

\textit{This part corresponds to Section~\ref{approach:mathematics} in the main article.}

Substituting $\mathbf{g}_2$ and $\mathbf{g}_2'$ into ${\left\langle\mathbf{g}_2',\mathbf{g}_2\right\rangle}$ gives:
\begin{eqnarray}
{\left\langle\mathbf{g}_2',\mathbf{g}_2\right\rangle}={\eta\cdot\nabla_{\boldsymbol{\omega}}\mathcal{L}_\mathrm{val}\!\left(\boldsymbol{\omega},\boldsymbol{\alpha}\right)|_{\boldsymbol{\omega}=\boldsymbol{\omega}^\star\!\left(\boldsymbol{\alpha}_t\right),\boldsymbol{\alpha}=\boldsymbol{\alpha}_t}^\top\cdot\mathbf{H}^{-1}\cdot\mathbf{A}\cdot\mathbf{H}\cdot\nabla_{\boldsymbol{\omega}}\mathcal{L}_\mathrm{val}\!\left(\boldsymbol{\omega},\boldsymbol{\alpha}\right)|_{\boldsymbol{\omega}=\boldsymbol{\omega}^\star\!\left(\boldsymbol{\alpha}_t\right),\boldsymbol{\alpha}=\boldsymbol{\alpha}_t}},
\end{eqnarray}
where ${\mathbf{A}}={\nabla_{\boldsymbol{\omega},\boldsymbol{\alpha}}^2\mathcal{L}_\mathrm{train}\!\left(\boldsymbol{\omega},\boldsymbol{\alpha}\right)|_{\boldsymbol{\omega}=\boldsymbol{\omega}^\star\!\left(\boldsymbol{\alpha}_t\right),\boldsymbol{\alpha}=\boldsymbol{\alpha}_t}^\top\cdot\nabla_{\boldsymbol{\alpha},\boldsymbol{\omega}}^2\mathcal{L}_\mathrm{train}\!\left(\boldsymbol{\omega},\boldsymbol{\alpha}\right)|_{\boldsymbol{\omega}=\boldsymbol{\omega}^\star\!\left(\boldsymbol{\alpha}_t\right),\boldsymbol{\alpha}=\boldsymbol{\alpha}_t}}$, is a semi-positive-definite matrix. According to the definition of Hesse matrix, $\mathbf{H}$ is a real symmetric positive-definite matrix.

Let $\left\{\boldsymbol{\psi}_k\right\}$ be an orthogonal set of eigenvectors with respect to $\mathbf{A}$ (\textit{i.e.}, ${\boldsymbol{\psi}_k^\top\cdot\boldsymbol{\psi}_k}={1})$, and $\left\{\lambda_k\right\}$ be the corresponding eigenvalues satisfying ${\lambda_k}\geqslant{\mathbf{0}}$. Let $\boldsymbol{\phi}_l$ be an eigenvector of $\mathbf{H}\cdot\mathbf{A}\cdot\mathbf{H}^{-1}+\mathbf{H}^{-1}\cdot\mathbf{A}\cdot\mathbf{H}$, and expanding $\mathbf{H}\cdot\boldsymbol{\phi}_l$ and $\mathbf{H}^{-1}\cdot\boldsymbol{\phi}_l$ based on $\left\{\boldsymbol{\psi}_k\right\}$ derives the following equation for any $l$:
\begin{eqnarray}
\label{eqn:expand_beta}
{\boldsymbol{\phi}_l}={\sum_{k=1}^Ka_k\cdot\left(\mathbf{H}\cdot\boldsymbol{\psi}_k\right)}={\sum_{k=1}^Kb_k\cdot\left(\mathbf{H}^{-1}\cdot\boldsymbol{\psi}_k\right)},
\end{eqnarray}
where $\left\{a_k\right\}$ and $\left\{b_k\right\}$ are the corresponding expansion coefficients. According to the definition of eigenvalues, we have:
\begin{eqnarray}
\label{eqn:nature_of_eigenvalues}
{\left(\mathbf{H}\cdot\mathbf{A}\cdot\mathbf{H}^{-1}+\mathbf{H}^{-1}\cdot\mathbf{A}\cdot\mathbf{H}\right)\cdot\boldsymbol{\phi}_l}={\mu_l\cdot\boldsymbol{\phi}_l},
\end{eqnarray}
where $\mu_l$ is the corresponding eigenvalue of $\boldsymbol{\phi}_l$. Substituting Eqn~\eqref{eqn:expand_beta} into Eqn~\eqref{eqn:nature_of_eigenvalues}, we obtain:
\begin{eqnarray}
\label{eqn:wait_to_multiply}
{\sum_{k=1}^Ka_k\cdot\lambda_k\left(\mathbf{H}\cdot\boldsymbol{\psi}_k\right)+\sum_{k=1}^Kb_k\cdot\left(\lambda_k-\mu_l\right)\cdot\left(\mathbf{H}^{-1}\cdot\boldsymbol{\psi}_k\right)}={\mathbf{0}}.
\end{eqnarray}
Left-multiplying Eqn~\eqref{eqn:wait_to_multiply} by $\left[\sum_{k=1}^K b_k\cdot\left(\lambda_k-\mu_l\right)\cdot\left(\mathbf{H}^{-1}\cdot\boldsymbol{\psi}_k\right)\right]^\top$ obtains:
\begin{eqnarray}
{\sum_{k=1}^K\left(\lambda_k-\mu_l\right)\cdot\lambda_k\cdot a_k\cdot b_k}={-\left\|\sum_{k=1}^Kb_k\cdot\left(\lambda_k-\mu_l\right)\cdot\left(\mathbf{H}^{-1}\cdot\boldsymbol{\psi}_k\right)\right\|_{2}^{2}}\leqslant{0}.
\end{eqnarray}
$\mathbf{A}$ is a real symmetric matrix sized $M_1\times M_1$, and it is obtained by multiplying a $M_1\times M_2$ matrix to a $M_2\times M_1$. Here, $M_1$ and $M_2$ are the dimensionality of $\boldsymbol{\omega}$ and $\boldsymbol{\alpha}$, respectively, and $M_1$ is often much larger than $M_2$ (\textit{e.g.}, millions vs. hundreds in DARTS). Hence, $\mathbf{A}$ has much fewer different eigenvalues compared to its dimensionality, and so we can choose many sets of $\left\{\boldsymbol{\alpha}_k\right\}$ which are orthogonal to each other, and generate many sets of $\left\{a_k\right\}$ and $\left\{b_k\right\}$ satisfying ${\sum_{k=1}^K\mathbf{a}_k\mathbf{b}_k}={\boldsymbol{\beta}^\top\cdot\boldsymbol{\beta}}\geqslant{0}$. That being said, in most of time, we can choose one set of eigenvectors from the subspace, $\left\{\boldsymbol{\alpha}_k\right\}$, and the elements within are orthogonal to each other, which makes ${\mathbf{a}_k\mathbf{b}_k}\geqslant{0}$ in most cases, and hence ${\mu_l}\geqslant{0}$ for any $l$.

Since all of the eigenvalues with respect to the real symmetric matrix $\mathbf{H}\cdot\mathbf{A}\cdot\mathbf{H}^{-1}+\mathbf{H}^{-1}\cdot\mathbf{A}\cdot\mathbf{H}$ is not smaller than zero, so it is semi-positive-definite. This derives that $\mathbf{H}^{-1}\cdot\mathbf{A}\cdot\mathbf{H}$ is semi-positive-definite, and thus ${\left\langle\mathbf{g}_2',\mathbf{g}_2\right\rangle}\geqslant{0}$.

\subsection{Computing $\mathbf{g}_2'$ Using Finite Difference Approximation}
\label{maths:computation}

\textit{This part corresponds to Section~\ref{approach:mathematics} in the main article.}

Although we avoid the computation of $\mathbf{H}^{-1}$, the inverse matrix of $\mathbf{H}$, computing $\mathbf{H}\cdot\left.\nabla_{\boldsymbol{\omega}}\mathcal{L}_\mathrm{val}\!\left(\boldsymbol{\omega},\boldsymbol{\alpha}\right)\right|_{\boldsymbol{\omega}=\boldsymbol{\omega}^\star\!\left(\boldsymbol{\alpha}_t\right),\boldsymbol{\alpha}=\boldsymbol{\alpha}_t}$ in $\mathbf{g}_2'$ is still a problem since the size of $\mathbf{H}$ is often very large, \textit{e.g.}, over one million. We use finite difference approximation just as used in the \textit{second-order} DARTS~\cite{liu2019darts}.

Let $\boldsymbol{\epsilon}$ be a small scalar, which is set to be $\mathbf{0.01}/\left\|\nabla_{\boldsymbol{\omega}}\mathcal{L}_\mathrm{val}\!\left(\boldsymbol{\omega},\boldsymbol{\alpha}\right)\right\|_{2}$ as in the original DARTS. Using finite difference approximation to compute the gradient around ${\boldsymbol{\omega}_1}={\boldsymbol{\omega}+\boldsymbol{\epsilon}\cdot\nabla_{\boldsymbol{\omega}}\mathcal{L}_\mathrm{val}\!\left(\boldsymbol{\omega},\boldsymbol{\alpha}\right)}$ and ${\boldsymbol{\omega}_2}={\boldsymbol{\omega}-\boldsymbol{\epsilon}\cdot\nabla_{\boldsymbol{\omega}}\mathcal{L}_\mathrm{val}\!\left(\boldsymbol{\omega},\boldsymbol{\alpha}\right)}$ gives:
\begin{equation}
{\mathbf{H}\cdot\nabla_{\boldsymbol{\omega}}\mathcal{L}_\mathrm{val}\!\left(\boldsymbol{\omega},\boldsymbol{\alpha}\right)}\approx{\frac{\nabla_{\boldsymbol{\omega}}\mathcal{L}_\mathrm{train}\!\left(\boldsymbol{\omega}_1,\boldsymbol{\alpha}\right)-\nabla_{\boldsymbol{\omega}}\mathcal{L}_\mathrm{train}\!\left(\boldsymbol{\omega}_2,\boldsymbol{\alpha}\right)}{2\boldsymbol{\epsilon}}}.
\end{equation}
Let ${\boldsymbol{\omega}_3}={\boldsymbol{\omega}+\frac{1}{2}\left[\nabla_{\boldsymbol{\omega}}\mathcal{L}_\mathrm{train}\!\left(\boldsymbol{\omega}_1,\boldsymbol{\alpha}\right)-\nabla_{\boldsymbol{\omega}}\mathcal{L}_\mathrm{train}\!\left(\boldsymbol{\omega}_2,\boldsymbol{\alpha}\right)\right]}$ and ${\boldsymbol{\omega}_4}={\boldsymbol{\omega}-\frac{1}{2}\left[\nabla_{\boldsymbol{\omega}}\mathcal{L}_\mathrm{train}\!\left(\boldsymbol{\omega}_1,\boldsymbol{\alpha}\right)-\nabla_{\boldsymbol{\omega}}\mathcal{L}_\mathrm{train}\!\left(\boldsymbol{\omega}_2,\boldsymbol{\alpha}\right)\right]}$, we have:
\begin{equation}
{\mathbf{g}_2'}\approx{-\boldsymbol{\eta}\times\frac{\nabla_{\boldsymbol{\alpha}}\mathcal{L}_\mathrm{train}\!\left(\boldsymbol{\omega}_3,\boldsymbol{\alpha}\right)-\nabla_{\boldsymbol{\alpha}}\mathcal{L}_\mathrm{train}\!\left(\boldsymbol{\omega}_4,\boldsymbol{\alpha}\right)}{2\boldsymbol{\epsilon}}}.
\end{equation}

\section{Technical Details and Search Costs}
\label{details}

\subsection{More about Hyper-Parameter Consistency}
\label{details:consistency}

An important contribution of this paper is to reveal the need of hyper-parameter consistency, \textit{i.e.}, using the same set of hyper-parameters, including the depth and width of the super-network, the Dropout ratio, whether to use the auxiliary loss term, \textbf{not} to prune edges at the end of search, \textit{etc}. This is a natural idea in both theory and practice, but most existing work ignored it, possibly because its importance was hidden behind the large optimization gap brought by the inaccurate gradient estimation in the bi-level optimization problem.

We point out that even when bi-level optimization works accurately, we may miss the optimal architecture(s) without hyper-parameter consistency. This is because each hyper-parameter will more or less change the value of the loss function and therefore impact the optimal architecture. We provide a few examples here.

\begin{itemize}
\item The most sensitive change may lie in the edge pruning process, \textit{i.e.}, preserving $8$ out of $14$ edges in each cell and eliminating others. This may incur significant accuracy drop even when only one edge gets pruned. For example, in the normal cell searched on the $\mathcal{S}_1$-A space (see Figure~\ref{fig:architectures_S1}), removing the \textit{skip-connect} operator will cause the re-training error to increase from $2.71\%\pm0.09$ to $3.36\%\pm0.08$. The dramatic performance drop is possibly due to the important role played by the pruned edge, \textit{e.g.}, the pruned \textit{skip-connect} may contribute to rapid information propagation in network training.
\item Some hyper-parameters will potentially change the optimal architecture. The basic channel width and network depth are two typical examples. When the channel number is small, the network may expect large convolutional kernels to guarantee a reasonable amount of trainable parameters; also, a deep architecture may lean towards small convolutional kernels since the receptive field is not a major bottleneck. However, the original DARTS did not unify these quantities in search and re-training, which often results in sub-optimality as noticed in~\cite{chen2019progressive}.
\item Other hyper-parameters, in particular those related to the training configuration, are also important. For example, if we expect Dropout to be used during the re-training stage, the optimal architecture may contain a larger number of trainable parameters; if the auxiliary loss tower is used, the optimal architecture may be deeper. Without hyper-parameter consistency, these factors cannot be taken into consideration.
\end{itemize}

\subsection{Details of the Two-Stage Search Process}
\label{details:edge_selection}

To avoid the optimization gap brought by edge pruning, we adopt a two-stage search process in which we perform edge search followed by operator search. In the DARTS setting, the first stage involves preserving $8$ out of $14$ edges to be preserved. For each node, indexed $j$, we use $\mathcal{E}_j$ to denote the set of all combinations of edges (in the DARTS setting, each node preserves two input edges, so $\mathcal{E}_j$ contains all $\left(i_1,i_2\right)$ pairs with ${0}\leqslant{i_1}<{i_2}<{j}$). The output of node $j$ is therefore computed as:
\begin{equation}
\label{eqn:node_output}
{\mathbf{y}_j}={\sum_{\left(i_1,i_2\right)\in\mathcal{E}_j}\frac{\exp\!\left(\beta^{\left(i_1,i_2\right)}\right)}{\sum_{\left(i_1',i_2'\right)\in\mathcal{E}_j}\exp\!\left(\beta^{\left(i_1',i_2'\right)}\right)}\cdot\left(\mathbf{y}_{i_1}+\mathbf{y}_{i_2}\right)}.
\end{equation}
where $\beta^{\left(i_1,i_2\right)}$ denotes the edge-selection parameter of the node combination of $\left(i_1,i_2\right)$. This mechanism is similar to the edge normalization introduced in~\cite{xu2020pc} but we take the number of preserved inputs into consideration.

After edge selection is finished, a regular operator search process, starting from scratch, follows on the preserved edges to determine the final architecture.

\subsection{Search Cost Analysis}
\label{details:costs}

%We are sincerely sorry that the search costs (GPU-days) provided in the last row of Tables~\ref{tab:comparison_CIFAR10} and Table~\ref{tab:comparison_ILSVRC2012} are incorrect due to a
%mistake in computation. They should be 3.1 GPU-day.

Note that the first (edge selection) stage requires around $2\times$ computational costs compared to the second (operator selection) stage. This is because edge selection works on $14$ edges while operator selection on $8$ edges. Fortunately, the former stage can be skipped (at no costs) if we choose to search operators on a fixed edge configuration, which also reports competitive search performance.

\begin{table}[!t]
\begin{center}
\small
\resizebox{\textwidth}{!}{\begin{tabular}{lcccccccc}
\hline
\textbf{\multirow{2}{*}{Search Method}} & \multicolumn{5}{c}{\textbf{Test Err. (\%)}} & \multicolumn{3}{c}{\textbf{Params (M)}} \\
\cmidrule(lr){2-6}\cmidrule(lr){7-9}
{} & \textbf{Seed \#1} &\textbf{Seed \#2} & \textbf{Seed \#3} & \textbf{average} & \textbf{best} & \textbf{Seed \#1} &\textbf{Seed 2} & \textbf{Seed \#3} \\
\hline
Random Search$^\dagger$          & - & - & - & - & 2.95 & - & - & - \\
\hline
DARTS (early termination)$^\dagger$    & - & - & - & 3.71$\pm$1.14 & 3.07 & - & - & - \\
\hline
R-DARTS (Dropout)$^\dagger$     \\
\quad Drop Prob. = 0.0 & 5.30 & 3.73 & 3.34 & 4.12$\pm$1.04 & 3.34 & 2.21 & 2.43 & 2.85 \\
\quad Drop Prob. = 0.2 & 3.22 & 3.16 & 3.26 & 3.21$\pm$0.05 & 3.16 & 3.62 & 4.04 & 2.99 \\
\quad Drop Prob. = 0.4 & 2.93 & 3.15 & 3.52 & 3.20$\pm$0.30 & 2.93 & 4.10 & 3.74 & 3.38 \\
\quad Drop Prob. = 0.6 & 3.09 & 5.44 & 3.56 & 4.03$\pm$1.24 & 3.09 & 4.46 & 2.30 & 2.66 \\
\hline
Our Approach, pruned edges & 2.76$\pm$0.10 & 2.64$\pm$0.07 & 2.68$\pm$0.09 & 2.69$\pm$0.06 & 2.64$\pm$0.07 & 3.09 & 3.21 & 3.40 \\
Our Approach, searched edges &2.58$\pm$0.07 &2.55$\pm$0.13 & 2.76$\pm$0.04 & 2.63$\pm$0.11 & 2.55$\pm$0.13 & 3.84 & 3.21 & 2.85 \\
\hline
\end{tabular}}
\end{center}
\vspace{-0.4cm}
\caption{Comparison between DARTS (early termination), R-DARTS (Dropout) and our approach in the reduced search space of $\mathcal{S}_3$ with different seeds. $^\dagger$: we borrow the experimental results from~\cite{zela2020understanding}.}
\label{tab:comparison_Zela}
\end{table}

\section{Additional Experimental Results}

\subsection{Comparison to~\cite{zela2020understanding}}
\label{extra:understanding}

\begin{figure*}[!t]
\centering
\includegraphics[width=0.35\textwidth]{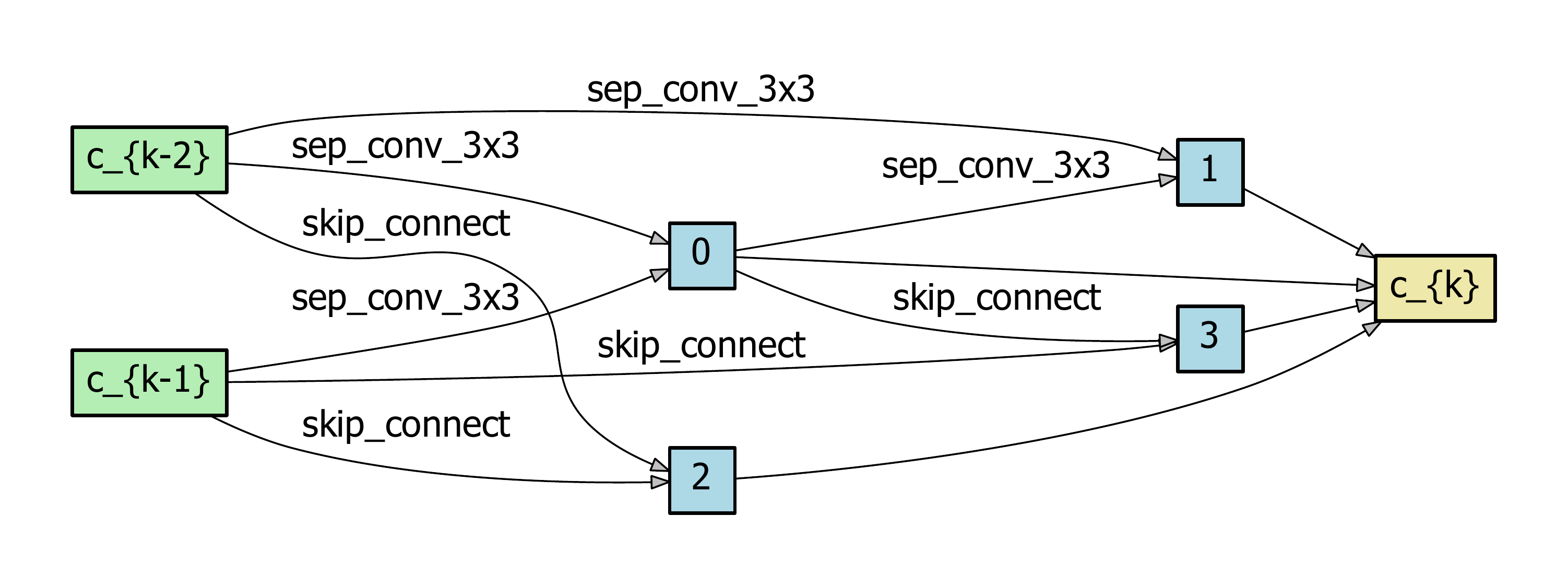}
\includegraphics[width=0.60\textwidth]{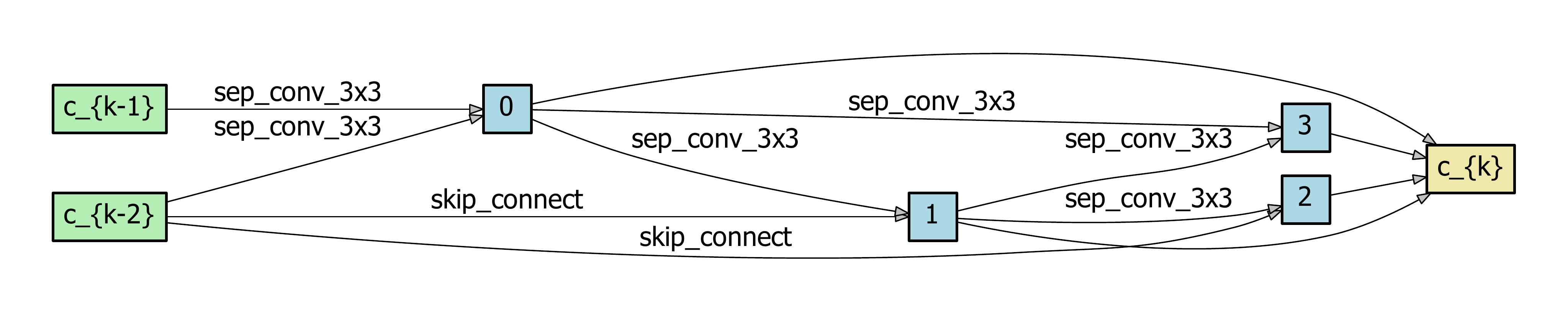} \\
\includegraphics[width=0.60\textwidth]{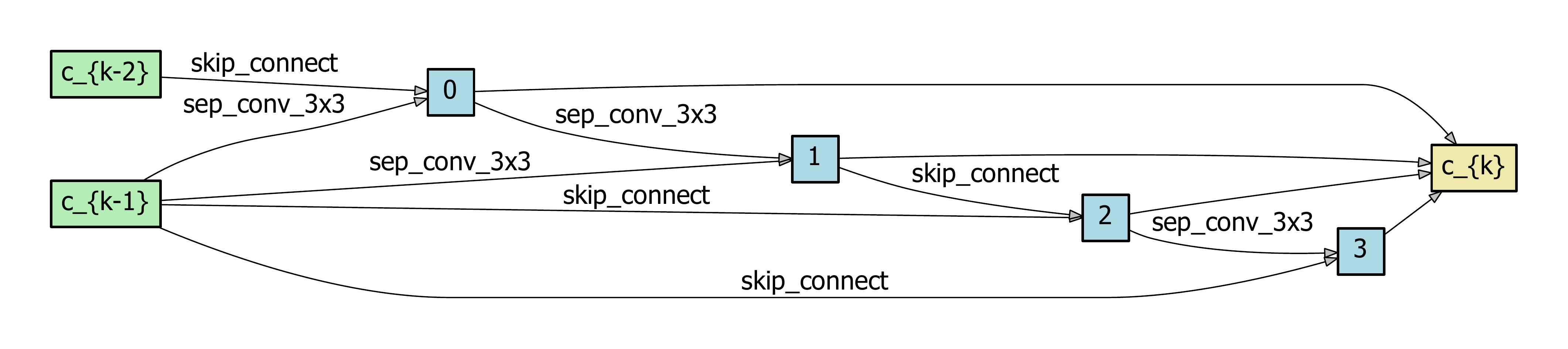}
\includegraphics[width=0.35\textwidth]{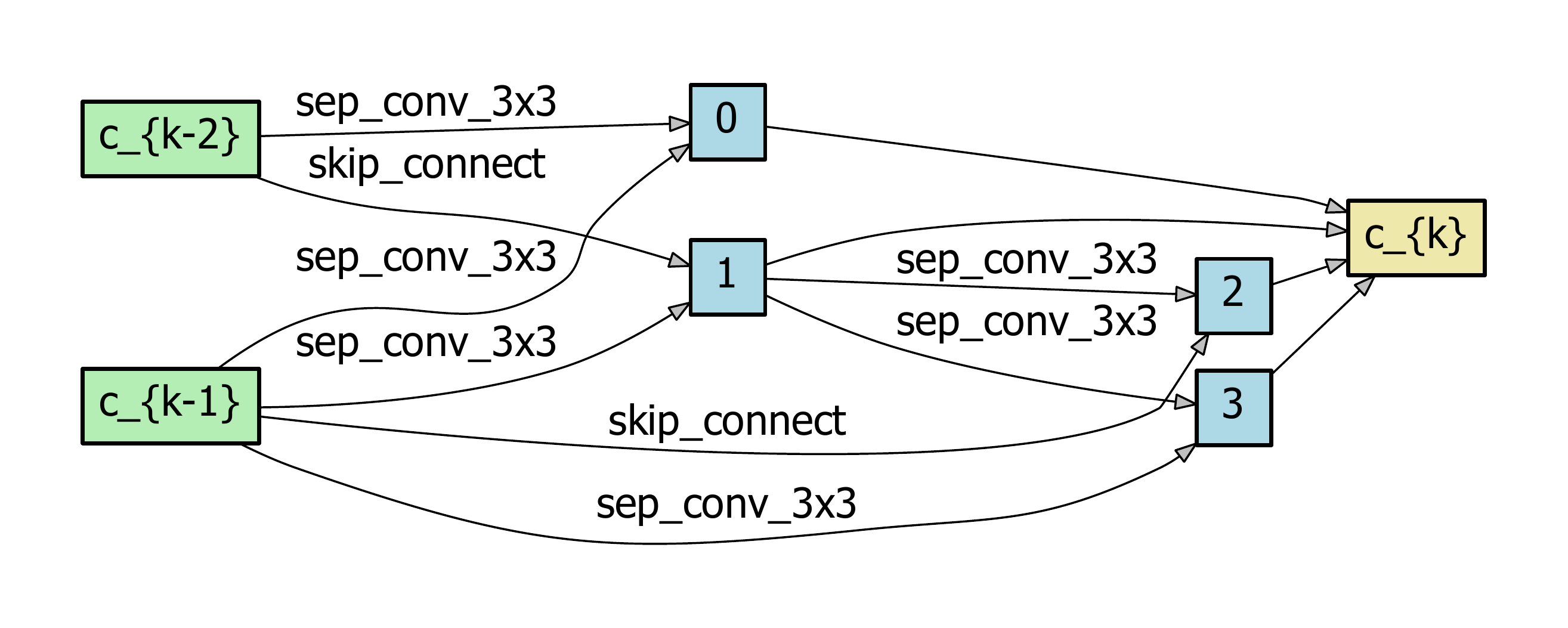} \\
\vspace{-0.4cm}
\caption{\textbf{Top}: the normal and reduction cells found in $\mathcal{S}_3$ with edge pruning. \textbf{Bottom}: the normal and reduction cells found in $\mathcal{S}_3$ with searched edges.}
\label{fig:architectures_Zela}
\end{figure*}

\cite{zela2020understanding} described various search spaces and demonstrated that the standard DARTS fails on them. They proposed various optimization strategies to robustify DARTS, including adding $\ell_2$-regularization, adding Dropout, and using early termination. We search in a reduced space which we denote as $\mathcal{S}_3$, in which the candidate operators only include \textit{sep-conv-3x3}, \textit{skip-connect}, and \textit{none}. Note that this is the `safest' search space identified in R-DARTS~\cite{zela2020understanding}, yet as shown in Table~\ref{tab:comparison_Zela}, R-DARTS often produced unsatisfying architectures, and choosing the best one over a few search trials can somewhat guarantee a reasonable architecture. Meanwhile, R-DARTS is sensitive to the search hyper-parameters such as the Dropout ratio.

Our approach works smoothly in this space, without the need of tuning hyper-parameters, and in either pruned or searched edges (see Figure~\ref{fig:architectures_Zela}). Thanks to the reduced search space, the best architecture often surpasses the numbers we have reported in Table~\ref{tab:comparison_CIFAR10}.

%A direct benefit of the search stability is that we can search only once while~\cite{zela2020understanding} needed several trials to determine the final architecture. The best architectures in three runs with pruned or searched edges are shown in Figure~\ref{fig:architectures_Zela}.

\subsection{$200$ Search Epochs for DARTS, P-DARTS, PC-DARTS, and Our Approach}
\label{extra:further_analysis}

\begin{figure*}
\centering
\includegraphics[width=0.40\textwidth]{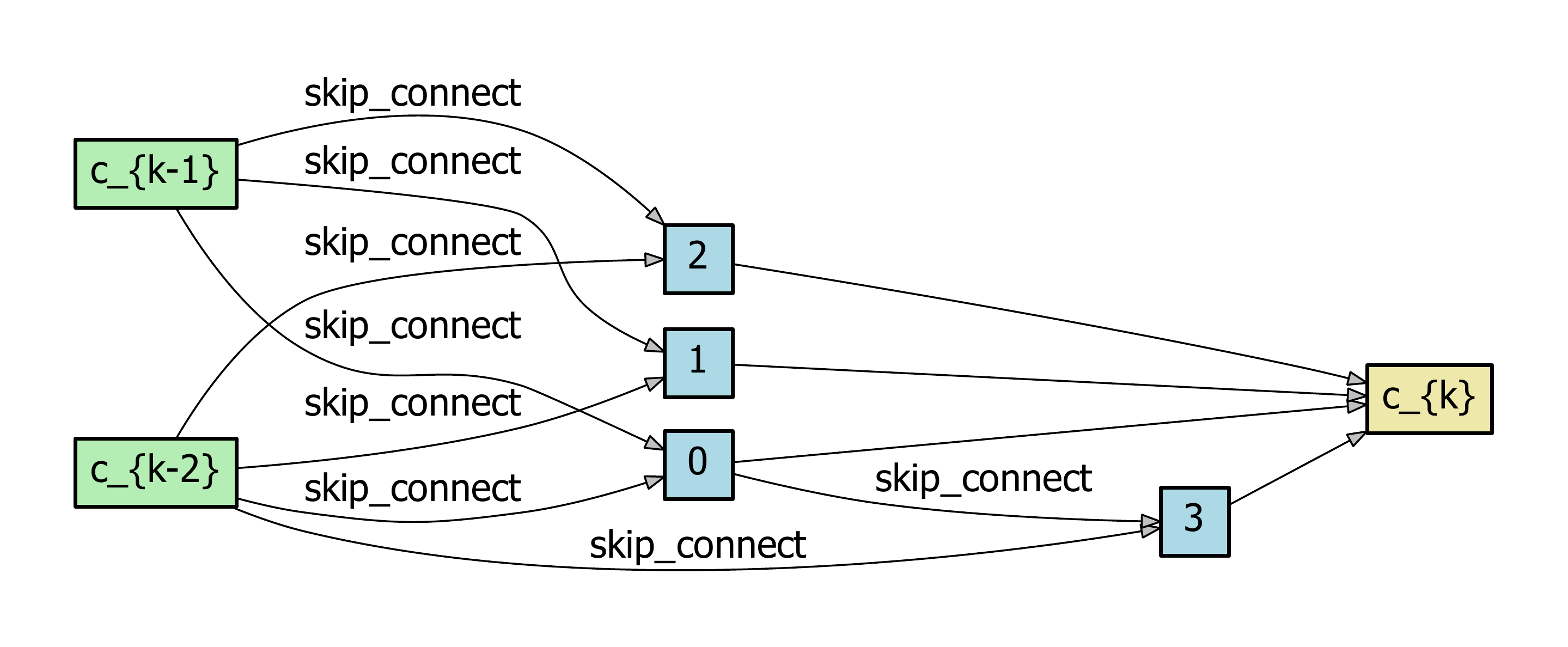}
\includegraphics[width=0.55\textwidth]{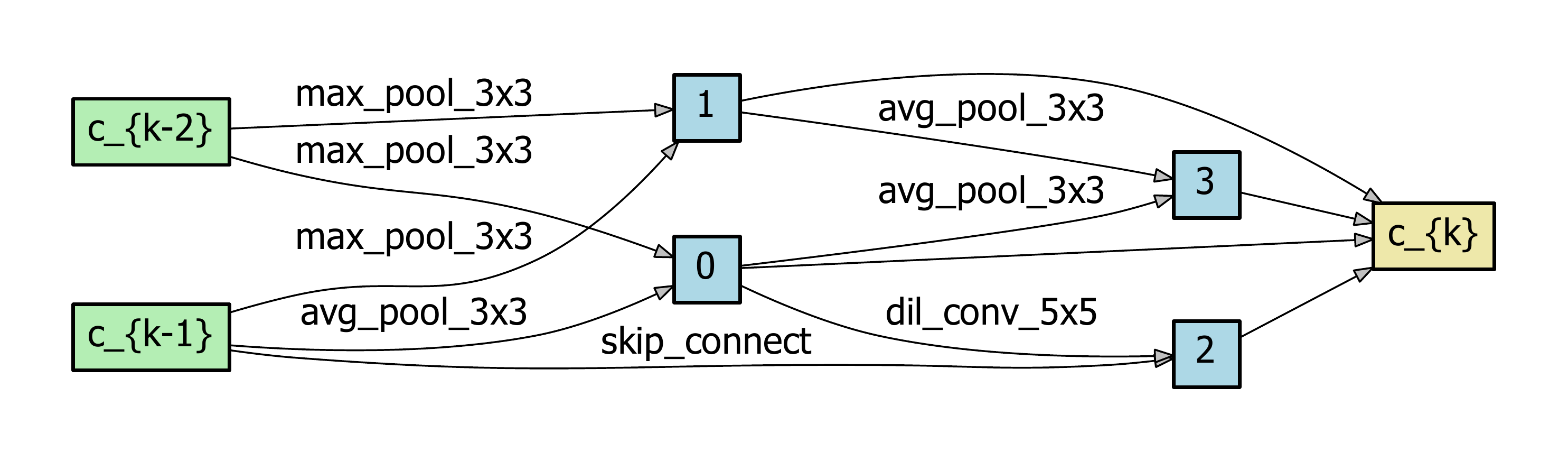} \\
\includegraphics[width=0.45\textwidth]{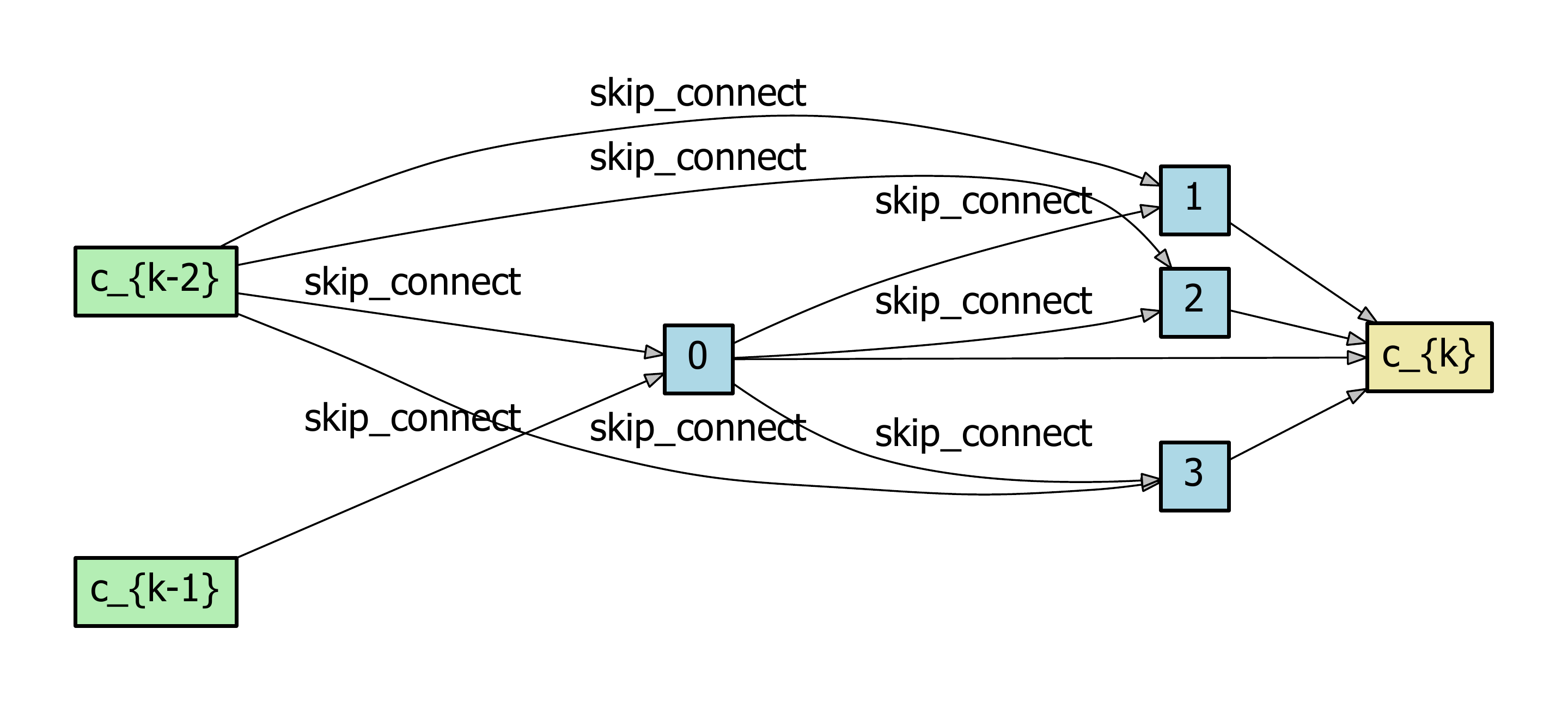}
\includegraphics[width=0.50\textwidth]{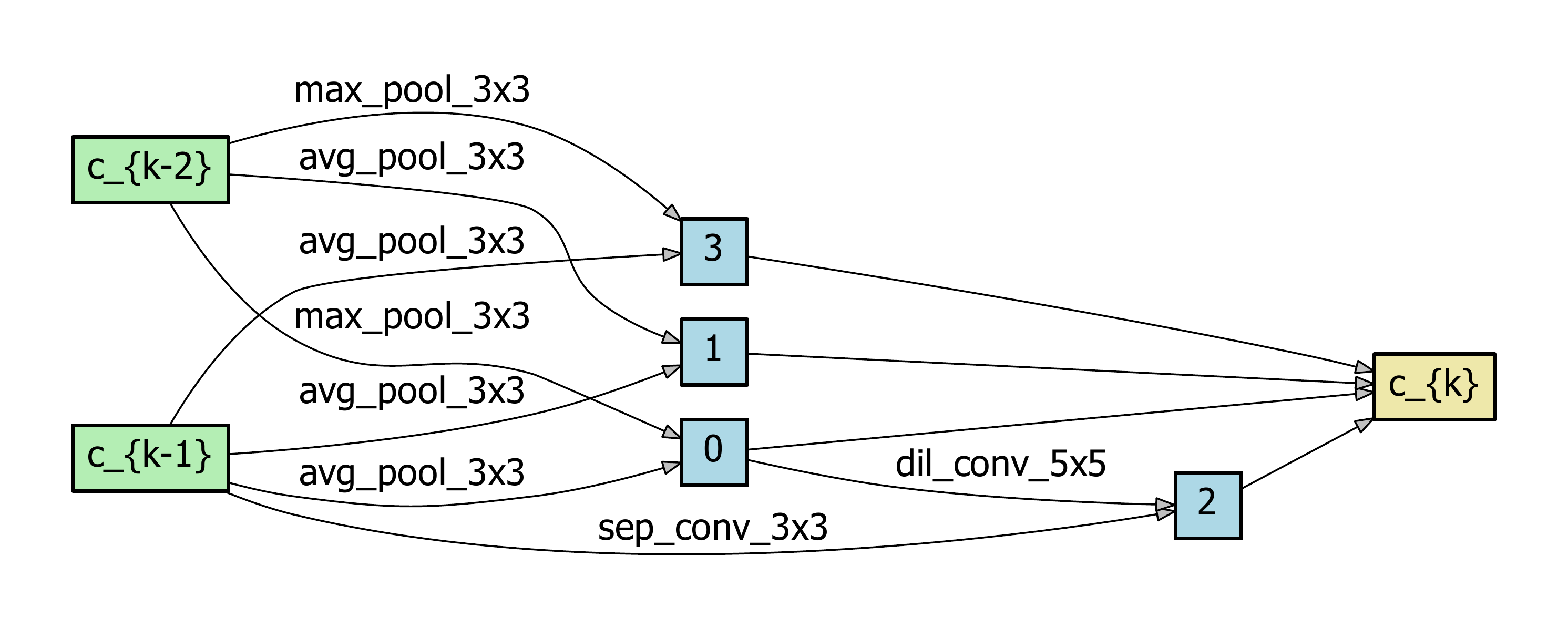} \\
\includegraphics[width=0.45\textwidth]{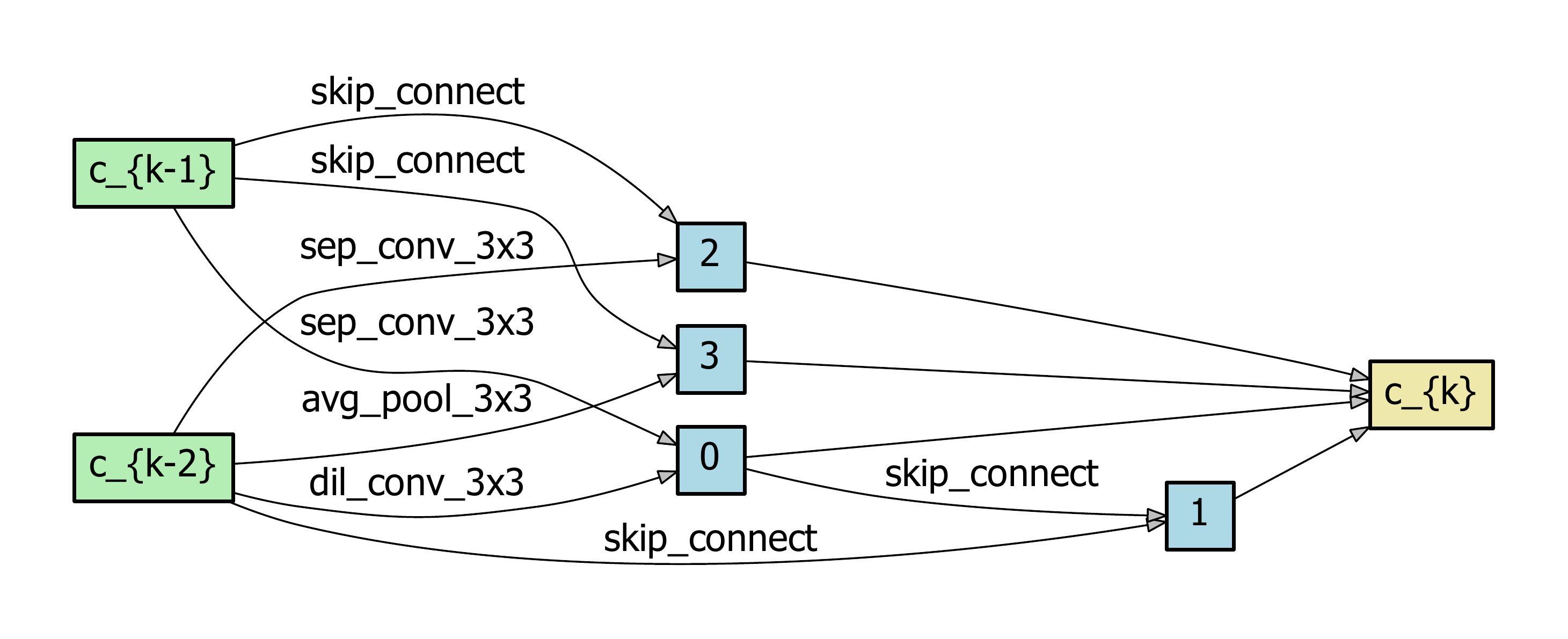}
\includegraphics[width=0.50\textwidth]{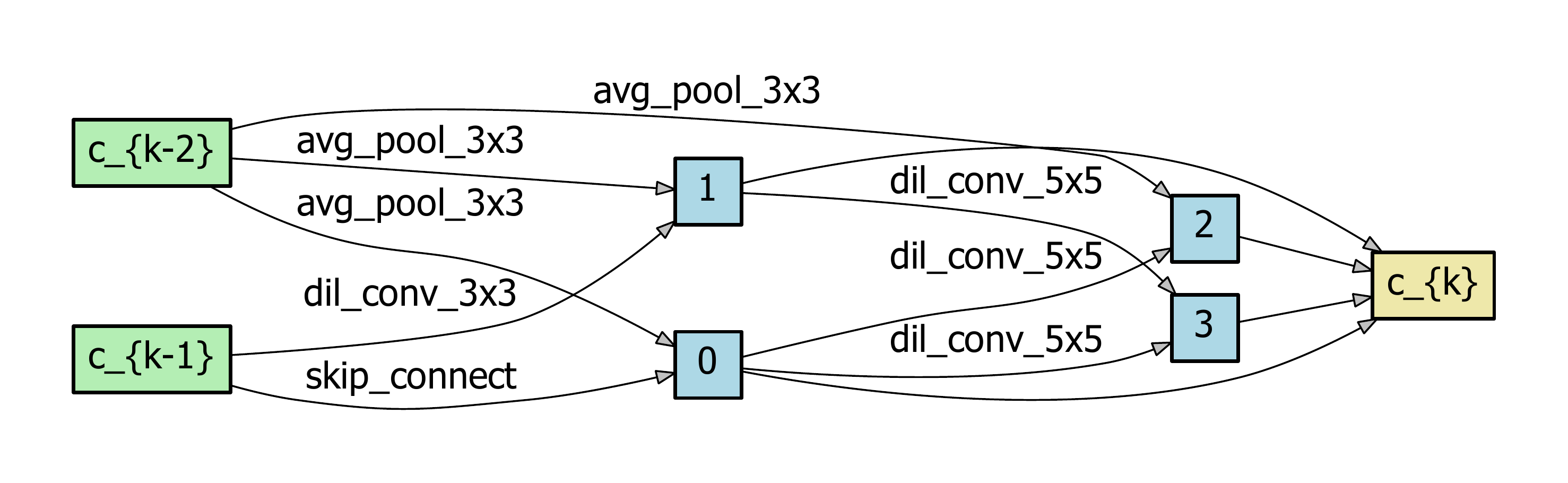} \\
\vspace{-0.4cm}
\caption{\textbf{Top}: the normal and reduction cells found by DARTS (test error: $6.18\%$). \textbf{Middle}: the normal and reduction cells found by P-DARTS (test error: $5.38\%$). \textbf{Bottom}: the normal and reduction cells found by PC-DARTS (test error: $3.15\%$).}
\label{fig:architectures_200_epochs}
\end{figure*}

We execute all algorithms for $200$ search epochs on CIFAR10. The searched architectures by DARTS, P-DARTS, and PC-DARTS are shown in Figure~\ref{fig:architectures_200_epochs}, and that of our approach shown in Figure~\ref{fig:architectures_CIFAR10}.

We find that both DARTS and P-DARTS failed completely into dummy architectures with all preserved operators to be \textit{skip-connect}. PC-DARTS managed to survive after $200$ epochs mainly due to two reasons: (i) the edge normalization technique is useful in avoiding \textit{skip-connect} to dominate; and (ii) PC-DARTS sampled $1/4$ channels of each operator, so that the algorithm converged slower -- $200$ epochs may not be enough for a complete collapse.

\begin{figure*}
\centering
\includegraphics[width=0.45\textwidth]{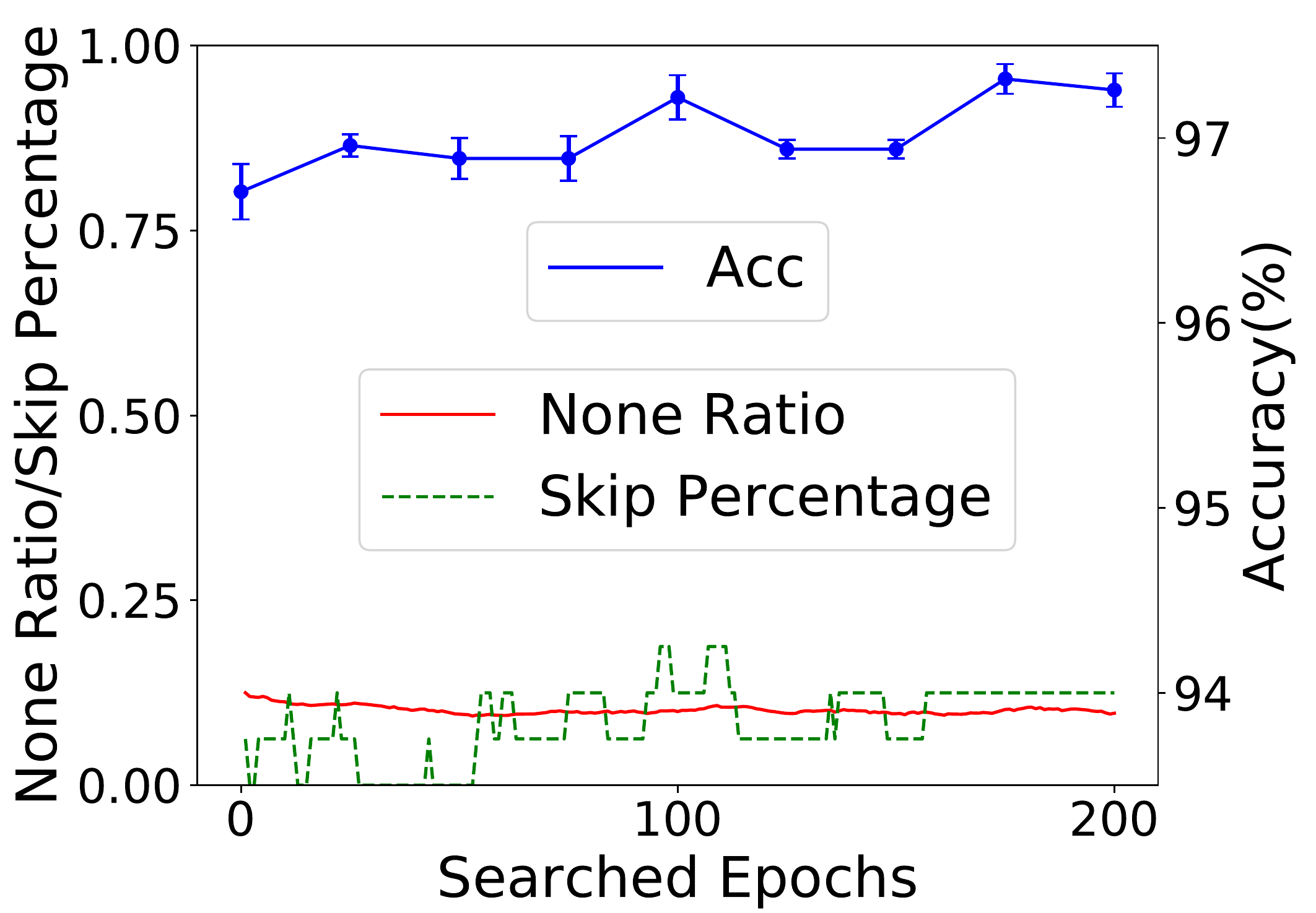}
\includegraphics[width=0.45\textwidth]{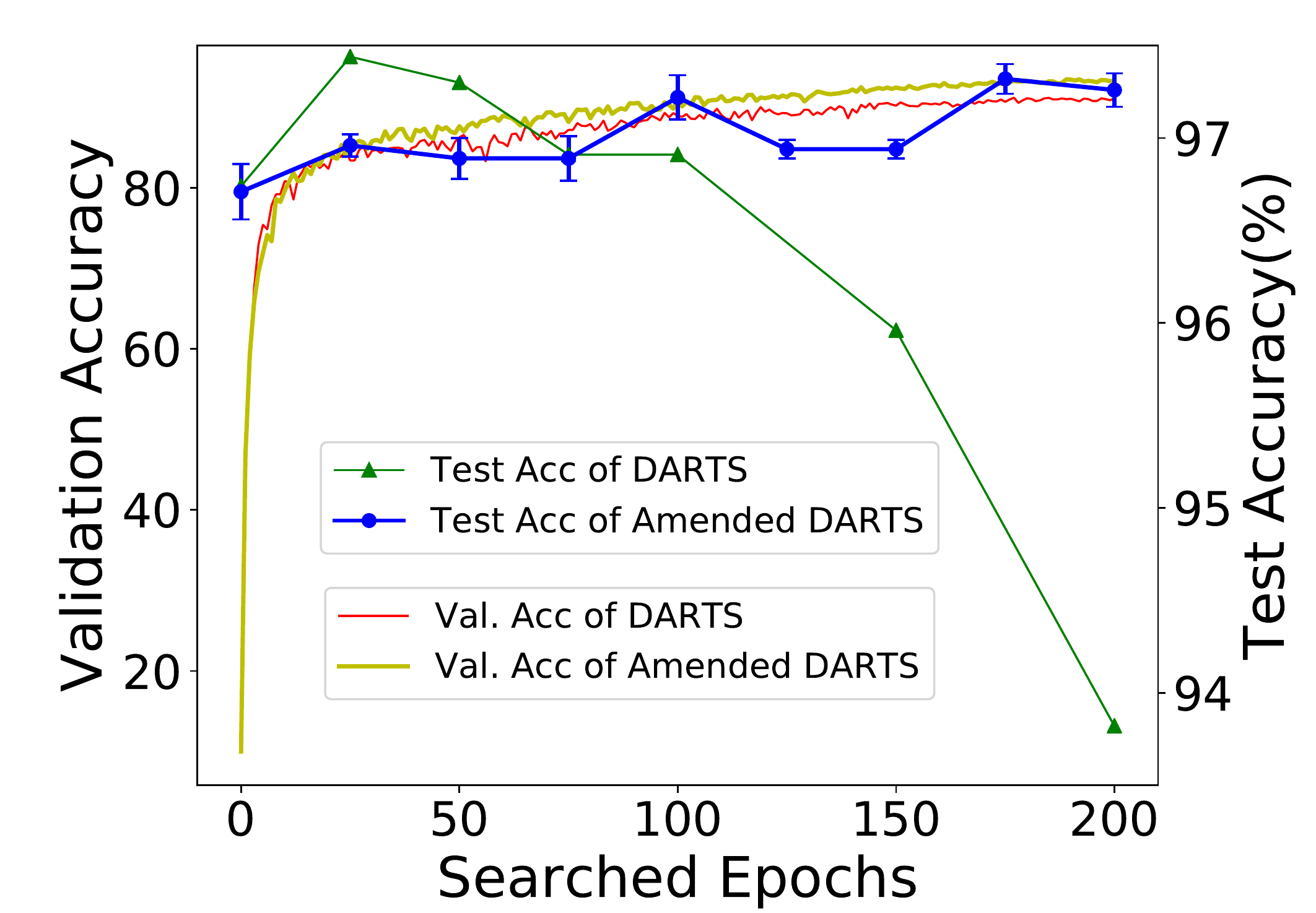}\\
\vspace{-0.4cm}
\caption{Details of a search process of our approach in $\mathcal{S}_1$ with $200$ search epochs. \textbf{Left}: Red, green and blue curves indicate the average weight of \textit{none}, the ratio of preserved \textit{skip-connect} operators, and the re-training accuracy, respectively (corresponding to Figure~\ref{fig:convergence_performance}). \textbf{Right}: Red and yellow curves indicate the validation accuracy of the super-network (in the search stage) searched by DARTS and our approach, and green and blue curves indicate the test accuracy (in the re-training stage) produced by the architectures found by DARTS and our approach, respectively.}
\label{fig:visualization_S1}
\end{figure*}

We also visualize our approach throughout $200$ epochs in $\mathcal{S}_1$ (under the edge-pruning setting) to investigate its behavior. In Figure~\ref{fig:visualization_S1}, the visualized factors include the average weight of \textit{none}, the ratio of preserved \textit{skip-connect} operators, the super-network validation accuracy, and the re-training accuracy of some checkpoints during the search process.

\newpage

\subsection{Searching with Different Seeds}
\label{extra:seeds}

On CIFAR10, we perform edge search (the first search stage) in $\mathcal{S}_2$ with different seeds (\textit{i.e.}, random initialization), yielding three sub-architectures named $\mathcal{S}_2$-A, $\mathcal{S}_2$-B, and $\mathcal{S}_2$-C. Then, we execute the operator search process with different seeds for $3$ times in $\mathcal{S}_1$ (with edge pruning), $\mathcal{S}_2$-A, $\mathcal{S}_2$-B, $\mathcal{S}_2$-C, and $\mathcal{S}_2$-F (fixed edges), and re-train each discovered architecture for $3$ times. The results are summarized in Table~\ref{tab:seeds_CIFAR10}. We also transfer some of the found architectures to ImageNet, and the corresponding results are listed in Table~\ref{tab:seeds_ILSVRC2012}. The edge-searched architectures in $\mathcal{S}_2$ report inferior performance compared to the edge-fixed ones, arguably due to the inconsistency of hyper-parameters.

\begin{figure*}
\centering
\subfigure[the normal cell of $\mathcal{S}_1$-A]{\includegraphics[width=0.45\textwidth]{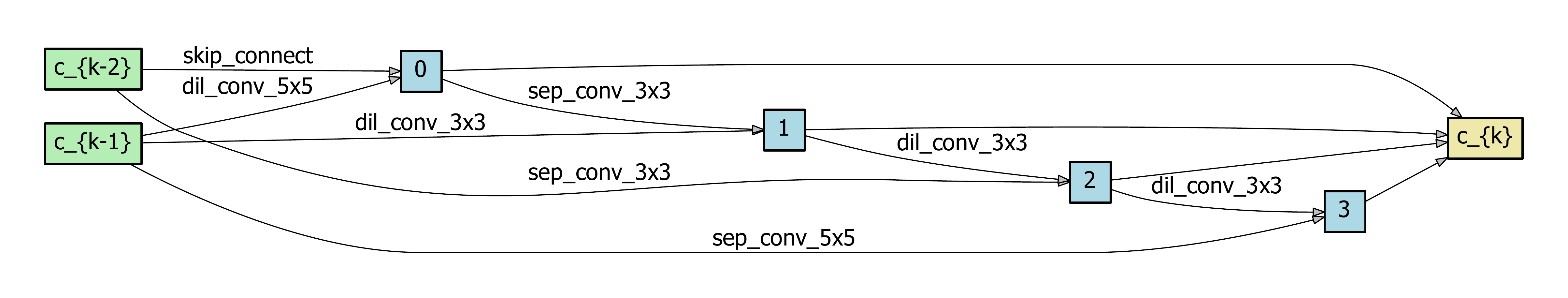}}
\subfigure[the reduction cell of $\mathcal{S}_1$-A]{\includegraphics[width=0.45\textwidth]{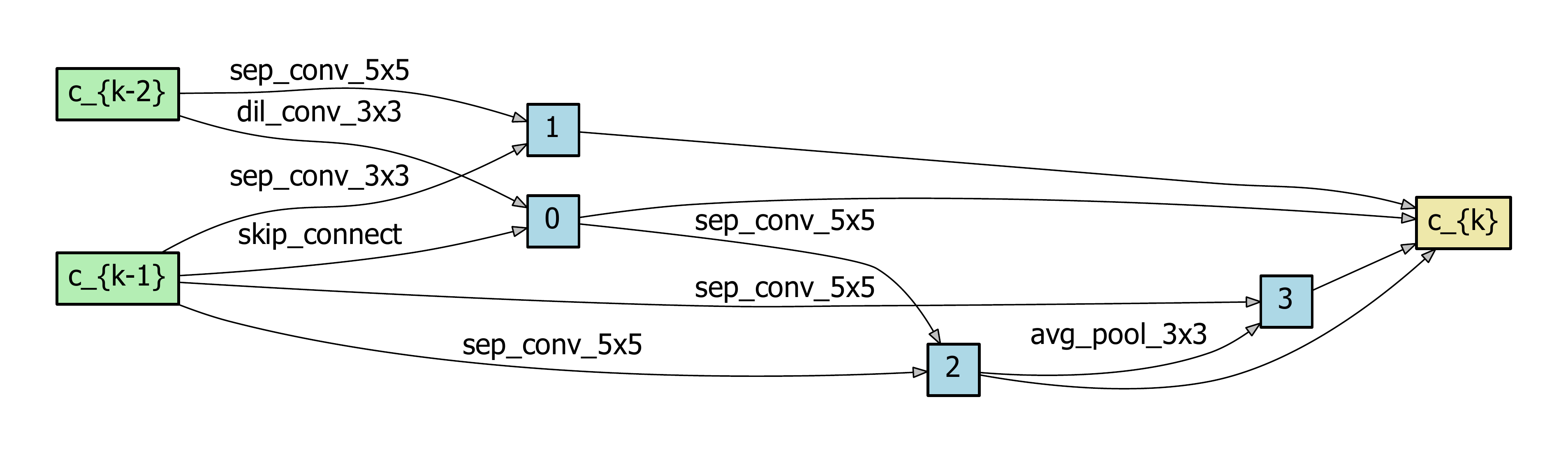}}\\
\vspace{0.2cm}
\subfigure[the normal cell of $\mathcal{S}_1$-B]{\includegraphics[width=0.45\textwidth]{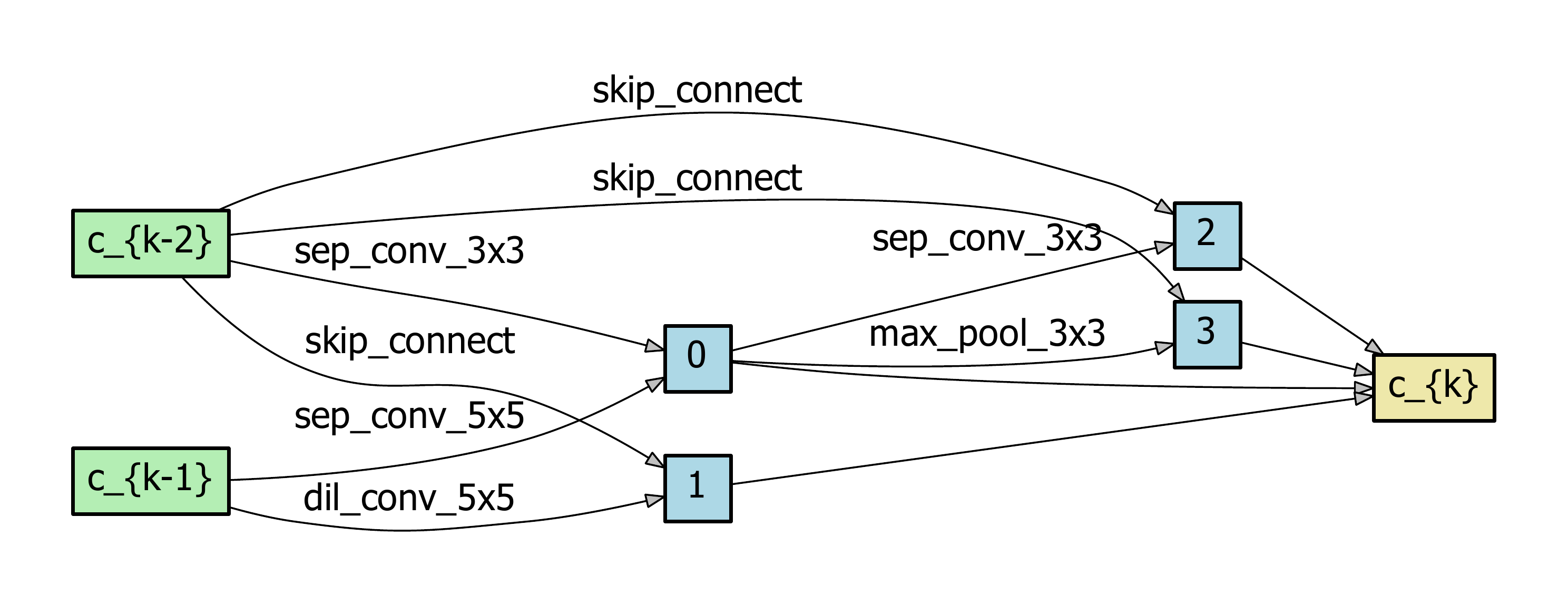}}
\subfigure[the reduction cell of $\mathcal{S}_1$-B]{\includegraphics[width=0.45\textwidth]{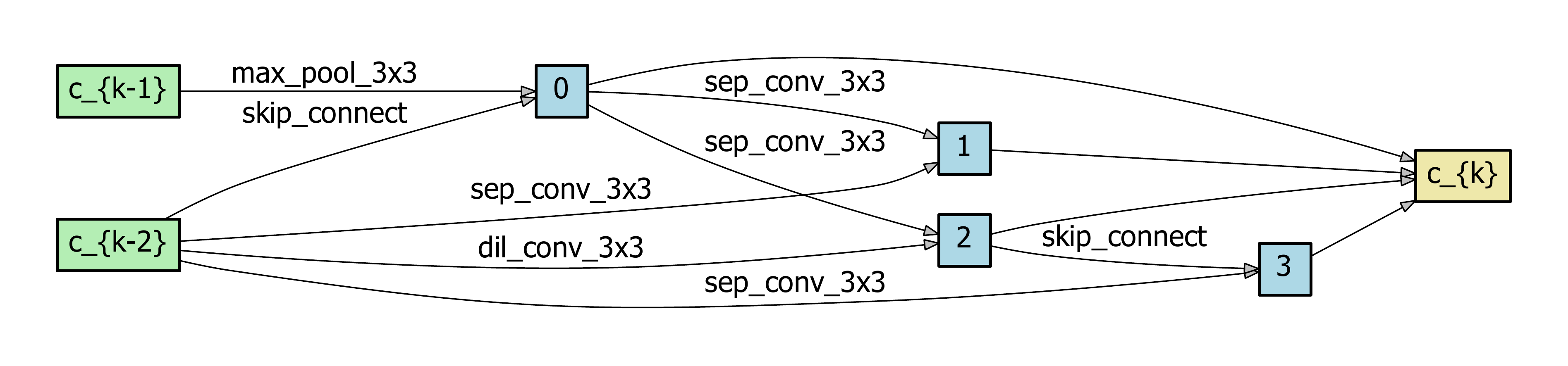}}\\
\vspace{0.2cm}
\subfigure[the normal cell of $\mathcal{S}_1$-C]{\includegraphics[width=0.45\textwidth]{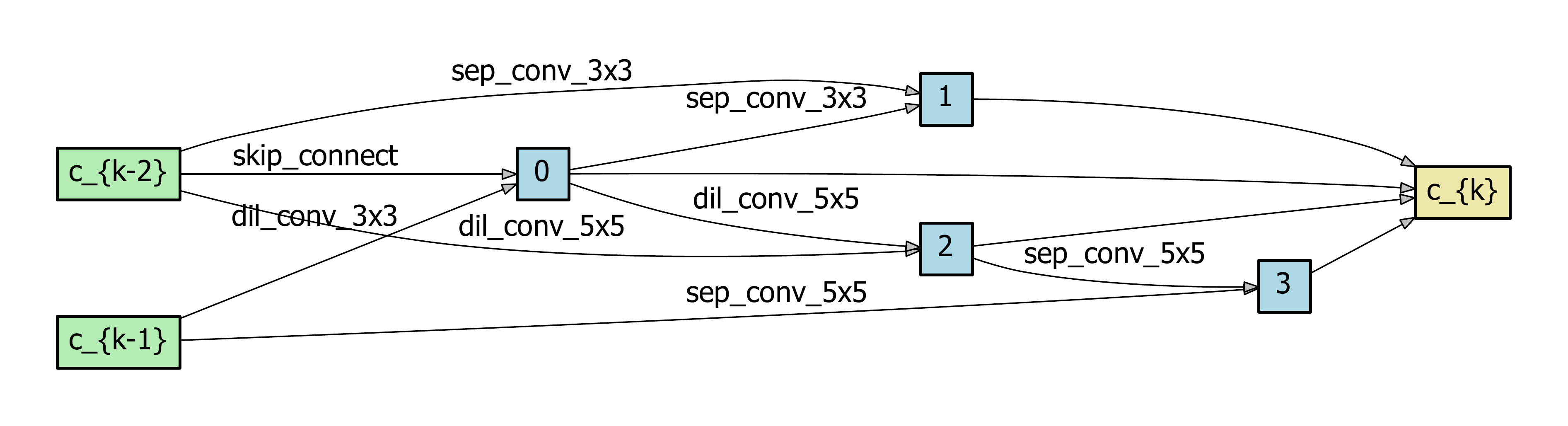}}
\subfigure[the reduction cell of $\mathcal{S}_1$-C]{\includegraphics[width=0.45\textwidth]{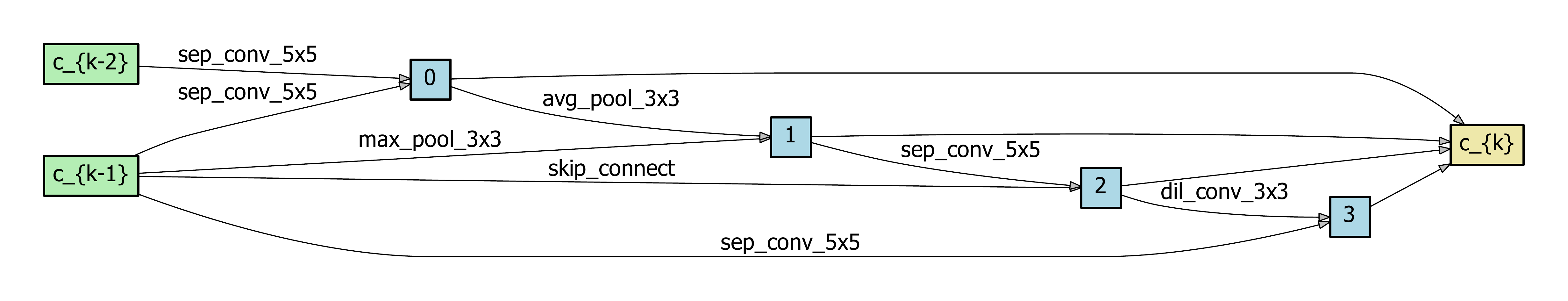}}\\
\vspace{0.2cm}
\subfigure[the normal cell of $\mathcal{S}_1$-F]{\includegraphics[width=0.45\textwidth]{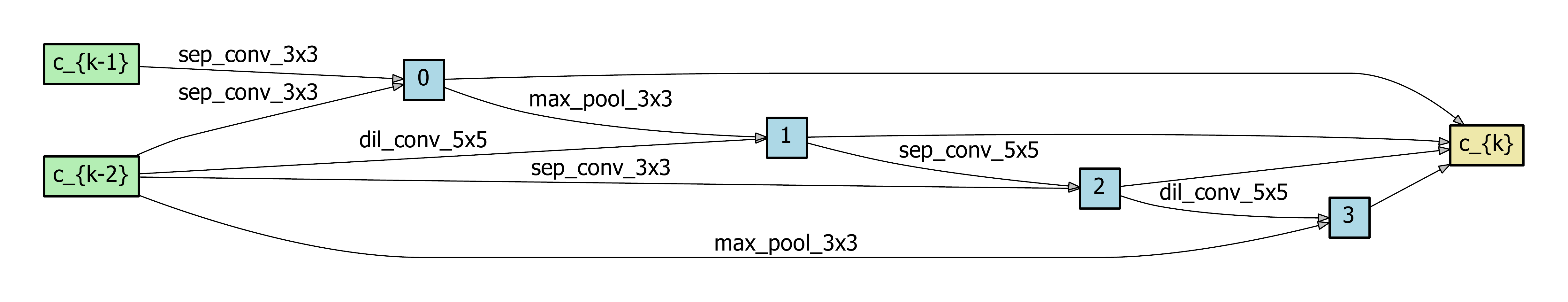}}
\subfigure[the reduction cell of $\mathcal{S}_1$-F]{\includegraphics[width=0.45\textwidth]{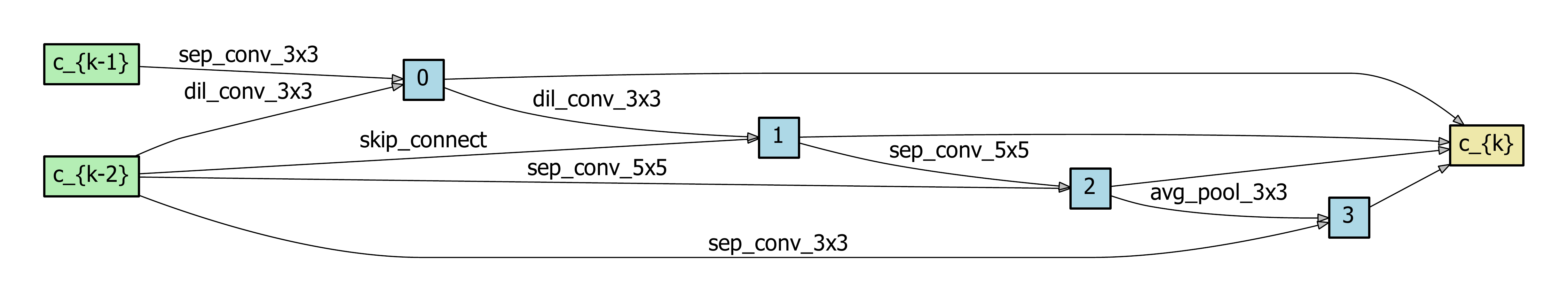}}\\
\vspace{0.4cm}
\caption{Architectures searched in $\mathcal{S}_1$ with pruned or fixed edges.}
\label{fig:architectures_S1}
\end{figure*}

\begin{figure*}
\centering
\subfigure[the architecture of $\mathcal{S}_2$-A-A]{\includegraphics[width=0.9\textwidth]{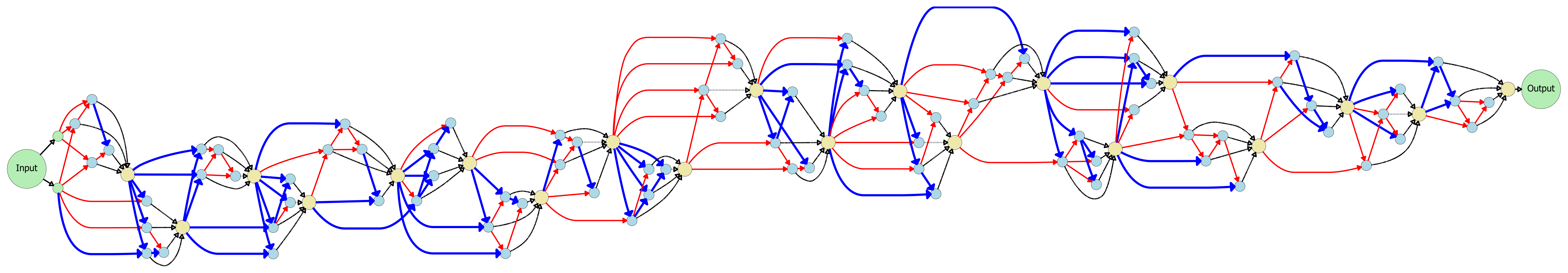}}
\subfigure[the architecture of $\mathcal{S}_2$-A-B]{\includegraphics[width=0.9\textwidth]{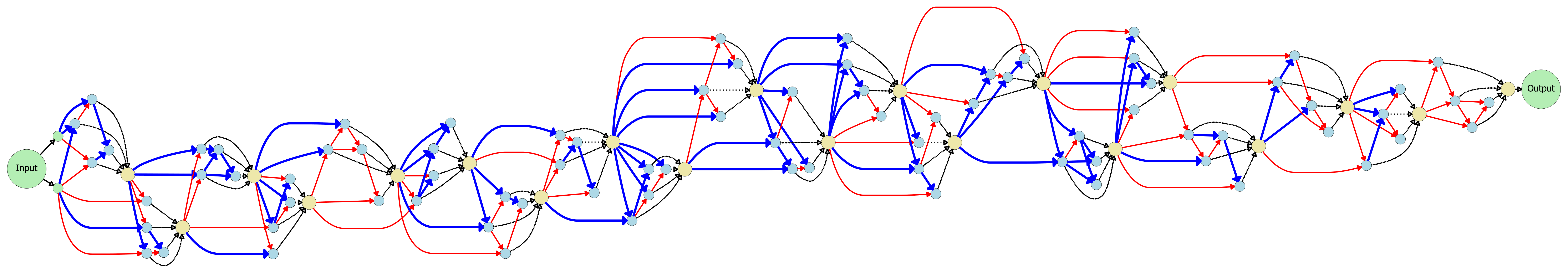}}
\subfigure[the architecture of $\mathcal{S}_2$-A-C]{\includegraphics[width=0.9\textwidth]{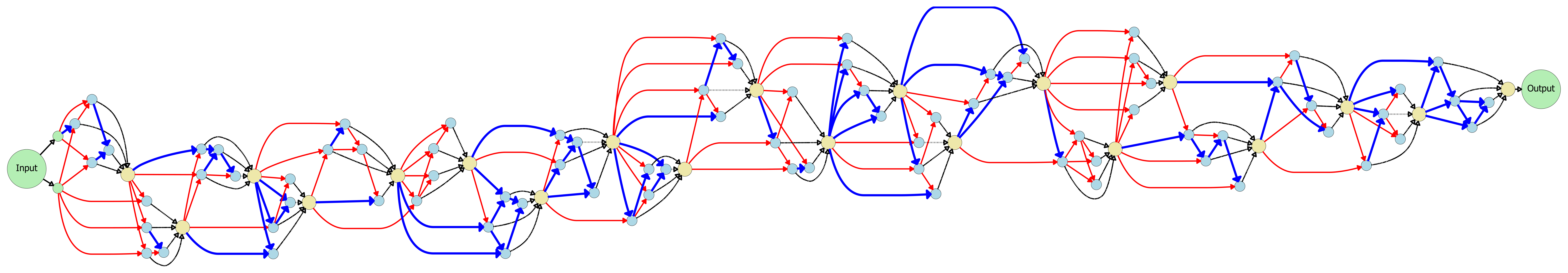}}
\subfigure[the architecture of $\mathcal{S}_2$-B-A]{\includegraphics[width=0.9\textwidth]{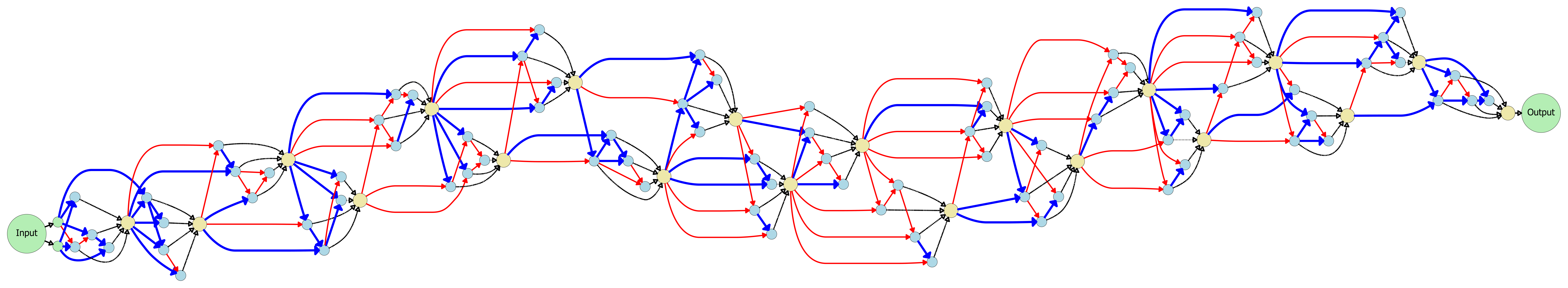}}
\subfigure[the architecture of $\mathcal{S}_2$-B-B]{\includegraphics[width=0.9\textwidth]{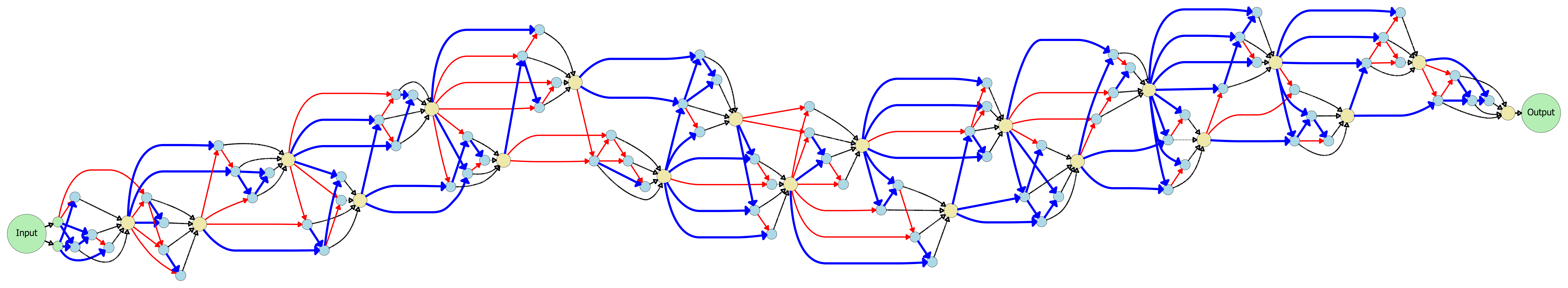}}
\subfigure[the architecture of $\mathcal{S}_2$-B-C]{\includegraphics[width=0.9\textwidth]{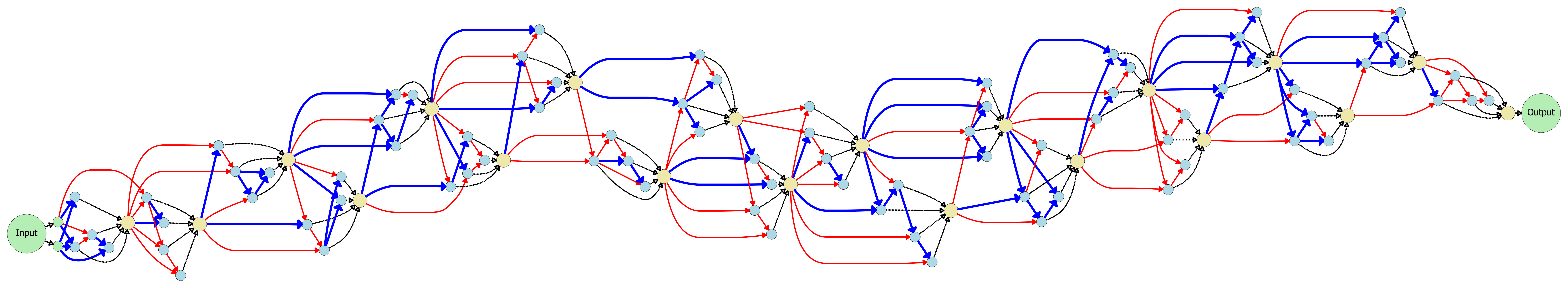}}
\caption{\textbf{(Part I)} Architectures searched in $\mathcal{S}_2$ with searched or fixed edges, in which red thin, blue bold, and black dashed arrows indicate \textit{skip-connect}, \textit{sep-conv-3x3}, and \textit{channel-wise concatenation}, respectively. \textit{This figure is best viewed in color.}.}
\label{fig:architectures_S2_P1}
\end{figure*}

\begin{figure*}
\centering
\subfigure[the architecture of $\mathcal{S}_2$-C-A]{\includegraphics[width=0.9\textwidth]{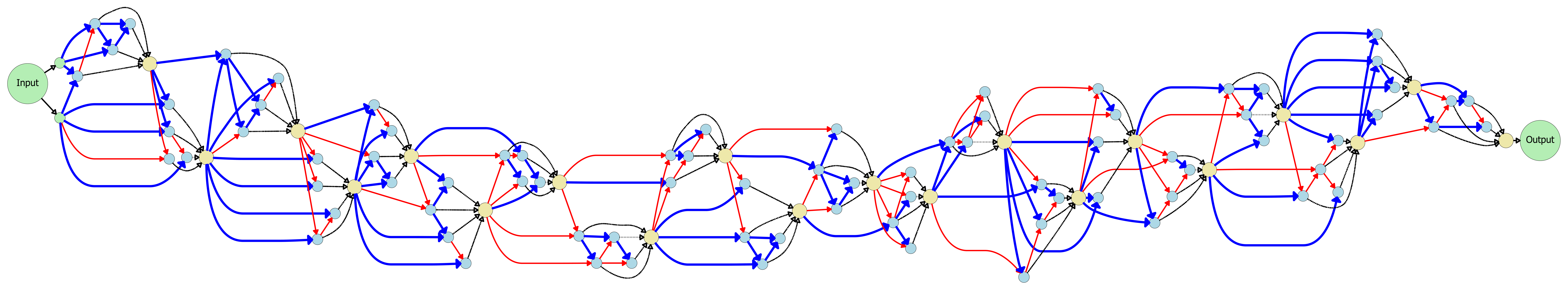}}
\subfigure[the architecture of $\mathcal{S}_2$-C-B]{\includegraphics[width=0.9\textwidth]{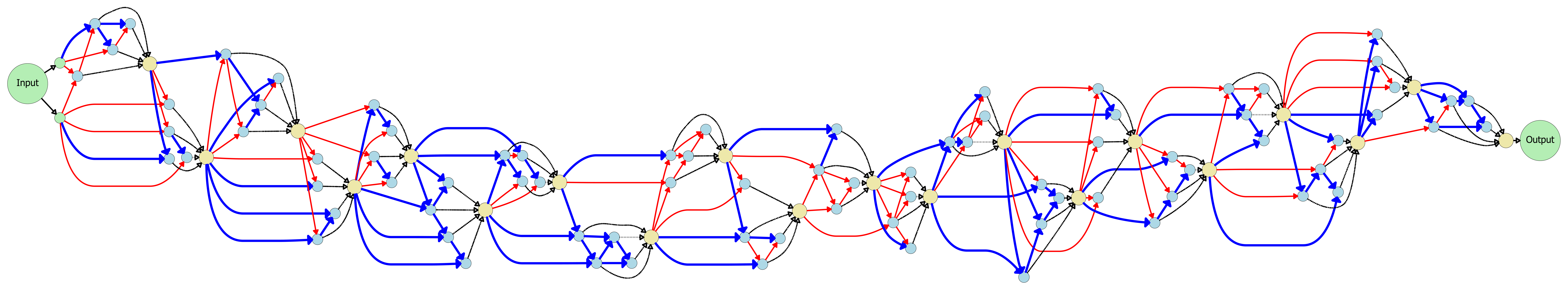}}
\subfigure[the architecture of $\mathcal{S}_2$-C-C]{\includegraphics[width=0.9\textwidth]{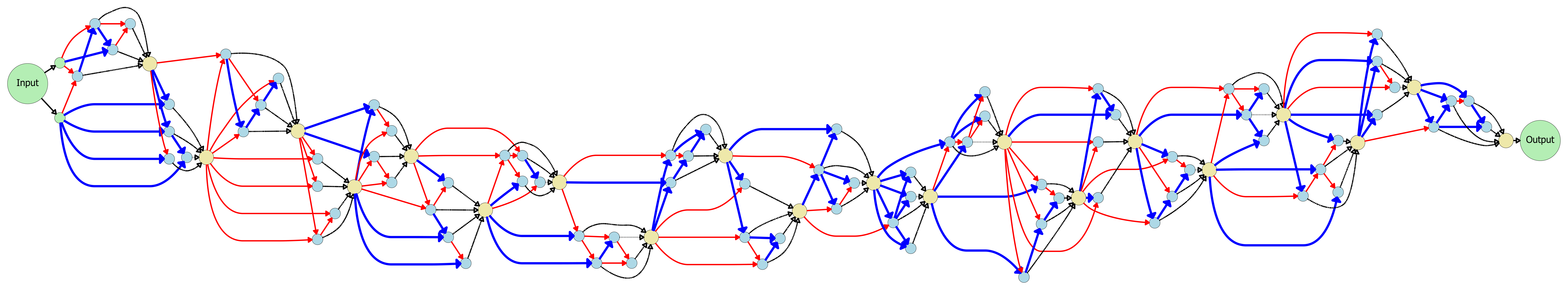}}
\subfigure[the architecture of $\mathcal{S}_2$-F-A]{\includegraphics[width=1.0\textwidth]{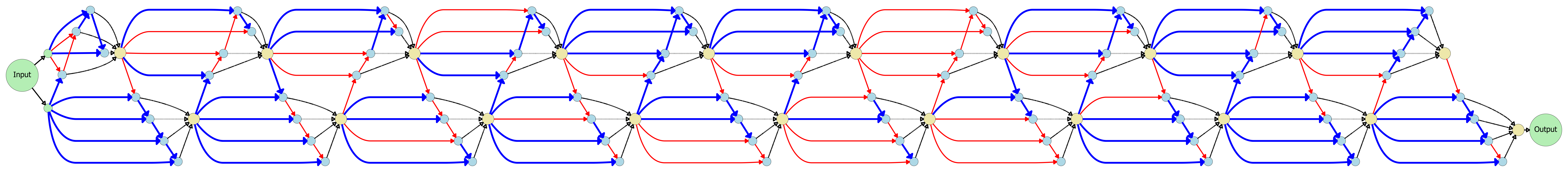}}
\subfigure[the architecture of $\mathcal{S}_2$-F-B]{\includegraphics[width=1.0\textwidth]{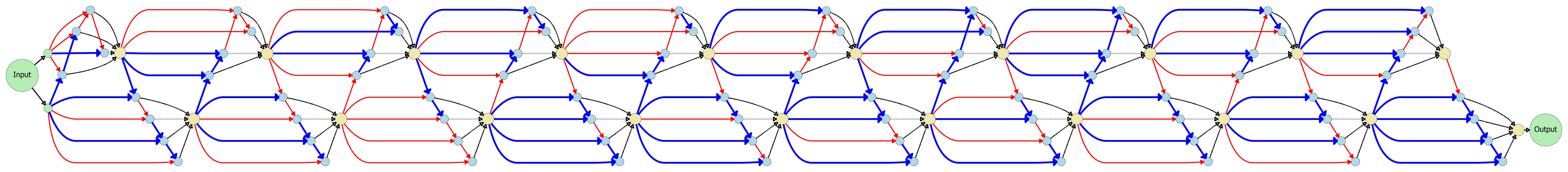}}
\subfigure[the architecture of $\mathcal{S}_2$-F-C]{\includegraphics[width=1.0\textwidth]{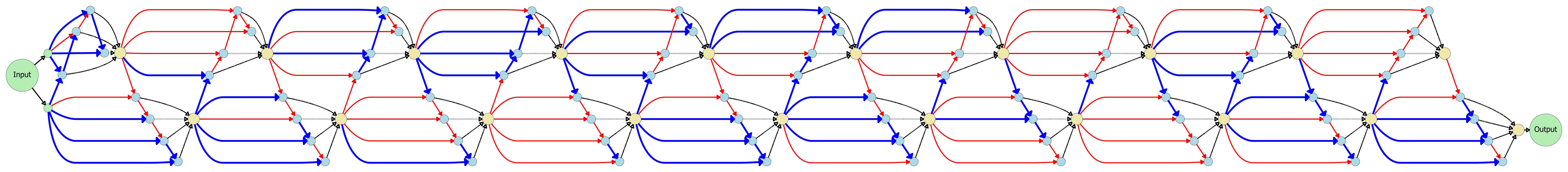}}
\caption{\textbf{(Part II)} Architectures searched in $\mathcal{S}_2$ with searched or fixed edges, in which red thin, blue bold, and black dashed arrows indicate \textit{skip-connect}, \textit{sep-conv-3x3}, and \textit{channel-wise concatenation}, respectively. \textit{This figure is best viewed in color.}.}
\label{architectures_S2_P2}
\end{figure*}

\begin{figure*}
\centering
\subfigure[the normal cell of $\mathcal{S}_3$-A]{\includegraphics[width=0.45\textwidth]{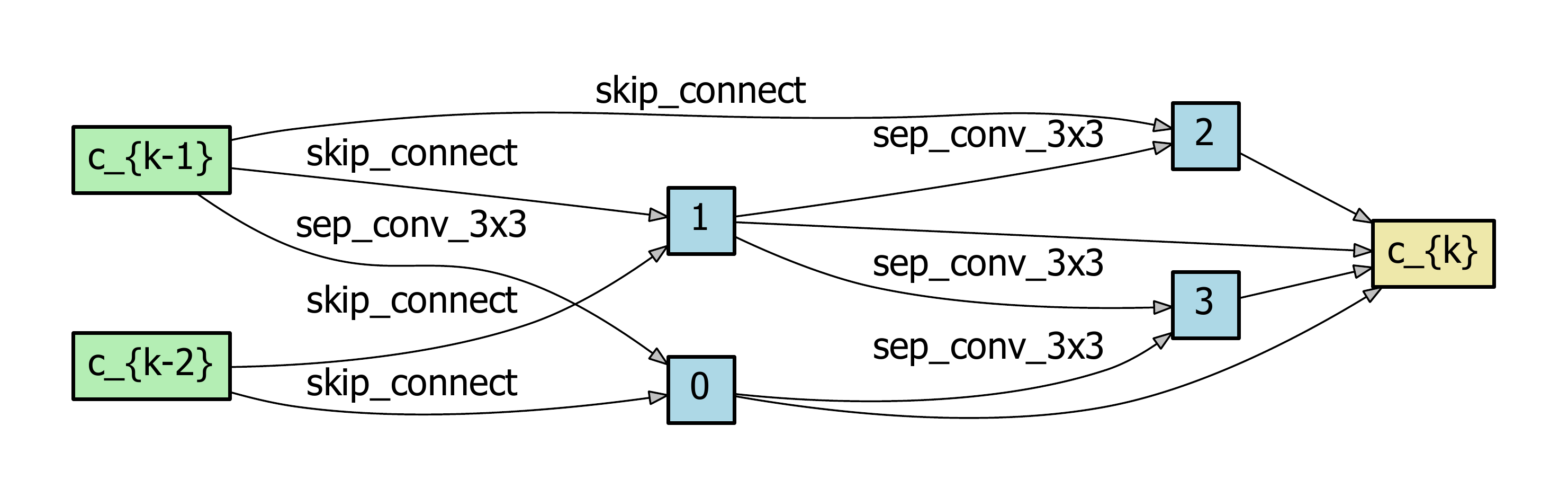}}
\subfigure[the reduction cell of $\mathcal{S}_3$-A]{\includegraphics[width=0.45\textwidth]{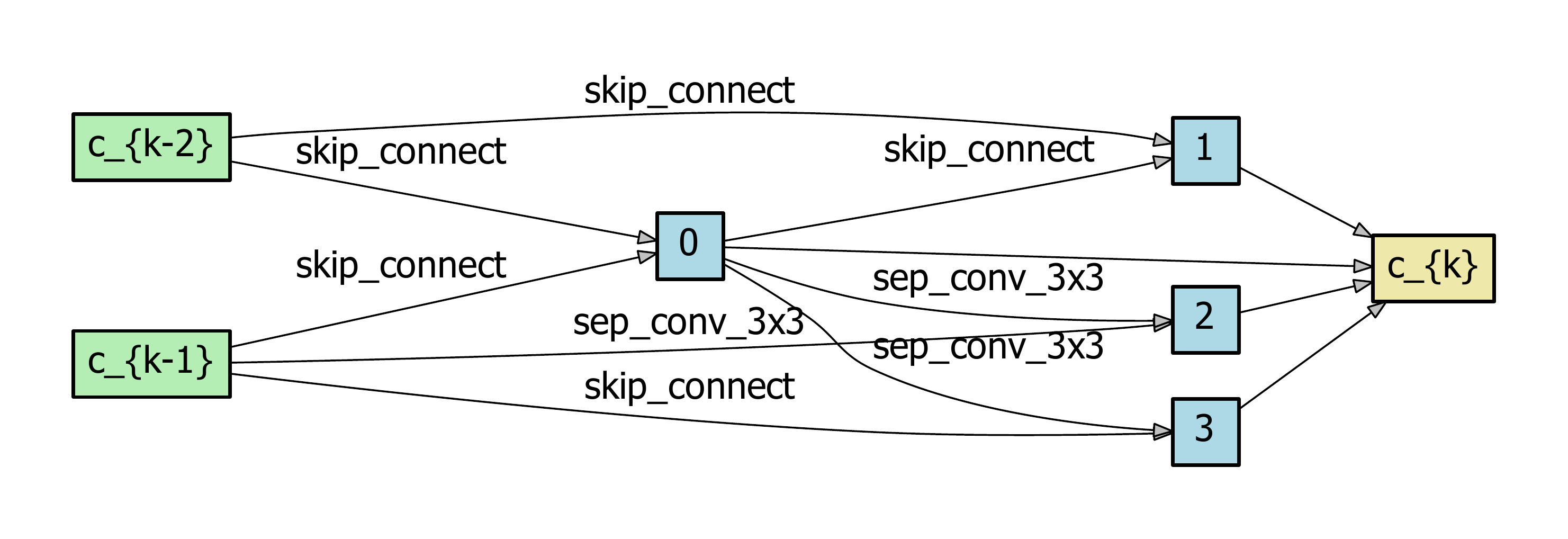}}\\
\vspace{0.2cm}
\subfigure[the normal cell of $\mathcal{S}_3$-B]{\includegraphics[width=0.45\textwidth]{S3_B_normal.pdf}}
\subfigure[the reduction cell of $\mathcal{S}_3$-B]{\includegraphics[width=0.45\textwidth]{S3_B_reduction.pdf}}\\
\vspace{0.2cm}
\subfigure[the normal cell of $\mathcal{S}_3$-C]{\includegraphics[width=0.45\textwidth]{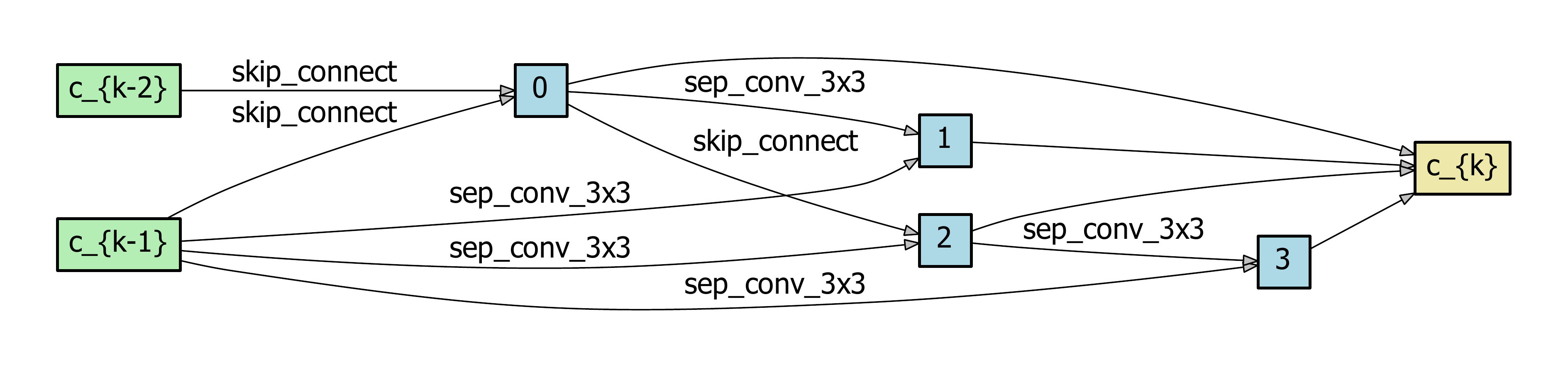}}
\subfigure[the reduction cell of $\mathcal{S}_3$-C]{\includegraphics[width=0.45\textwidth]{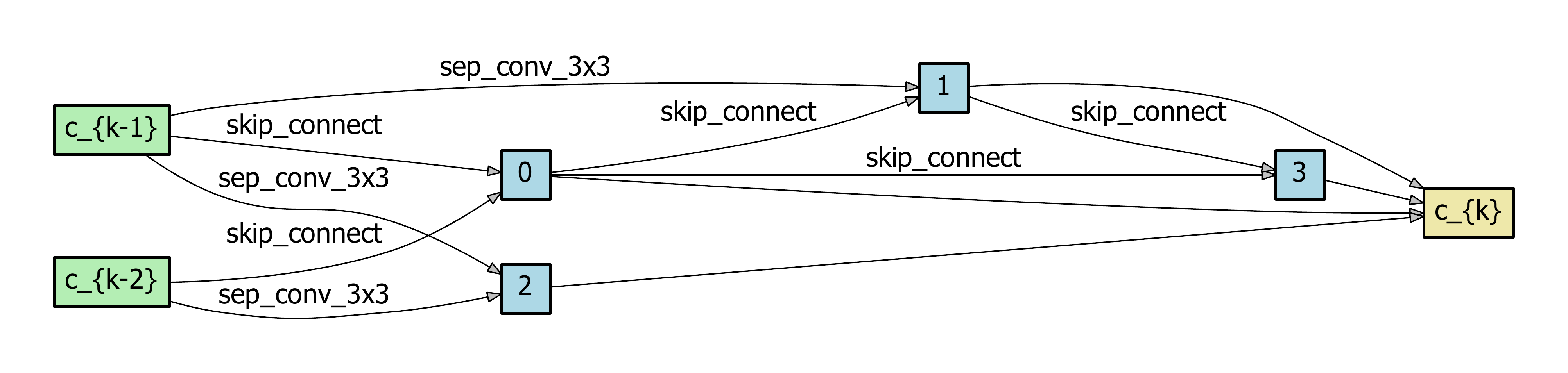}}\\
\subfigure[the normal cell of $\mathcal{S}_3$-D-A]{\includegraphics[width=0.45\textwidth]{S3_D_A_normal.pdf}}
\vspace{0.2cm}
\subfigure[the reduction cell of $\mathcal{S}_3$-D-A]{\includegraphics[width=0.45\textwidth]{S3_D_A_reduction.pdf}}\\
\vspace{0.2cm}
\subfigure[the normal cell of $\mathcal{S}_3$-D-B]{\includegraphics[width=0.45\textwidth]{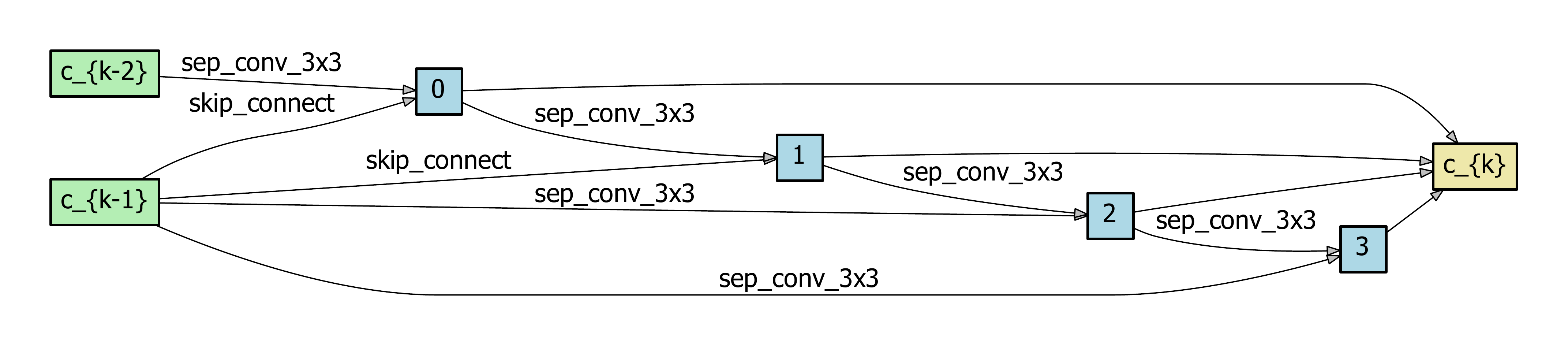}}
\subfigure[the reduction cell of $\mathcal{S}_3$-D-B]{\includegraphics[width=0.45\textwidth]{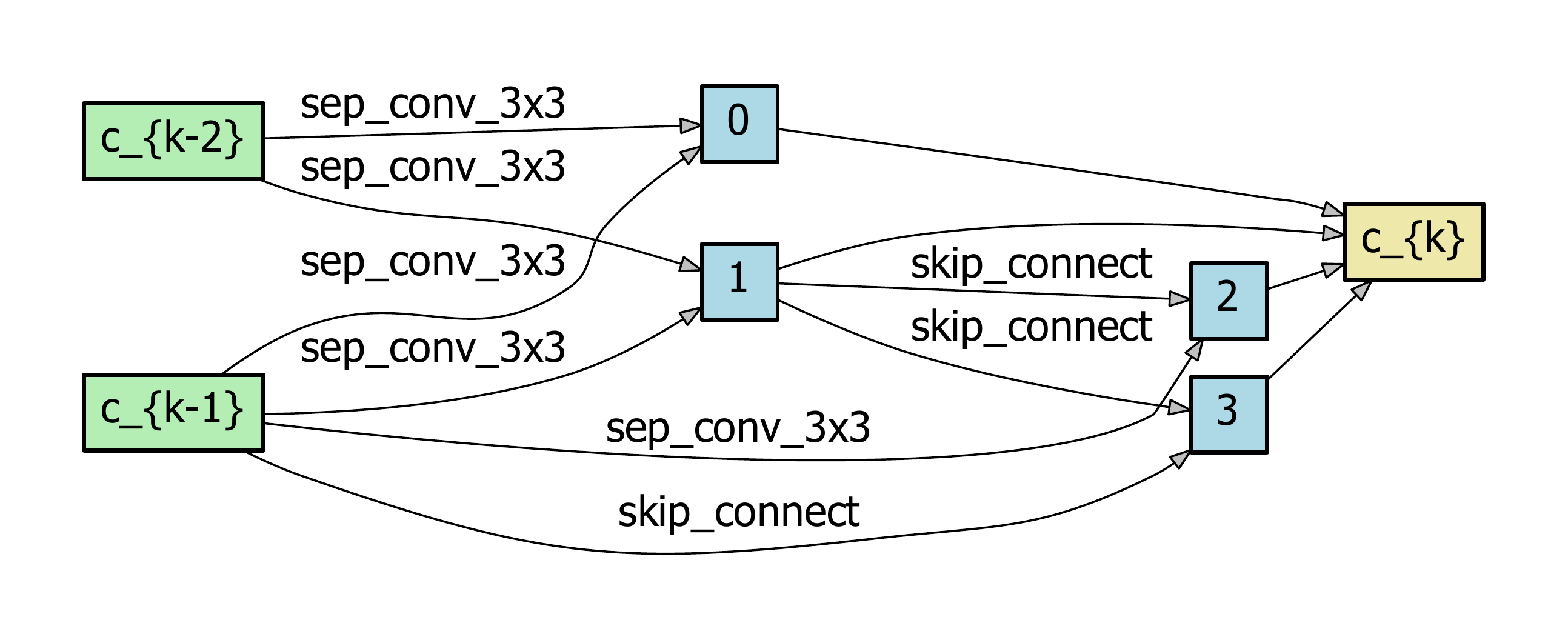}}\\
\vspace{0.2cm}
\subfigure[the normal cell of $\mathcal{S}_3$-D-C]{\includegraphics[width=0.45\textwidth]{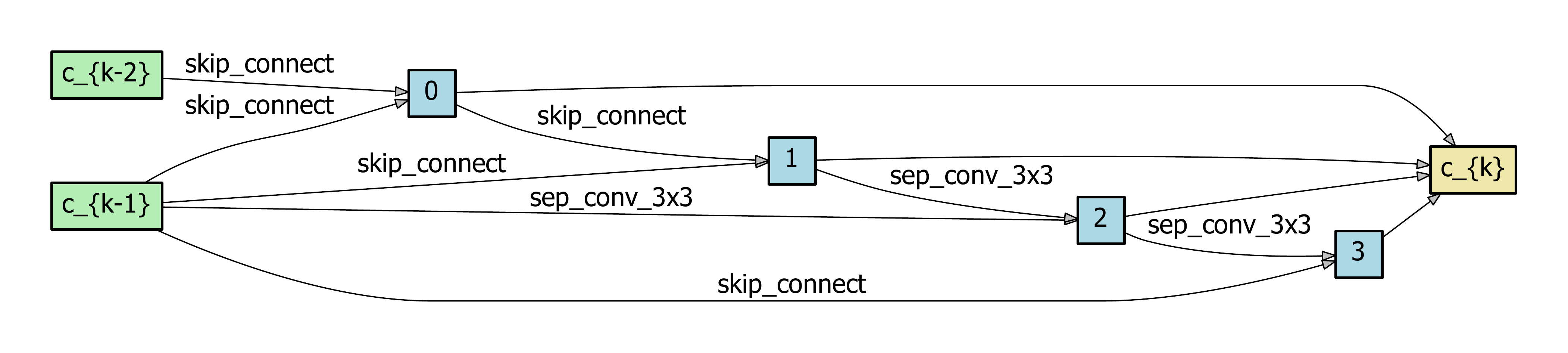}}
\subfigure[the reduction cell of $\mathcal{S}_3$-D-C]{\includegraphics[width=0.45\textwidth]{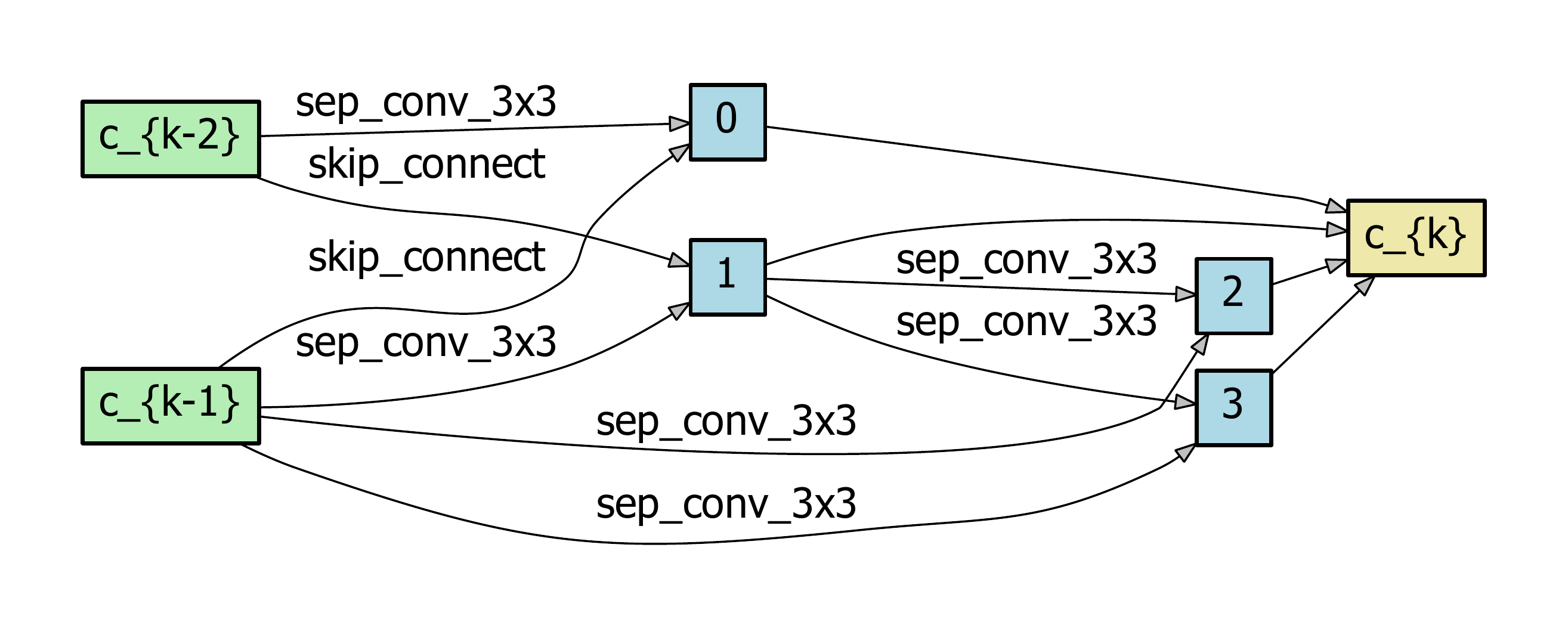}}\\
\vspace{-0.4cm}
\caption{Architectures searched in $\mathcal{S}_3$.}
\label{fig:architectures_S3}
\end{figure*}

All the searched architectures are shown in Figures~\ref{fig:architectures_S1}--\ref{fig:architectures_S3}.

\begin{table*}[!h]
\begin{center}
\begin{tabular}{lcccccc}
\hline
\textbf{\multirow{2}{*}{Architecture}} & \multicolumn{4}{c}{\textbf{Test Err. (\%)}} & \textbf{Params} & \textbf{Search Cost}\\
\cmidrule(lr){2-5} & \textbf{Run \#1} & \textbf{Run \#2} & \textbf{Run \#3} & \textbf{average} & \textbf{(M)} & \textbf{(GPU-days)}\\
\hline
Amended-DARTS, $\mathcal{S}_1$-A, pruned edges & 2.81 & 2.65 & 2.67 & 2.71$\pm$0.09 & 3.3 & 1.7\\
Amended-DARTS, $\mathcal{S}_1$-B, pruned edges & 2.82 & 2.73 & 2.90 & 2.82$\pm$0.09 & 2.9 & 1.7\\
Amended-DARTS, $\mathcal{S}_1$-C, pruned edges & 2.82 & 2.68 & 2.73 & 2.74$\pm$0.07 & 3.8 & 1.7\\
Amended-DARTS, $\mathcal{S}_1$-F, fixed edges & 2.69 & 2.68 & 3.05 & 2.81$\pm$0.21 & 3.5 & 1.0\\
\hline
Amended-DARTS, $\mathcal{S}_2$-A-A, searched edges & 2.67 & 2.79 & 2.66 & 2.71$\pm$0.07 & 2.8 & 3.1\\
Amended-DARTS, $\mathcal{S}_2$-A-B, searched edges & 2.62 & 2.79 & 2.56 & 2.66$\pm$0.12 & 3.0 & 3.1\\
Amended-DARTS, $\mathcal{S}_2$-A-C, searched edges & 2.67 & 2.69 & 2.69 & 2.68$\pm$0.01 & 2.9 & 3.1\\
Amended-DARTS, $\mathcal{S}_2$-B-A, searched edges & 2.59 & 2.73 & 2.57 & 2.63$\pm$0.09 & 3.0 & 3.1\\
Amended-DARTS, $\mathcal{S}_2$-B-B, searched edges & 2.64 & 2.57 & 2.48 & 2.56$\pm$0.08 & 3.4 & 3.1\\
Amended-DARTS, $\mathcal{S}_2$-B-C, searched edges & 2.64 & 2.66 & 2.80 & 2.70$\pm$0.09 & 3.0 & 3.1\\
Amended-DARTS, $\mathcal{S}_2$-C-A, searched edges & 2.56 & 2.63 & 2.69 & 2.66$\pm$0.09 & 3.3 & 3.1\\
Amended-DARTS, $\mathcal{S}_2$-C-B, searched edges & 2.61 & 2.70 & 2.77 & 2.69$\pm$0.08 & 3.2 & 3.1\\
Amended-DARTS, $\mathcal{S}_2$-C-C, searched edges & 2.76 & 2.62 & 2.57 & 2.65$\pm$0.10 & 3.2 & 3.1\\
Amended-DARTS, $\mathcal{S}_2$-F-A, fixed edges & 2.51 & 2.51 & 2.77 & 2.60$\pm$0.15 & 3.6 & 1.1\\
Amended-DARTS, $\mathcal{S}_2$-F-B, fixed edges & 2.59 & 2.64 & 2.58 & 2.60$\pm$0.03 & 3.5 & 1.1\\
Amended-DARTS, $\mathcal{S}_2$-F-C, fixed edges & 2.64 & 2.60 & 2.72 & 2.65$\pm$0.06 & 2.9 & 1.1\\
\hline
Amended-DARTS, $\mathcal{S}_3$-A, pruned edges & 2.65 & 2.81 & 2.83 & 2.76$\pm$0.10 & 3.1 & 0.8\\
Amended-DARTS, $\mathcal{S}_3$-B, pruned edges & 2.57 & 2.71 & 2.64 & 2.64$\pm$0.07 & 3.2 & 0.8\\
Amended-DARTS, $\mathcal{S}_3$-C, pruned edges & 2.59 & 2.67 & 2.77 & 2.68$\pm$0.09 & 3.4 & 0.8\\
Amended-DARTS, $\mathcal{S}_3$-D-A, searched edges & 2.52 & 2.43 & 2.69 & 2.55$\pm$0.13 & 3.2 & 3.0\\
Amended-DARTS, $\mathcal{S}_3$-D-B, searched edges & 2.65 & 2.58 & 2.52 & 2.58$\pm$0.07 & 3.8 & 3.0\\
Amended-DARTS, $\mathcal{S}_3$-D-C, searched edges & 2.80 & 2.74 & 2.73 & 2.76$\pm$0.04 & 2.9 & 3.0\\
\hline
\end{tabular}
\end{center}
\vspace{-0.4cm}
\caption{Results of architecture search with different seeds in $\mathcal{S}_1$, $\mathcal{S}_2$, and $\mathcal{S}_3$ on CIFAR10.}
\label{tab:seeds_CIFAR10}
\end{table*}

\begin{table*}
\begin{center}
\begin{tabular}{lccccc}
\hline
\textbf{\multirow{2}{*}{Architecture}} & \multicolumn{2}{c}{\textbf{Test Err. (\%)}} & \textbf{Params} & $\times+$ & \textbf{Search Cost}\\
\cmidrule(lr){2-3}
&                            \textbf{top-1} & \textbf{top-5} & \textbf{(M)} & \textbf{(M)} & \textbf{(GPU-days)}\\
\hline
Amended-DARTS, $\mathcal{S}_1$-A, pruned edges & 24.7 & 7.6  & 5.2 & 586 & 1.7\\
Amended-DARTS, $\mathcal{S}_1$-F, fixed edges & 24.6 & 7.4  & 5.2 & 587 & 1.0\\
\hline
Amended-DARTS, $\mathcal{S}_2$-A-A, searched edges & 24.7 & 7.5 & 4.9 & 576 & 3.1\\
Amended-DARTS, $\mathcal{S}_2$-B-A, searched edges & 24.3 & 7.3 & 5.2 & 596 & 3.1\\
Amended-DARTS, $\mathcal{S}_2$-B-B, searched edges & 24.6 & 7.5 & 5.3 & 585 & 3.1\\
Amended-DARTS, $\mathcal{S}_2$-B-C, searched edges & 24.3 & 7.3 & 5.3 & 592 & 3.1\\
Amended-DARTS, $\mathcal{S}_2$-C-A, searched edges & 24.3 & 7.5 & 5.2 & 592 & 3.1\\
Amended-DARTS, $\mathcal{S}_2$-F-A, fixed edges & 24.3 & 7.4  & 5.5 & 590 & 1.1\\
\hline
\end{tabular}
\end{center}
\vspace{-0.4cm}
\caption{Transferring some of the searched architectures on CIFAR10 to ILSVRC2012. The \textit{mobile setting} is used to determine the basic channel width of each transferred architecture.}
\label{tab:seeds_ILSVRC2012}
\end{table*}

\end{document}